\documentclass[lettersize,journal]{IEEEtran}
\usepackage[caption=false,font=footnotesize,labelfont=rm,textfont=rm]{subfig}
\usepackage{cite}
\usepackage{color}
\usepackage{amssymb,amsfonts,mathrsfs,mathtools}
\usepackage{algorithmic}
\usepackage{algorithm}
\usepackage{array}
\usepackage{stfloats}
\usepackage{url}
\usepackage{verbatim}
\usepackage{graphicx}
\usepackage{textcomp}
\usepackage{amsthm}
\usepackage{multicol}
\usepackage{multirow}

\theoremstyle{definition}
\newtheorem{theorem}{Theorem}

\newtheorem{remark}{Remark}

\begin{document}

\title{Epigraph Form Reinforcement Learning-based Safe Trajectory Tracking Control for Connected and Autonomous Vehicles}

\author{Xiaoke Yang, Jin Chen, ~\IEEEmembership{Member,~IEEE}, Yafei Wang, ~\IEEEmembership{Member,~IEEE}, Qi Zhang, Ming Li, ~\IEEEmembership{Member,~IEEE}, Haiwen Wu, ~\IEEEmembership{Member,~IEEE}, Chengliang Yin, Yong Zhang
\thanks{Xiaoke Yang, Jin Chen, Yafei Wang, Qi Zhang, Chengliang Yin, and Yong Zhang are with the School of Mechanical Engineering, Shanghai Jiao Tong University, Shanghai 200240, China (email: xkyang1997@sjtu.edu.cn, chenjin920414@sjtu.edu.cn, wyfjlu@sjtu.edu.cn, zhangqi\textunderscore mech@sjtu.edu.cn, clyin1965@sjtu.edu.cn, yongzhang1977@sjtu.edu.cn).}
\thanks{Ming Li is with KTH Royal Institute of Technology, Stockholm SE-10044, Sweden (email: ming3@kth.se).}
\thanks{Haiwen Wu is with the Department of Mechanical and Automation Engineering, The Chinese University of Hong Kong, Hong Kong 999077, China (email: haiwenwu@cuhk.edu.hk).}
}



\maketitle

\begin{abstract}
Safe trajectory tracking (STT) aims to enable autonomous vehicles to follow a reference trajectory accurately while avoiding collisions. This problem is particularly challenging for Connected and Autonomous Vehicles (CAVs) in highly dynamic environments. Existing non-learning-based STT methods often use simplified vehicle and interaction models to meet real-time requirements, which may result in suboptimal tracking accuracy. Although Deep Reinforcement Learning (DRL) can help address this limitation, widely used Lagrangian DRL methods often suffer from unstable training because increasing Lagrange multipliers amplify policy-gradient magnitudes. To address these issues, we propose a novel reinforcement learning framework for STT, called STT Control Epigraph Form RL (STTC-EFRL). In this framework, the STT problem is formulated as an epigraph-form min-max optimization problem. Based on this formulation, we develop a learning mechanism that uses a decreasing scalar related to the epigraph to adjust the policy gradient descent direction without increasing the gradient magnitude, thereby improving training stability. We also develop a computationally efficient ray-casting method to compute a polyhedron scaling factor for safety quantification. Experimental results, including Vehicle-in-the-Loop tests, show that STTC-EFRL achieves higher tracking accuracy than the considered non-learning-based baseline and more stable training than widely used Lagrangian DRL methods, while satisfying safety requirements in connected environments.
\end{abstract}

\begin{IEEEkeywords}
Safe trajectory tracking, connected and autonomous vehicles, deep reinforcement learning, obstacle avoidance, epigraph form.
\end{IEEEkeywords}

\section{Introduction}
\label{sec:introduction}

\subsection{Background}
\begin{figure}
\centering
\includegraphics[width=0.9\linewidth]{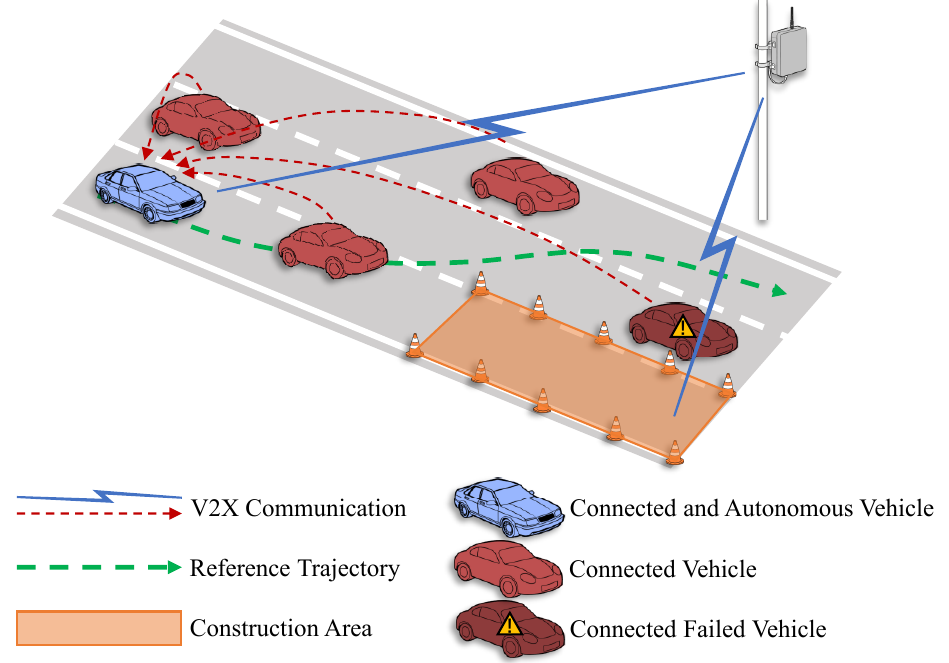}
\caption{Illustration of a typical STT task for a CAV in a V2X-enabled environment. The ego vehicle tracks a reference trajectory while interacting with roadside infrastructure and other connected vehicles.}
\label{fig:scenario_v2x}
\end{figure}

\IEEEPARstart{S}{afe} trajectory tracking (STT) is a fundamental task for autonomous vehicles (AVs), requiring a vehicle to follow a reference trajectory accurately while avoiding collisions. High-quality STT depends on accurate tracking of both the spatial path and the temporal speed profile. This task becomes especially challenging in complex scenarios such as roundabouts \cite{lin2025a}, intersections \cite{li2023safe}, and lane-merging zones \cite{liu2025lane, chen2026}, where multiple traffic participants interact in a rapidly changing environment. In such settings, on-board sensing alone often suffers from limited field of view and occlusions, making it difficult to accurately estimate the states and geometry of surrounding obstacles \cite{zha2025heterogeneous}. Vehicle-to-Everything (V2X) communication can help alleviate this limitation by allowing connected vehicles and roadside units to share real-time information about surrounding obstacles and road conditions, as illustrated in Fig. \ref{fig:scenario_v2x}.

\IEEEpubidadjcol

\begin{figure}[tbp]
\centering
\subfloat[]{%
\includegraphics[width=0.95\linewidth]{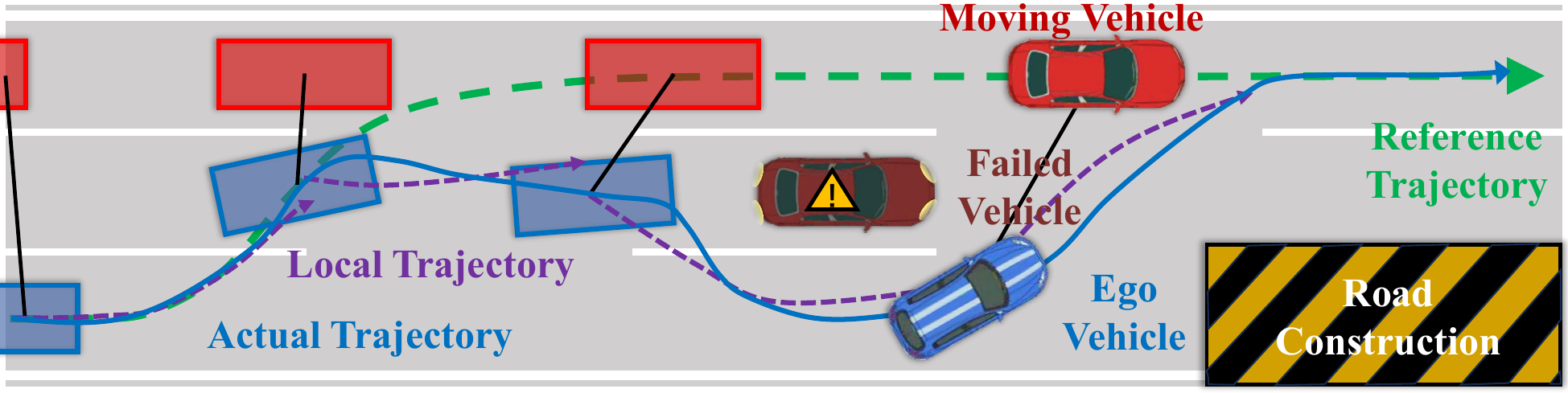}
\label{fig:lane_change_scenario_for_introduction_hierarchical}}
\vfill
\subfloat[]{%
\includegraphics[width=0.95\linewidth]{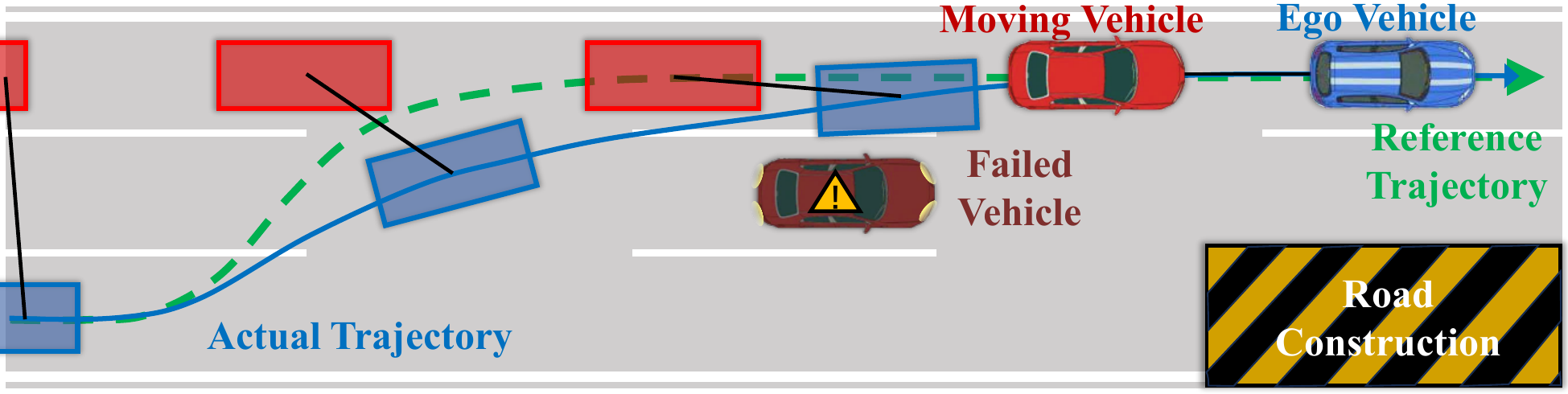}
\label{fig:lane_change_scenario_for_introduction_integrated}}
\caption{Comparison of STT performance in an obstructed-lane scenario. Rectangles denote vehicles at historical time steps. In (a), the hierarchical controller deviates from the reference trajectory to remain safe, which reduces tracking accuracy. In (b), the integrated controller better follows the reference trajectory while maintaining safety.}
\label{fig:lane_change_scenario_for_introduction}
\end{figure}

A common solution for STT of connected and autonomous vehicles (CAVs) is a hierarchical architecture, which decomposes the task into local trajectory planning and trajectory tracking control \cite{han2024uniform}. Recently, hierarchical safe Deep Reinforcement Learning (DRL) methods have also adopted this paradigm for STT by using a high-level Reinforcement Learning (RL) policy for trajectory planning and a low-level controller for execution and safety filtering \cite{huang2022efficient, qi2022learning}. Although modular and widely used to reduce learning dimensionality, this architecture may suffer from error accumulation and mismatched assumptions at the module interface between planning and control \cite{paden2016survey}. Specifically, the high-level policy or planner usually relies on simplified kinematic models and generates reference commands that may exceed the physical limits of the low-level vehicle dynamics in highly dynamic scenarios. The low-level controller struggles to track these commands while enforcing safety, leading to tracking latency. Because the high-level policy assumes ideal downstream execution, these discrepancies may compound over time, forcing the vehicle into conservative maneuvers or degraded tracking accuracy in highly dynamic scenarios, as shown in Fig. \ref{fig:lane_change_scenario_for_introduction_hierarchical}.

Integrated approaches address these structural limitations by handling planning and control within a unified framework, ensuring that trajectory tracking and safety constraints are considered simultaneously \cite{reza2021integrated, erlien2016shared, zhang2021human}. By mapping environmental states directly to physically feasible control actions, these methods eliminate the explicit module interfaces, reducing the aforementioned error accumulation and dynamic mismatches. Consequently, integrated approaches have potential in maintaining consistency between decision-making and control execution, demonstrating a superior capability to balance tracking accuracy and safety in highly dynamic STT tasks, as illustrated in Fig. \ref{fig:lane_change_scenario_for_introduction_integrated}.

Nevertheless, integrated STT for CAVs in complex scenarios remains difficult because of nonlinear vehicle dynamics, nonconvex constraints, high dimensionality, and rapidly changing environments. Conventional optimization-based methods can therefore be computationally expensive and may get trapped in poor local minima. DRL provides a promising alternative \cite{hsu2021safety}, as it can learn integrated control policies with low online computational cost while balancing tracking accuracy and safety. However, existing safe DRL frameworks often suffer from training instability, which limits controller reliability \cite{so23solving}.

\subsection{Related Works and Contributions}
Integrated STT methods can be broadly divided into non-learning-based and learning-based approaches. Non-learning-based approaches usually incorporate safety constraints into model-based control methods. Zhang et al. \cite{zhang2021secure} combined barrier functions with backstepping control for STT, while Han et al. \cite{han2024uniform} employed barrier-function-based yaw-angle tracking to solve the STT task. In addition, many studies combine Control Lyapunov Functions (CLFs) and Control Barrier Functions (CBFs) with dynamic models to formulate online STT optimization problems \cite{chen2017obstacle, li2024quadrotor, chen2025range}. However, to maintain real-time responsiveness in highly dynamic CAV scenarios, these methods often rely on simplified vehicle dynamics and interaction models, which can limit the achievable tracking accuracy.

Learning-based integrated control methods learn policies from data, aiming to reduce online computational burden while balancing safety and control performance. Recently, DRL algorithms have demonstrated remarkable success in various complex integrated autonomous control tasks. For instance, Yang et al. \cite{yang2026optimal} achieve optimal trajectory tracking for autonomous Unmanned Aerial Vehicles (UAVs) by proposing a predefined-time DRL algorithm. To resist cyber attacks, Yang et al. \cite{yang2025secure} solve the secure containment control problem for multi-UAV systems utilizing a fixed-time convergent DRL method. Similarly, Gong et al. \cite{gong2025secure} accomplish secure formation control against false data injection attacks by developing a game-based fixed-time convergent DRL algorithm. Furthermore, for high-level decision-making, Yang et al. \cite{yang2025aav} execute weapon target assignment tasks by designing a state split flow deep Q-learning network.

In the context of integrated DRL-based safety-critical STT control, recent schemes have gained attention by utilizing CBFs either as rigorous safety shields to filter unsafe actions \cite{zhang2025control, hou2024safe} or as modules to modify primitive evaluation metrics \cite{ma2021model}. Though effective in theory, these methods typically require substantial domain knowledge and rely heavily on accurate, explicit dynamic models and environment models to synthesize the barrier. Consequently, their performance can be highly sensitive to model mismatches in complex and highly dynamic CAV scenarios.

Other approaches formulate safe DRL as constrained Markov decision processes (CMDPs) \cite{chen2022reinforcement}. Conventional CMDP-based methods typically combine control objectives and safety constraints through weighted sums. Although such formulations are easy to optimize, selecting appropriate weights is difficult, and the choice of weights directly affects training stability and safety violations. Lagrangian methods introduce adaptive weight variables to enforce constraints more systematically. Algorithms such as PPO-Lagrangian \cite{gu2023safe}, Lagrangian Soft Actor-Critic \cite{zhang2021safe}, and hierarchical safety enhancement \cite{gu2023safevehicle} have shown promising performance in safety-critical control problems. However, these methods can still suffer from training instability because the weight variables scale policy gradients \cite{so23solving}.

As an alternative, Altarovici et al. \cite{altarovici2013general} introduced epigraph transformations for continuous-time control optimization. So et al. \cite{so23solving} later proposed Epigraph Form Reinforcement Learning (EFRL) for discrete-time stabilization problems and demonstrated convergence and safety through simulations and quadrotor experiments. However, these studies focus on safe goal-reaching problems and do not directly apply to STT for CAVs, where the reference trajectory changes over time. Motivated by this gap, this paper studies STT for CAVs in highly dynamic environments with multiple static and dynamic obstacles. To improve tracking accuracy, maintain safety, and reduce online computational burden, we develop a DRL-based integrated STT framework based on an epigraph-form reformulation of the STT problem.

In summary, the main contributions of this paper are as follows:
\begin{enumerate}
\item We propose a two-stage epigraph-form framework and develop the associated STT Control Epigraph Form Reinforcement Learning (STTC-EFRL) algorithm for the integrated STT of CAVs. The framework reformulates the STT problem into a min-max optimization problem: the unconstrained inner problem is effectively solved via DRL with a decreasing epigraph variable to guide policy updates and improve training stability; the outer problem is efficiently solved by a bisection-based root-finding procedure.

\item We develop a computationally efficient ray-casting method to calculate the polyhedron scaling factor for rigorous primitive evaluation. This method introduces a direct approach to handle polyhedron obstacles with varying shapes, thereby significantly accelerating the precise primitive evaluation process during the training phase for DRL.

\item We analyze the theoretical properties of the controller learned by STTC-EFRL. We first study the closed-loop stability of the controller. Then, we conduct simulations and Vehicle-in-the-Loop (ViL) experiments as necessary validations. The results confirm that our approach achieves higher tracking accuracy than state-of-the-art methods. It also shows better training stability than Lagrangian DRL methods while meeting safety requirements.
\end{enumerate}

\subsection{Paper Organization}
The remainder of this paper is organized as follows. Section \ref{sec:preliminaries} introduces the notation used in this paper, presents the vehicle model, and formulates the STT problem for CAVs as a constrained optimal control problem (OCP). Section \ref{sec:methodology} presents the epigraph form of the constrained OCP and the proposed STTC-EFRL algorithm. It also analyzes the stability of the resulting controller and presents the scaling-factor computation method for primitive evaluation. Section \ref{sec:experiments} reports simulation training results and ViL experiments in lane-changing, overtaking and intersection scenarios. Section \ref{sec:conclusion} concludes the paper and discusses limitations and future research directions.
\section{Preliminaries}
\label{sec:preliminaries}
In this section, we introduce the notations and dynamic model used in this paper, and then mathematically formulate the STT control problem for CAVs.

\subsection{Notations}
Let $\mathbb{R}, \ \mathbb{N}, \ \mathbb{N}_{+}$ denote the sets of real numbers, natural numbers and positive natural numbers, respectively. For scalar $a \in \mathbb{R}$, $|a|$ denotes its absolute value. For $n \in \mathbb{N}$, we define $\mathbb{R}^n$ as the $n$-dimensional Euclidean space. The transpose of a vector $v \in \mathbb{R}^{n\times1}$ is denoted as $v^\top \in \mathbb{R}^{1\times n}$, and $\| v \|$ denotes its $L_2$ norm. Additionally, $\mathbf{E}$ denotes the expectation, $f \circ g$ denotes a composite function that is $f(g(\cdot))$, $\text{diag}(\cdot)$ denotes the diagonal matrix, and $\partial \mathcal{A}$ and $\operatorname{int}(\mathcal{A})$ respectively denote the boundary and the interior of the set $\mathcal{A}$.

\subsection{Vehicle Dynamics}
\begin{figure}[tbp]
\centering
\includegraphics[width=0.8\linewidth]{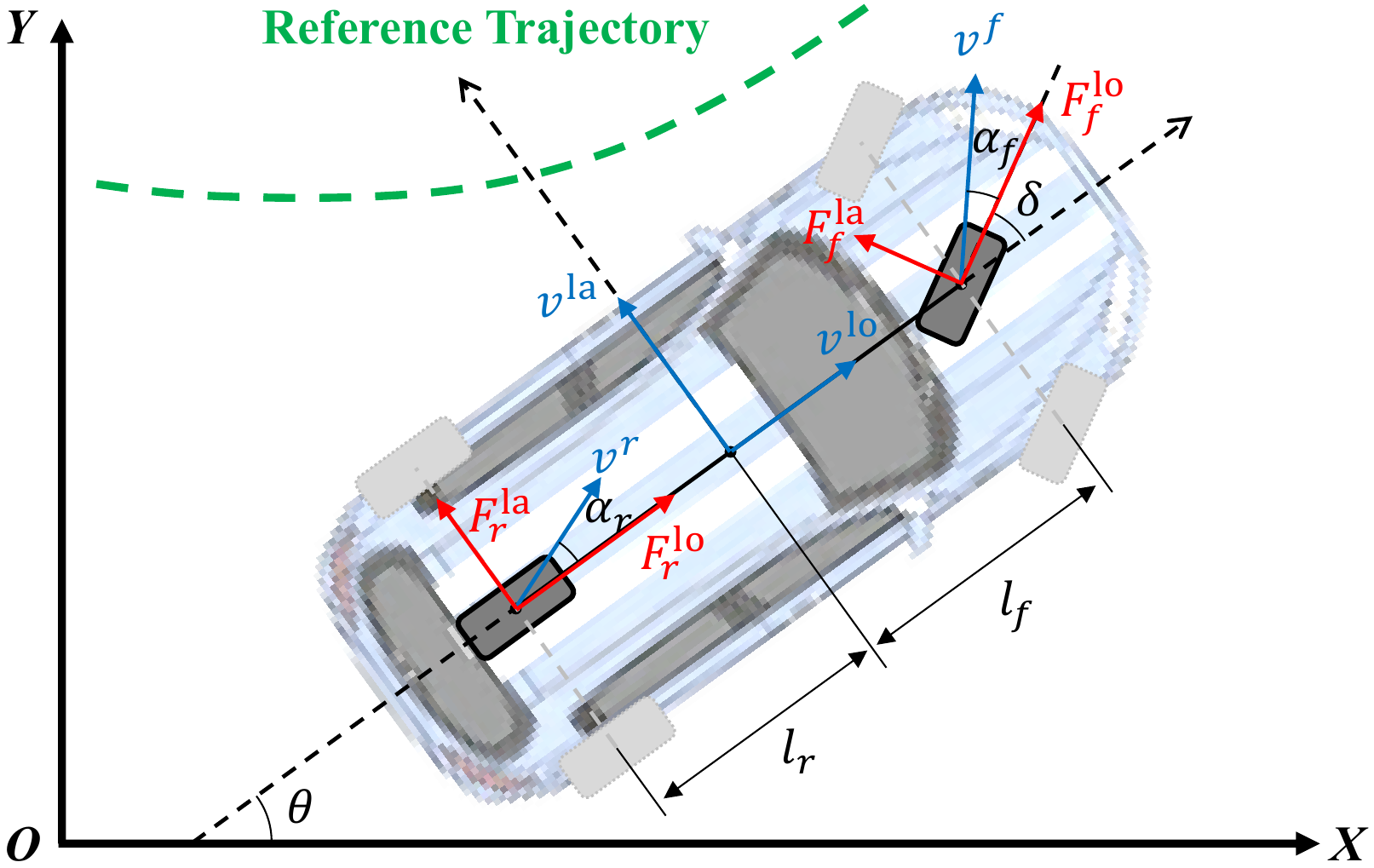}
\caption{Diagram of the 3-DOF bicycle model of a vehicle.}
\label{fig:3_DOF_dynamic_model}
\end{figure}

For the STT task of CAVs, a three-degree-of-freedom (3-DOF) vehicle dynamic model \cite{kong2015kinematic} is widely utilized, which captures both longitudinal and lateral dynamic characteristics, as illustrated in Fig. \ref{fig:3_DOF_dynamic_model}.
By leveraging fundamental physical laws and kinematic constraints, the associated differential equations are derived as
\begin{equation}
\label{dynamic_model_physical_laws}
\begin{split}
    M(\dot{v}^\text{lo} - v^\text{la}\dot{\theta}) &= F^{\text{lo}}_f\cos{\delta} - F^{\text{la}}_{f}\sin{\delta} + F^{\text{lo}}_r \\
    M(\dot{v}^\text{la} + v^\text{lo}\dot{\theta}) &= F^{\text{lo}}_f\sin{\delta} + F^{\text{la}}_{f}\cos{\delta} + F^{\text{la}}_r \\
    I_z\ddot{\theta} &= (F^{\text{lo}}_{f}\sin{\delta} + F^{\text{la}}_{f}\cos\delta)l_f - F^{\text{la}}_{r}l_r
\end{split}
\end{equation}
where $v^\text{lo}$ and $v^\text{la}$ denote the longitudinal and lateral velocities in the body-fixed coordinates respectively; $F^{\text{lo}}_{f}$ and $F^{\text{lo}}_{r}$ represent the longitudinal tire forces acting on the front and rear tires respectively; $F^{\text{la}}_{f}$ and $F^{\text{la}}_{r}$ denote the lateral tire forces for the front and rear tires respectively; $\theta$ is the yaw angle; $M$ and $I_z$ are the vehicle mass and yaw moment of inertia, and $\delta$ denotes the front-wheel steering angle. In this paper, we limit $\delta$ to $[-7^\circ,7^\circ]$.

Under the small slip-angle assumption, the lateral tire forces can be approximated as linear functions of the slip angles, expressed as
\begin{equation}
\label{lateral_tire_forces_approximation}
\begin{split}
    F^{\text{la}}_{f} &= -C_f\alpha_f \\
    F^{\text{la}}_{r} &= -C_r\alpha_r \\
    \alpha_f &= (v^\text{la} + l_f\dot{\theta})/v^\text{lo} - \delta \\
    \alpha_r &= (v^\text{la} - l_r\dot{\theta})/v^\text{lo}.
\end{split}
\end{equation}
where $C_f$ and $C_r$ denote the cornering stiffness coefficients of the front and rear tires; $\alpha_f$ and $\alpha_r$ are the slip angles of the front and rear tires. Substituting \eqref{lateral_tire_forces_approximation} into \eqref{dynamic_model_physical_laws} results in the discrete-time 3-DOF vehicle dynamic model:
\begin{equation}
\label{eq:3_DOF_dynamics_without_state}
\begin{split}
    x^c_{k+1} &= x^c_k + (v^\text{lo}_k \cos{\theta_k} - v^\text{la}_k \sin{\theta_k}) \tau \\
    y^c_{k+1} &= y^c_k + (v^\text{lo}_k \sin{\theta_k} + v^\text{la}_k \cos{\theta_k}) \tau \\
    v^\text{lo}_{k+1} &= v^\text{lo}_k + (v^\text{la}_k\dot{\theta}_k + a^\text{lo}_k \cos \delta_k + \frac{F^\Delta_k}{M} \sin \delta_k) \tau \\
    v^\text{la}_{k+1} &= v^\text{la}_k - (v^\text{lo}_k\dot{\theta}_k - a^\text{lo}_k \sin \delta_k + \frac{F^\Delta_k}{M} \cos \delta_k  + \frac{\bar{F}^\Delta_k}{M}) \tau \\
    \theta_{k+1} &= \theta_k + \dot{\theta}_k \tau \\
    \dot{\theta}_{k+1} &= \dot{\theta}_k + (\frac{M l_f}{I_z} a^\text{lo}_k \sin \delta_k - \frac{l_f F^\Delta_k}{I_z} \cos \delta_k + \frac{l_r\bar{F}^\Delta_k}{I_z}) \tau
\end{split}
\end{equation}
where the subscript $k$ denotes the variable at the $k$-th time step; $a^\text{lo}_k$ is the longitudinal acceleration, and $F^\Delta_k = C_f(v^\text{la}_k - v^\text{lo}_k \delta_k + l_f \dot{\theta}_k)/v^\text{lo}_k, \ \bar{F}^\Delta_k = C_r(v^\text{la}_k - l_r \dot{\theta}_k)/v^\text{lo}_k$; $x^c_k$ and $y^c_k$ are the position of the center of mass (COM) of the ego vehicle in the world coordinates; $\tau$ is the step time for the discrete-time system. In this paper, we define $s_k \in \bar{\mathcal{S}} \subset \mathbb{R}^6$ as the vehicle state and $s_k = \begin{bmatrix} x^c_k & y^c_k & v^\text{lo}_k & v^\text{la}_k & \theta_k & \dot{\theta}_k \end{bmatrix}^\top$. The discrete-time nonlinear vehicle dynamics derived from \eqref{eq:3_DOF_dynamics_without_state} is written as:
\begin{equation}
\label{eq:dynamic_model_simplified}
s_{k+1} = f(s_k, u_k)
\end{equation}
where $u_k \in \mathcal{U} \subset \mathbb{R}^2$ is the control input, and is defined as $u_k = 
\begin{bmatrix} 
a^\text{lo}_k & \delta_k 
\end{bmatrix}^\top$.

Similar to $s_k$, we denote the state of the $i$-th obstacle as $s^{o,i}_k \in \bar{\mathcal{S}}$, where $i \in \{ 1,2\dots,N_o \}$ and $N_o$ is the number of obstacles. Then we define the environment state as 
\begin{equation}
\label{eq:environment_state_definition}
S_k = \begin{bmatrix} s_k & s^{o,1}_k & s^{o,2}_k & \dots & s^{o,N_o}_k\end{bmatrix}.
\end{equation}
The state space set of $S_k$ is $\mathcal{S} \subset \mathbb{R}^{6\times(N_o+1)}$.

Considering the tracking objective, the state error between the ego vehicle and the reference trajectory is defined as
\begin{equation}
\label{eq:error_state_definition_goal}
e_k = 
\begin{bmatrix}
x^c_k - \tilde{x}^c_k & 
y^c_k - \tilde{y}^c_k &
v^\text{lo}_k - \tilde{v}^\text{lo}_k &
\theta_k - \tilde{\theta}_k
\end{bmatrix}^\top
\end{equation}
where 
$\tilde{x}^c_k$ and $ \tilde{y}^c_k, \tilde{v}^\text{lo}_k, \ \tilde{\theta}_k$ are respectively the position, the longitudinal speed profile, and the yaw angle of the desired state at the $k$-th time step. The state space set of $e_k$ is $\mathcal{S}_e \subset \mathbb{R}^4$.

\subsection{Problem Formulation}
Considering the vehicle system \eqref{eq:dynamic_model_simplified}, the objective of STT is to minimize the cumulative error between ego states and desired states in the reference trajectory while maintaining safety. Assume the reference trajectory contains $T+1$ desired states and $T \in \mathbb{N}_{+}$. Let us denote $H_k(S_k)$ as the safety factor and $H_k(S_k) \leq 0 $ as the safety constraint, indicating that the vehicle maneuvers inside the road boundaries, avoids obstacles and meets the state boundary requirements. Then, $\forall k \in \{0,1,\dots,T-1\}$, the STT can be mathematically formulated as the following optimization problem:
\begin{equation}
    \label{eq:safe_trajectory_tracking_problem_formulation}
    \begin{split}
    \min_{U} \quad & \sum_{i=k}^{T} \sqrt{e_i^\top W e_i}, \\
    \text{s.t.} \quad & \max_{k \leq i \leq T} H_i(S_i) \leq 0, \\
    & s_{k+1} = f(s_k, u_k)
    \end{split}
\end{equation}
where $U=\{ u_k, u_{k+1}, \dots, u_{T-1}\}$ is the set of the control inputs that need to be optimized; $W$ is a symmetric positive-definite weight matrix; $s_{k+1} = f(s_k, u_k)$ represents the vehicle dynamics in \eqref{eq:dynamic_model_simplified}. For conciseness, we omit the argument $S_k$ of $H_k(\cdot)$ and define $l_k = \sqrt{e_k^\top W e_k}$ to represent the tracking cost for each step in the following.

\section{Methodology}
\label{sec:methodology}
In this section, we first introduce the concept of epigraph to reformulate the STT task as a min-max optimization problem, including both tracking and safety performance. Subsequently, an EFRL algorithm is proposed to solve such an optimization problem and derive a corresponding optimal controller. Following this, the stability of the closed-loop system is analyzed under unconstrained conditions. Finally, a specific scaling factor is designed for CAVs along with a computationally efficient algorithm, ensuring precise safety constraint representations and fast implementation in training.

\subsection{Epigraph Form Reinforcement Learning}
\label{subsec:EFRL}
\begin{figure*}[htbp]
\centering
\includegraphics[width=0.9\linewidth]{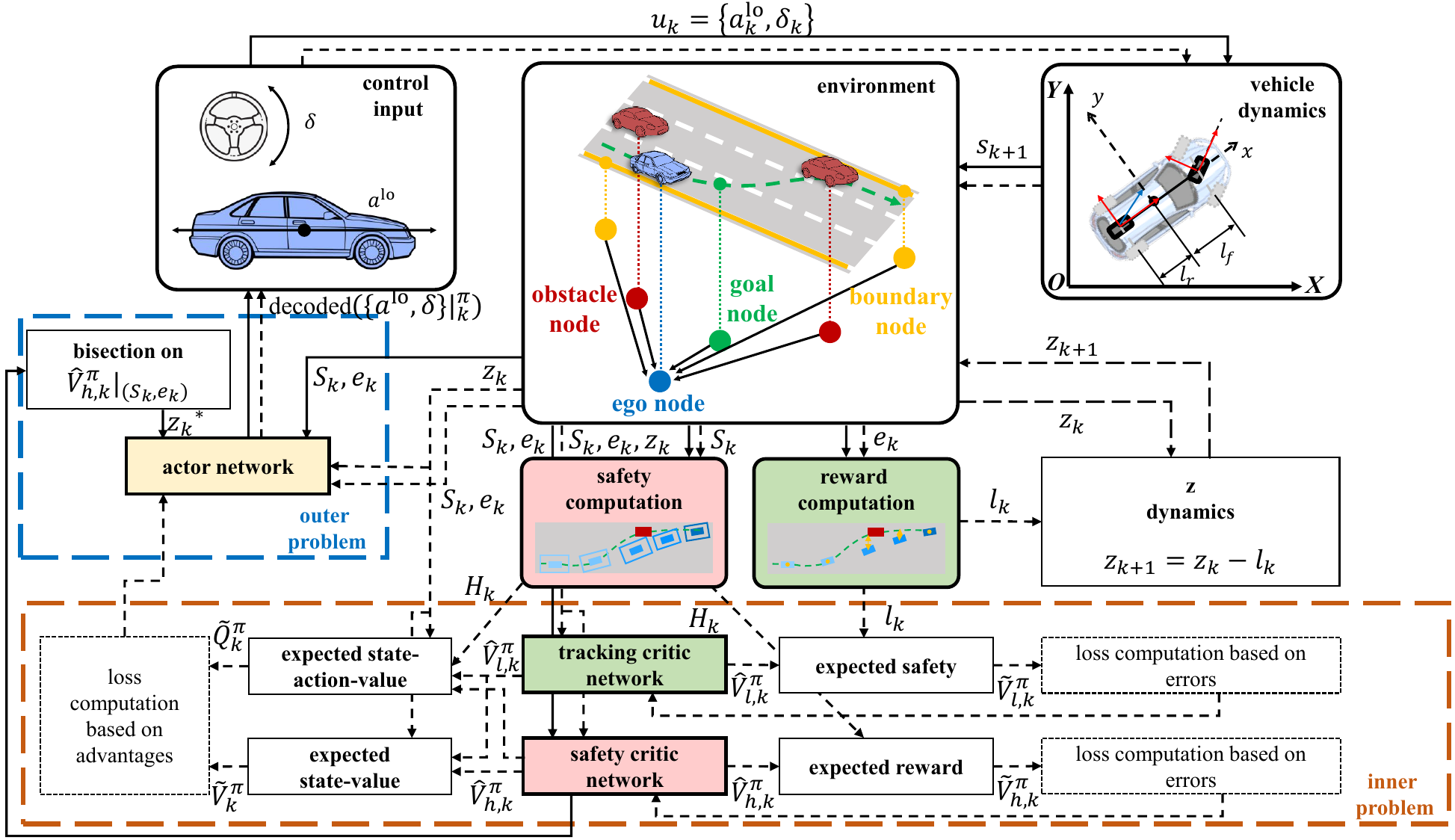}
\caption{STTC-EFRL framework architecture. The solid lines denote the information flow during evaluation and implementation, and the dashed lines denote the information flow during training.}
\label{fig:framework_architecture}
\end{figure*}
To solve the problem \eqref{eq:safe_trajectory_tracking_problem_formulation}, one common method is to redesign the objective function as a weighted form that incorporates both tracking and safety terms. However, this method implicitly treats safety constraints as soft constraints in the optimization problem formulation. Then it is critical but hard to decide the proper weights that balance the training stability and safety violations. Another widely-used approach is to reformulate the optimization problem into a Lagrangian form based on the Karush-Kuhn-Tucker (KKT) conditions. This method is hard to solve, especially in high dimensional nonlinear systems, even using DRL-based approaches as discussed in Remark \ref{rmk:RL_of_Lagrangian}.

Motivated by \cite{altarovici2013general, so23solving}, we introduce a scalar epigraph variable and formulate the corresponding epigraph form representation of problem \eqref{eq:safe_trajectory_tracking_problem_formulation}. $\forall k \in \{0,1,\dots,T-1\}$,
\begin{equation}
\label{eq:constrained_OCP_EF}
\begin{split}
\min_{z_k} \quad & z_k, \\
\text{s.t.} \quad & \min_{\pi} \max \{V^{\pi}_{h,k}, V^{\pi}_{l,k}-z_k\} \leq 0, \\
&  s_{k+1}=f(s_k, u^{\pi}_k)
\end{split}
\end{equation}
where $V^{\pi}_{h,k} = \max_{k \leq i \leq T} H_i$ denotes the system safety in the optimization horizon, and $V^{\pi}_{l,k} = \sum_{i=k}^{T}l_i$ is the total tracking cost; $z_k = \sup\{V^{\pi}_{l,k}\}$ denotes the epigraph variable of the total tracking cost. The min-max logical formulation combined with the epigraph thus imposes more stringent safety constraint representations.

Solving \eqref{eq:constrained_OCP_EF} directly by numerical optimization solvers suffers from local optimality and high computational cost, especially considering highly dynamic obstacles, high-speed, and safety requirements. Therefore, we adopt a DRL-based solving approach that exhibits rapid environmental responsiveness and superior optimality exploration, indicating greater potential for practical implementations. The optimization over the control inputs $U$ is transferred to the optimization of a policy $\pi: \mathcal{S} \times \mathcal{S}_e \times \mathbb{R}_{\geq 0} \to \mathcal{U}$ that takes $ S_k, \ e_k$ and $z_k$ as inputs and $u_k$ as output. Through the min-max epigraph form reformulation in \eqref{eq:constrained_OCP_EF}, the original constrained problem \eqref{eq:safe_trajectory_tracking_problem_formulation} can be decomposed into two interconnected subproblems: an unconstrained inner one over $\pi$ and a constrained outer one over $z_k$.

\begin{remark}
\label{rmk:RL_of_Lagrangian}
Of course, the DRL-based approaches can be used in Lagrangian optimization form \cite{gu2024review}. The Lagrangian optimization inner problem is usually formulated as $\max_{\lambda \geq 0} \ \min_{\pi} \ V^{\pi}_{l,k} + \lambda \max \{ V^{\pi}_{h,k} , 0 \}$. Then the gradient of the inner problem w.r.t. $\lambda$ is a non-negative value at all times. For this reason, $\lambda$ will rise continuously unless all constraints are satisfied. Meanwhile, an extremely large $\lambda$ occurs when constraints are unfulfilled, as the policy gradient displays a linear scaling relationship with $\lambda$. Consequently, the Lagrangian DRL-based approaches usually suffer from training instability.
\end{remark}

The inner problem can be given by: $\forall k \in \{0,1,\dots,T-1\}$,
\begin{subequations}
    \label{eq:epigraph_inner_problem}
    \begin{align}
    \min_{\pi} \quad & \max \{V^{\pi}_{h,k}, V^{\pi}_{l,k}-z_k\},
    \label{eq:epigraph_inner_problem:objective} \\
    \text{s.t.} \quad & 
    s_{k+1}=f(s_k, u^{\pi}_k),
    \label{eq:epigraph_inner_problem:x_dynamics} \\
    & z_{k+1} = z_k - l_k.
    \label{eq:epigraph_inner_problem:z_dynamics}
    \end{align}
\end{subequations}
It is formulated as an unconstrained OCP that can be smoothly solved via reinforcement learning. Moreover, $z_0 \in [0, z_{\max}]$, where $z_{\max}$ should be the ideal value of $\sup\{V^{\pi^*}_{l,k}\}$ corresponding to the optimal solution $\pi^*$.

Additionally, the outer problem can be expressed as
\begin{subequations}
\label{eq:epigraph_outer_problem}
\begin{align}
\min_{z_k} \quad & z_k,
\label{eq:epigraph_outer_problem:objective} \\
\text{s.t.} \quad & \min_{\pi} V^{\pi}_k \leq 0
\label{eq:epigraph_outer_problem:constraints}
\end{align}
\end{subequations}
where $V^{\pi}_k = \max \{V^{\pi}_{h,k}, V^{\pi}_{l,k}-z_k\}$ aligning with \eqref{eq:epigraph_inner_problem:objective}. The objective is to find the optimal epigraph variable ${z_k}^*$ for given $S_k$ and $e_k$.

Therefore, the solving process for \eqref{eq:constrained_OCP_EF} is divided into two stages, whose structure is illustrated in Fig.\ref{fig:framework_architecture}. At Stage 1, $\pi^*$ is computed by actor-critic style DRL. 
Two separate Neural Networks (NNs) are introduced as the critics to approximate one state-value $V^{\pi}_{l,k}$ w.r.t. the tracking cost and another state-value $V^{\pi}_{h,k}$ w.r.t. the safety. Here, $\pi^*$ is conditioned on $z_k$. Denoting the theoretical optimal solution of the outer problem as $z_k^*$, the optimal action can be written as
\begin{equation}
\label{eq:optimal_action}
{u_k}^* = \pi^*(S_k, e_k, {z_k}^*).
\end{equation}
In the practical implementation of the outer problem, obtaining the theoretical optimal solution $z_k^*$ is non-trivial. Therefore, we employ a bisection-based root-finding procedure using the learned safety critic NN to obtain a numerical estimate $\hat z_k$ of the epigraph variable. Further details are provided in Remarks \ref{rmk:explain_dynamics_zk} and \ref{rmk:robust_bisection}. Accordingly, the control action during inference is given by
\begin{equation}
\label{eq:inference_action}
u_k = \pi^*(S_k,e_k,\hat{z}_k).
\end{equation}

Then, we focus on explaining how to solve the inner problem using an actor-critic-style DRL method. We define $\langle a,b \rangle = \max\{ a,b \}$ for conciseness. In the training process, each sampled trajectory is rolled out for $T$ steps. For such training set-ups, the action-state-value of the tracking cost $Q^\pi_{l,k}$ and the safety $Q^\pi_{h,k}$ can be expanded as follows:
\begin{equation}
\label{eq:critic_expansion}
\begin{split}
Q^{\pi}_{l,k} &= l_k + \mathbf{E}^{u,s}_{k+1}[\dots + \mathbf{E}^{u,s}_{j-1}[l_{j-1} +\mathbf{E}^s_{j}[V^{\pi}_{l,j}]]], \\
Q^{\pi}_{h,k} &= \langle H_k, \mathbf{E}^{u,s}_{k+1}[ \langle \dots, \mathbf{E}^{u,s}_{j-1}[ \langle H_{j-1},\mathbf{E}^s_{j}[V^{\pi}_{h,j}] \rangle ] \rangle ] \rangle  
\end{split}
\end{equation}
where $j \in \{1,2,\dots,T\}$ and $k \in \{0,1,\dots,j-1\}$; $\mathbf{E}^{u}_k \coloneqq \mathbf{E}_{u_k \sim \pi(S_k,e_k,z_k)}, \ \mathbf{E}^s_{k+1} \coloneqq \mathbf{E}_{S_{k+1} \sim S_k \mid u_k}$, and $\mathbf{E}^{u,s}_{k} \coloneqq \mathbf{E}^u_k \circ \mathbf{E}^s_{k} $. During training, the mathematical expectations appearing in \eqref{eq:critic_expansion} are replaced by estimations obtained from sampled trajectories.

\begin{algorithm}[htbp]
\caption{STTC-EFRL: Inner Problem (Training)}
\label{alg:STTC_EFRL_inner}
\begin{algorithmic}
\STATE Construct training environment, initialize the actor NN, the tracking and safety critic NNs.
\STATE \textbf{while} training \textbf{do}:
\STATE \hspace{0.5cm} Initialize the ego state $s_0$, the epigraph $z_{0}$, the reference trajectory and obstacles. Roll out a sampled trajectory under the policy $\pi$ for $T$ steps.
\STATE \hspace{0.5cm} \textbf{for} $k$ from $0$ to $T-1$ \textbf{do}:
\STATE \hspace{1cm} $\hat{V}^{\pi}_{l,k} \gets$ tracking critic NN output.
\STATE \hspace{1cm} $\hat{V}^{\pi}_{h,k} \gets$ safety critic NN output.
\STATE \hspace{1cm} $p^{\pi}_{k} \gets$ action distribution under $\pi$.
\STATE \hspace{1cm} $u_{k} \gets$ action sampled from $p^{\pi}_{k}$.
\STATE \hspace{1cm} $s_{k+1} \gets$ next ego state under system dynamics \eqref{eq:epigraph_inner_problem:x_dynamics}.
\STATE \hspace{1cm} $l_{k} \gets$ tracking cost.
\STATE \hspace{1cm} $z_{k+1} \gets$ next epigraph under z-dynamics \eqref{eq:epigraph_inner_problem:z_dynamics}.
\STATE \hspace{1cm} $H_{k} \gets$ safety factor.
\STATE \hspace{0.5cm} \textbf{end for}
\STATE \hspace{0.5cm} $\mathscr{L}_l \gets$ tracking critic NN loss calculation using \eqref{eq:critic_loss_function:tracking_cost}.
\STATE \hspace{0.5cm} $\mathscr{L}_h \gets$ safety critic NN loss calculation using \eqref{eq:critic_loss_function:safety_factor}.
\STATE \hspace{0.5cm} Update critic NNs with critic gradients.
\STATE \hspace{0.5cm} \textbf{while} in policy update epochs \textbf{do}:
\STATE \hspace{1cm} $p^{\pi_\text{old}}_{k} \gets$ action distribution under old policy $\pi_\text{old}$.
\STATE \hspace{1cm} $\mathscr{L}_{\pi} \gets$ policy loss calculation using \eqref{eq:policy_loss}.
\STATE \hspace{1cm} Update $\pi$ with policy gradient.
\STATE \hspace{0.5cm} \textbf{end while}
\STATE \textbf{end while}
\end{algorithmic}
\end{algorithm}

\begin{algorithm}[htbp]
\caption{STTC-EFRL: Outer Problem (Evaluation)}
\label{alg:STTC_EFRL_outer}
\begin{algorithmic}
\STATE Construct evaluation environment, initialize ego state $s_0$, reference trajectory and obstacles.
\STATE \textbf{for} $k$ from $0$ to $T-1$ \textbf{do}:
\STATE \hspace{0.5cm} $\hat{z}_k \gets$ bisection-based root-finding with safety margin.
\STATE \hspace{0.5cm} $p^{\pi}_k \gets$ action distribution under $\pi$.
\STATE \hspace{0.5cm} $u_k \gets$ action with the highest probability density.
\STATE \hspace{0.5cm} $s_{k+1} \gets$ next ego state.
\STATE \hspace{0.5cm} $l_k \gets$ tracking cost.
\STATE \hspace{0.5cm} $H_k \gets$ safety factor.
\STATE \textbf{end for}
\STATE Compute terminal tracking cost $l_T$ and safety factor $H_T$.
\STATE \textbf{return:}
\STATE \hspace{0.5cm} $\sum_{i=0}^{T} l_i$ to evaluate total tracking cost.
\STATE \hspace{0.5cm} $\max_{0 \leq i \leq T}H_i$ to evaluate safety.
\end{algorithmic}
\end{algorithm}

Consequently, the losses of the critic NNs can be given by
\begin{subequations}
\label{eq:critic_loss_function}
\begin{align}
\mathscr{L}_l &= \frac{1}{T}\sum_{k=0}^{T-1} \| \tilde{Q}^{\pi}_{l,k} - \hat{V}^{\pi}_{l,k} \|, 
\label{eq:critic_loss_function:tracking_cost} \\
\mathscr{L}_h &= \frac{1}{T}\sum_{k=0}^{T-1} \| \tilde{Q}^{\pi}_{h,k} - \hat{V}^{\pi}_{h,k} \|
\label{eq:critic_loss_function:safety_factor}
\end{align}
\end{subequations}
where $\hat{V}^{\pi}_{l,k}$ and $\hat{V}^{\pi}_{h,k}$ are the outputs of the tracking critic NN and the safety critic NN at the $k$-th step, respectively; $\tilde{Q}^{\pi}_{l,k}$ and $\tilde{Q}^{\pi}_{h,k}$ are the estimation of $Q^{\pi}_{l,k}$ and $Q^{\pi}_{h,k}$, respectively. The estimations $\tilde{Q}^{\pi}_{l,k}$ and $\tilde{Q}^{\pi}_{h,k}$ can be expanded to the temporal difference style according to \eqref{eq:critic_expansion}. We define:
\begin{equation}
\label{eq:critic_time_difference_expansion}
\begin{split}
\tilde{Q}^l_j &\coloneqq l_k + l_{k+1} + \dots + l_{j-1} + \hat{V}^{\pi}_{l,j}, \\
\tilde{Q}^h_j &\coloneqq \langle H_k, \langle H_{k+1}, \dots, \langle H_{j-1}, \hat{V}^{\pi}_{h,j} \rangle \rangle \rangle.
\end{split}
\end{equation}

Then, the $\lambda$-Temporal Difference ($\lambda$-TD) algorithm is used to derive the weighted sum of action-state-value estimations, which can be described as
\begin{equation}
\label{eq:critic_time_difference_sum}
\begin{split}
\tilde{Q}^{\pi}_{l,k} =& (1-\lambda_Q)(\sum_{i=k+1}^{T-1}\lambda_Q^{i-k-1}\tilde{Q}^l_i + \frac{\lambda_Q^{T-k-1}}{1-\lambda_Q}\tilde{Q}^l_T) \\
\tilde{Q}^{\pi}_{h,k} =& (1-\lambda_Q)(\sum_{i=k+1}^{T-1}\lambda_Q^{i-k-1}\tilde{Q}^h_i + \frac{\lambda_Q^{T-k-1}}{1-\lambda_Q}\tilde{Q}^h_T)
\end{split}
\end{equation}
where $\lambda_Q \in (0,1)$ is the discount factor for $\lambda$-TD.

Since the $\max$ operator does not satisfy the linearity of expectation, we define the Bellman equation for RL for the inner problem as
\begin{equation}
\label{eq:bellman_equation}
V^{\pi}_k = \mathbf{E}^u_k[\langle H_k, \mathbf{E}^s_{k+1}[V^{\pi}_{k+1}] \rangle ]
\end{equation}
where $V^\pi_k$ denotes the overall state-value balanced by $z_k$ and follows the definition in \eqref{eq:epigraph_outer_problem:constraints}. The estimation of $V^\pi_k$ is defined as $\tilde{V}^{\pi}_k \coloneqq \langle \hat{V}^{\pi}_{h,k}, \hat{V}^{\pi}_{l,k} - z_k \rangle$.

Additionally, the overall action-state-value balanced by $z_k$ is denoted as $Q^\pi_k$ and can be expanded as
\begin{equation}
\label{eq:action_value_expansion}
Q^{\pi}_k = \langle H_k, \mathbf{E}^{u,s}_{k+1} [ \langle \dots, \mathbf{E}^{u,s}_{j-1}[ \langle H_{j-1}, \mathbf{E}^s_{j}[V^{\pi}_{j}] \rangle ] \rangle ] \rangle.
\end{equation}
Similarly, by replacing expectations with estimations, we define:
\begin{equation}
\label{eq:actor_time_difference_expansion}
\tilde{Q}_j \coloneqq \langle H_k, \langle H_{k+1}, \dots, \langle H_{j-1}, \tilde{V}^\pi_j \rangle \rangle \rangle.
\end{equation}
Therefore, we can derive the overall action-state-value estimation using $\lambda$-TD:
\begin{equation}
\label{eq:actor_time_difference_sum}
\tilde{Q}^{\pi}_k = (1-\lambda_Q)(\sum_{i=k+1}^{T-1}\lambda^{i-k-1}_Q \tilde{Q}_i + \frac{\lambda^{T-k-1}_Q}{1-\lambda_Q} \tilde{Q}_T).
\end{equation}

Finally, to update actor NN, Generalized Advantage Estimation (GAE) \cite{schulman2015high} is introduced and can be described as
\begin{equation}
\label{eq:advantage_calculation_GAE}
A^{\pi}_k = \tilde{Q}^{\pi}_k - \tilde{V}^{\pi}_k
\end{equation}
where $A^{\pi}_k$ denotes the advantage. To this end, the policy loss can be calculated using the Proximal Policy Optimization (PPO) algorithm \cite{schulman2017proximal}:
\begin{equation}
\label{eq:policy_loss}
\begin{split}
\mathscr{L}_{\pi} &= -\frac{1}{T} \sum_{k=0}^{T-1} [ \min \{ J_{k}A^{\pi}_{k}, \bar{J}_{k}
A^{\pi}_{k} \} ], \\
\bar{J}_{k} &= \operatorname{clip}(J_{k}, 1-\epsilon_p, 1+\epsilon_p)
\end{split}
\end{equation}
where $J_{k} = {p^{\pi}_{k}} / {p^{\pi_{\text{old}}}_{k}}$ indicates the action probability density ratio between the updated policy $\pi(S_k, e_k, z_{k})$ and old policy $\pi_{\text{old}}(S_k, e_k, z_{k})$; $\epsilon_p \in (0,1)$ is the clip coefficient for PPO.

The solving processes of the inner and outer problems are summarized in the pseudocode of Algorithm \ref{alg:STTC_EFRL_inner} and Algorithm \ref{alg:STTC_EFRL_outer}. In what follows, we present a stability analysis of the closed-loop system under the trajectory tracking controller. 

In classical control theory analysis on constrained Lyapunov stability, to handle the state constraints, an intermediate feasibility set $\Phi$ is first defined as the intersection of two sets: the set where the control inputs satisfy physical limits while enforcing Lyapunov function decay, and the set defining the state constraints. However, as the system's trajectory might temporarily violate state constraints before stabilizing, a more strictly safe invariant convex subset $\Omega$ is then constructed by identifying the largest closed level set of the Lyapunov function completely contained within $\Phi$. If the initial condition of the system is within $\Omega$, then the bounded controller ensures the state trajectory continuously moves toward lower energy levels without crossing the boundaries, rendering the system safe and asymptotically stable \cite{mhaskar2006stabilization}.

However, this theoretical proof requires the guaranteed existence of such a static convex set $\Omega$ that contains the equilibrium point. In the STT problem, the presence of multiple obstacles makes the safe region non-convex and time-varying directly. Furthermore, in highly dynamic STT scenarios where obstacles could appear directly on the reference trajectory, the existence of $\Omega$ can no longer be guaranteed. Therefore, rigorous Lyapunov stability analysis for STT problems under highly dynamic and non-convex safety regions remains an open research topic.

Despite these limitations under active constraints, establishing Lyapunov stability under unconstrained conditions remains theoretically and practically essential. First, unconstrained asymptotic stability guarantees the system's fundamental capability to track the reference trajectory and gracefully recover after obstacle avoidance. Second, although active safety constraints inevitably perturb nominal stability near boundaries, the scenario dynamically reverts to an unconstrained tracking task once obstacles are bypassed. Thus, the unconstrained stability analysis accurately reflects the core system mechanism when operating outside the active influence of safety constraints. 

While acknowledging the theoretical limitation in the fully constrained setting, we empirically validate the practical safety and stability of our STTC-EFRL approach through comprehensive ViL experiments, with detailed results and observations presented in Section \ref{sec:experiments}.

\begin{theorem}
\label{thm:global_asymptotic_stable}
Consider problem \eqref{eq:safe_trajectory_tracking_problem_formulation} without safety constraints. Define the infinite-horizon policy value function as $V^{\pi,\infty}_{l,k} \coloneqq \sum_{i=k}^{\infty} l_i$, and the domain where the value function is finite as $\mathcal{V} \coloneqq \{ e_k \mid V^{\pi,\infty}_{l,k} < +\infty \}$. Let $\mathcal{K}_\infty$ denote the infinite class $\mathcal{K}$ functions \cite{ames2016control}.
Suppose for all \(e_k \in \mathcal{V}\):
\begin{enumerate}
\item There exists \(\beta \in \mathcal{K}_\infty\) such that
\begin{equation}
\label{V_is_lyapunov_assumption1}
V^{\pi,\infty}_{l,k} \leq \beta(l_k).
\end{equation}
\item There exists \(\bar{\beta} \in \mathcal{K}_\infty\) such that
\begin{equation}
\label{V_is_lyapunov_assumption2}
l_k \geq \bar{\beta}(l_k).
\end{equation}
\end{enumerate}
Then the closed-loop system of \eqref{eq:dynamic_model_simplified} under policy $\pi$
is globally asymptotically stable with respect to the equilibrium $e_k = 0$.
\end{theorem}

\begin{figure}[tbp]
\centering
\subfloat[]{
\includegraphics[width=0.29\linewidth]{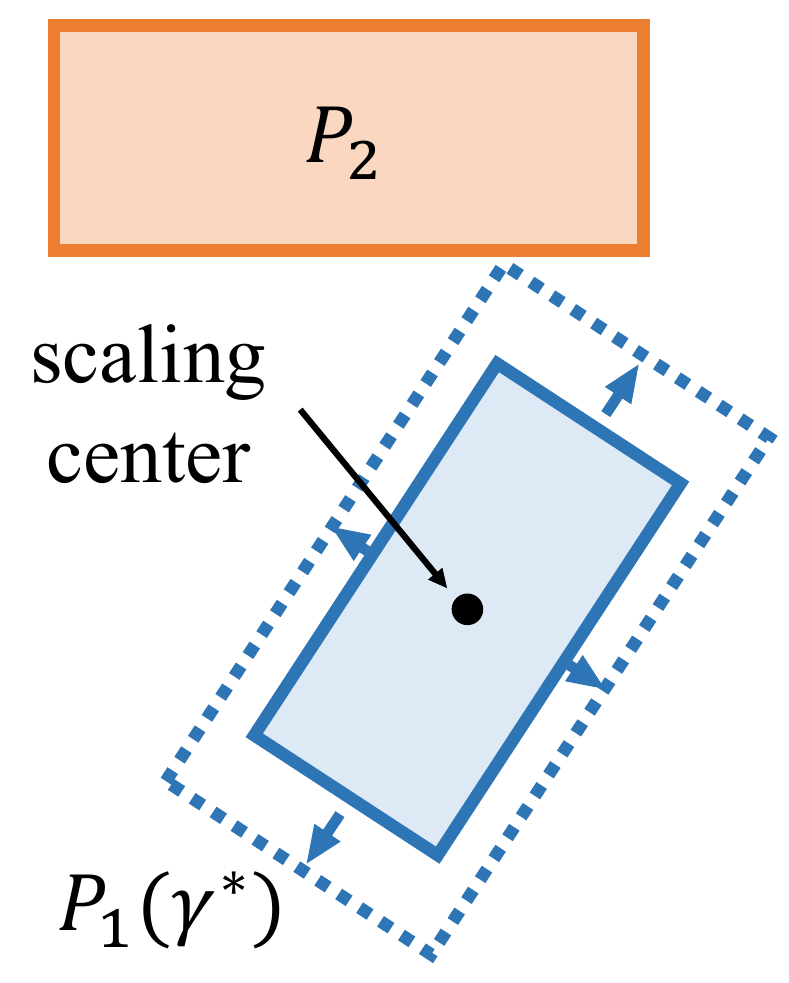}
\label{fig:gamma_star_greater_than_1}
}
\hfill
\subfloat[]{%
\includegraphics[width=0.29\linewidth]{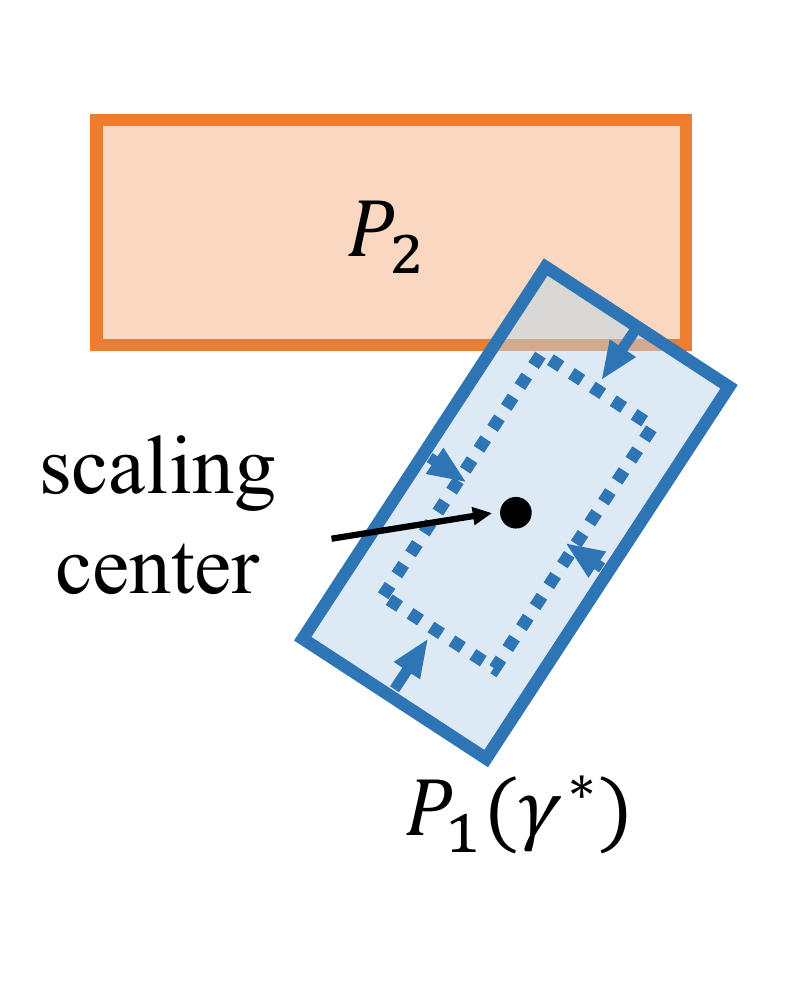}
\label{fig:gamma_star_less_than_1}
}
\hfill
\subfloat[]{%
\includegraphics[width=0.29\linewidth]{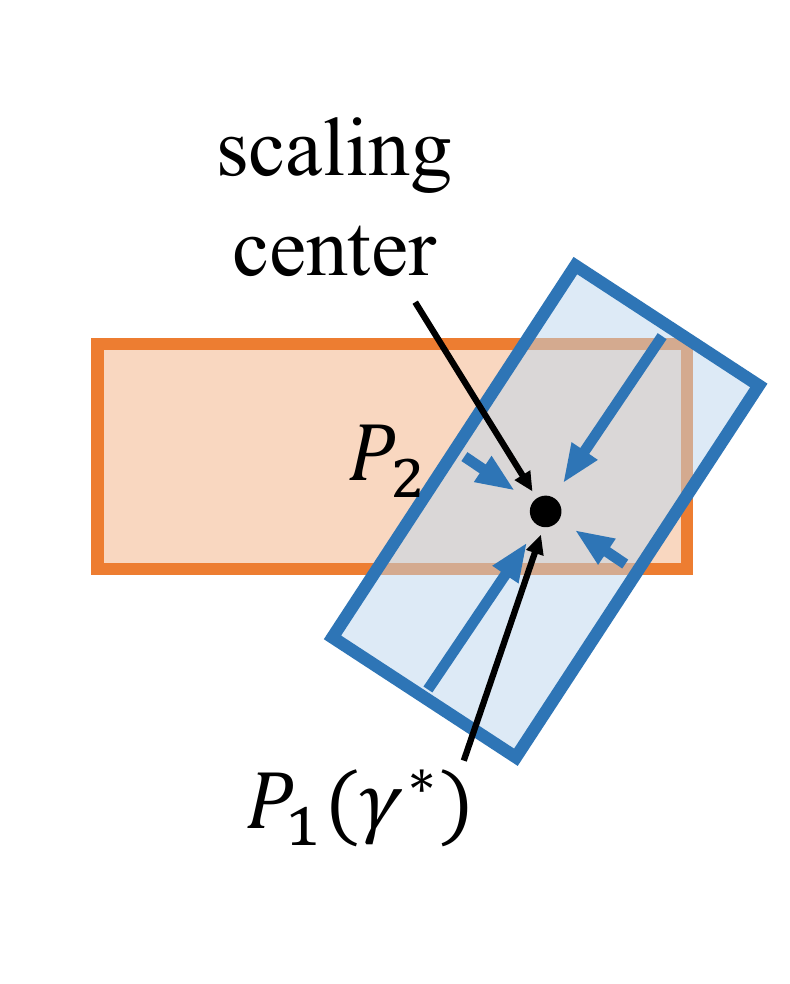}
\label{fig:gamma_star_equal_to_0}
}
\caption{Examples of calculating the optimal scaling factor $\gamma^*$ between two rectangles. In (a), $\gamma^* > 1$, which means no collision occurs between the two rectangles. In (b), $0 < \gamma^* \leq 1$, which means $P_1$ collides with $P_2$, where a smaller $\gamma^*$ indicates a more severe collision. In (c),  $\gamma^* = 0$, which means the scaling center of $P_1$ lies inside $P_2$, and $\gamma^*$ is consistently 0.}
\label{fig:scaling_factor}
\end{figure}

\begin{proof}
First, we analyze $\mathcal{V}$ and organize the proof. Since $l_k \geq 0$, we have $\forall e_k \in \mathcal{V}, \ \sum_{i=k}^{\infty} l_i$ converges. A necessary condition for the convergence of a non-negative infinite series is $\lim_{k \to \infty} l_k = 0$. Since $l_k$ is a positive-definite weighted norm with respect to $e_k$, this further implies $\lim_{k \to \infty} e_k = 0$. However, it only means the trajectory converges to the equilibrium $e_k=0$. Global asymptotic stability requires both Lyapunov stability and convergence to the equilibrium. Thus, we must prove $V^{\pi,\infty}_{l,k}$ is a Lyapunov function.

Next, we verify the rationality of Assumptions \eqref{V_is_lyapunov_assumption1} and \eqref{V_is_lyapunov_assumption2} for the STT task. Assumption \eqref{V_is_lyapunov_assumption2} holds trivially because the tracking cost $l_k$ is a positive-definite weighted norm by design, so we can choose an appropriate $\bar{\beta} \in \mathcal{K}_\infty$ to satisfy $l_k \geq \bar{\beta}(l_k)$. For Assumption \eqref{V_is_lyapunov_assumption1}, the reference trajectory generated by the high-level global planner is continuous and bounded to ensure ride comfort and feasibility. This boundedness limits the growth of the cumulative tracking cost, so $V^{\pi,\infty}_{l,k}$ can be bounded by a $\mathcal{K}_\infty$ function of the current cost $l_k$. Both $\beta$ and $\bar{\beta}$ can be explicitly determined from the weight matrix $W$ and the bounds on trajectory variations.

With the above clarifications, we proceed to prove $V^{\pi,\infty}_{l,k}$ is a Lyapunov function. By non-negativity of $l_k$, the value function satisfies the dynamic programming principle:
\begin{equation}
\label{dynamic_programming_for_v_at_k}
V^{\pi,\infty}_{l,k} = l_k + V^{\pi,\infty}_{l,k+1}.
\end{equation}
Using Assumption \eqref{V_is_lyapunov_assumption2} and non-negativity of $V^{\pi,\infty}_{l,k+1}$, we derive the lower bound of $V^{\pi,\infty}_{l,k}$:
\begin{equation}
\label{satisfy_lyapunov1}
V^{\pi,\infty}_{l,k}
\geq \bar{\beta}(l_k) + V^{\pi,\infty}_{l,k+1}
\geq \bar{\beta}(l_k).
\end{equation}
Rearranging \eqref{dynamic_programming_for_v_at_k} and substituting Assumption \eqref{V_is_lyapunov_assumption2}, we obtain the decay condition of $V^{\pi,\infty}_{l,k}$ along the trajectory:
\begin{equation}
\label{satisfy_lyapunov2}
V^{\pi,\infty}_{l,k+1}
= V^{\pi,\infty}_{l,k} - l_k
\leq V^{\pi,\infty}_{l,k} - \bar{\beta}(l_k).
\end{equation}
Combining \eqref{satisfy_lyapunov1}, \eqref{satisfy_lyapunov2} and Assumption \eqref{V_is_lyapunov_assumption1}, we have:
\begin{equation}
\label{V_satisfies_lyapunov_function}
\begin{split}
\bar{\beta}(l_k) \leq V^{\pi,\infty}_{l,k} &\leq \beta(l_k), \\
V^{\pi,\infty}_{l,k+1} \leq V^{\pi,\infty}_{l,k} &- \bar{\beta}(l_k).
\end{split}
\end{equation}
Then by Definition 2.18 in \cite{grune2016nonlinear}, $V^{\pi,\infty}_{l,k}$ is a Lyapunov function. According to Theorem 2.19 in \cite{grune2016nonlinear}, the existence of such a function implies global asymptotic stability for the closed-loop system of \eqref{eq:dynamic_model_simplified} under policy $\pi$ at $e_k = 0$.
\end{proof}

\begin{remark}
\label{rmk:explain_dynamics_zk}
The dynamics of $z_k$ is defined such that it decreases with the tracking cost $l_k$ as given in \eqref{eq:epigraph_inner_problem:z_dynamics}, since the global asymptotic stability of $\pi$ ensures that $V^\pi_{l,k}$ decreases with $k$. As $z_k$ represents the budget between stability and safety \cite{so23solving}, after training, the value of $z_k$ acts as a regulator between the tracking cost and safety for $\pi$. A smaller $z_k$ imposes a tighter tracking-cost budget, whereas a larger $z_k$ relaxes this budget and allows greater emphasis on safety. This regulation mechanism provides the motivation for numerically estimating the epigraph variable $\hat z_k$ during policy evaluation.

For a policy satisfying $V^\pi_{l,k}=z_k$, the feasibility condition in \eqref{eq:epigraph_outer_problem:constraints} reduces to $V^\pi_{h,k}\leq 0$. Accordingly, the safety value function provides the criterion for evaluating the safety constraint under this condition. Since this function is approximated by the learned safety critic NN, the epigraph variable can be numerically estimated by locating where the predicted safety value reaches the prescribed threshold. Therefore, the estimated epigraph variable $\hat{z}_k$ is determined by searching for a root of the learned safety critic NN within a prescribed interval. Although the learned safety critic NN is not explicitly constrained to be monotonic with respect to $z_k$, the bisection-based root-finding procedure relies on a bracketed sign change rather than global monotonicity of the safety critic. When no valid sign-changing interval is identified, a boundary value is instead returned based on the evaluated safety critic values. Specifically, the lower bound is selected when all evaluated values satisfy the root-finding criterion, whereas the upper bound is returned as a fallback when none of the evaluated values satisfies this criterion.
\end{remark}

\begin{remark}
\label{rmk:robust_bisection}
In practical implementations, the safety of the ego vehicle may be compromised by the learning error of the safety critic NN. Consequently, searching for a root of $\hat{V}^{\pi}_{h,k}=0$ may lead to an overly aggressive epigraph estimate.
To enhance robustness, we employ the bisection-based root-finding procedure on $\hat{V}^{\pi}_{h,k} = -\nu$, where $\nu$ is a small positive threshold parameter. Introducing $\nu$ provides an additional safety margin in the numerical estimation of $\hat{z}_k$, thereby encouraging more conservative control actions and improving robustness against safety critic approximation errors.
\end{remark}

\subsection{Safety Constraints based on Scaling Factor}
Circles and ellipses are extensively adopted to model the geometric shape of vehicles in planning and control tasks for CAVs \cite{han2024uniform, liu2025lane, funada2025collision}. However, such representations either lack sufficient accuracy for most vehicles, such as modeling vehicles as circles and using the minimum distance between the centers of the circles to interpret collision conditions, or suffer from heavy computational burden, such as modeling vehicles as ellipses and using the signed distance function \cite{singletary2022safety} or the supporting hyperplane \cite{funada2025collision} between the ellipses to interpret collision conditions. 

In this paper, given the inherent box-shaped geometric features of most vehicles, rectangles are employed to depict the geometric form of vehicles. Furthermore, the scaling factor is utilized to characterize the corresponding geometric relationships and safety situations. Inspired by \cite{tracy2022differentiable}, instead of scaling two primitives at once, we propose that scaling only the ego primitive that envelops the ego vehicle is more reasonable in the STT tasks for CAVs. This is because the counterpart primitive usually represents the obstacles, road boundaries or other vehicles. The shape of the counterpart primitive is usually irregular, sometimes even unbounded. Therefore, the scaling center of the counterpart primitive is hard to determine. In the STT, we use polyhedra as the geometric models of other obstacles and the road boundaries in the bird's-eye-view. Subsequently, we design the ego-scaled scaling factor $\gamma$ via the following optimization problem,
\begin{subequations}
\label{eq:scaling_optimization_linear_form}
\begin{align}
\min_{x,\gamma} \quad & \gamma, 
\label{eq:scaling_optimization_linear_form:objective} \\
\text{s.t.} \quad & A_1 Q_1^\top(x - O_1) \leq \gamma b_1,
\label{eq:scaling_optimization_linear_form:ego_primitive} \\
& A_2 x \leq b_2
\label{eq:scaling_optimization_linear_form:counterpart_primitive}
\end{align}
\end{subequations}
where $x \in \mathbb{R}^2$ denotes a point in the 2-D space; $\gamma \geq 0$ denotes the scaling factor; $A_1 \in \mathbb{R}^{m_1 \times 2}, A_2 \in \mathbb{R}^{m_2 \times 2}, b_1 \in \mathbb{R}^{m_1}, b_2 \in \mathbb{R}^{m_2}$ are constraint parameters from the scaled primitive and the counterpart primitive; $m_1,m_2 \in \mathbb{N}_{+}$; $Q_1 \in \mathbb{R}^{2\times2}$ is the rotation matrix calculated from the yaw angle and $O_1 \in \mathbb{R}^2$ is the position of the scaling center for the ego vehicle. Given that the shape of the ego vehicle is fixed, the constant constraint $A_1x \leq b_1$ denotes the original primitive of the ego vehicle, and the primitive at consecutive time steps can be defined via $Q_1$ and $O_1$ as shown in \eqref{eq:scaling_optimization_linear_form:ego_primitive}. In this paper, $O_1$ is set as the geometric center of the rectangle of the ego primitive. We further denote the ego primitive  as $P_1 \coloneqq \{x \mid A_1 Q_1^\top (x-O_1) \leq b_1\}$ and the counterpart primitive as $P_2 \coloneqq \{x \mid A_2 x \leq b_2\}$. For brevity, we denote the scaled version of $P_1$ with scaling factor $\gamma$ as $P_1(\gamma)$, and write $P_1(\gamma^*)$ as ${P_1}^*$ in the subsequent analysis. The optimal solution $\gamma^*$ is then adopted to assess the safety condition between the two primitives ${P_1}^*$ and $P_2$. Three representative cases associated with $\gamma^*$ are depicted in Fig. \ref{fig:scaling_factor}.

\begin{algorithm}[tbp]
\caption{Ray-Casting-based $\gamma$ Calculation.}
\label{alg:ray_casting_alpha_calc}
\begin{algorithmic}
\STATE 
\STATE $A_1 \in \mathbb{R}^{m_1\times2} \gets$ constraint matrix for $P_1$.
\STATE $b_1 \in \mathbb{R}^{m_1} \gets$ constraint vector for $P_1$.
\STATE $O_1 \in \mathbb{R}^2 \gets$ scaling origin of $P_1$. 
\STATE $Q_1 \in \mathbb{R}^{2\times2} \gets$ rotation matrix of $P_1$.
\STATE $V_w \in \mathbb{R}^2 \gets$ vertices of $P_1$, $w=1,2,\dots, n_v$, and $n_v$ denotes the total number of the vertices of $P_1$.
\STATE $A_2 \in \mathbb{R}^{m_2\times2} \gets$ constraint matrix for $P_2$.
\STATE $b_2 \in \mathbb{R}^{m_2} \gets$ constraint vector for $P_2$.
\STATE $\gamma_0 \gets \max \{A_2 O_1- b_2 \}$
\STATE $E_{w,i} \gets$ intersections between ray $\overrightarrow{O_1 V_w}$ and the sides of $P_2$ for $ w=1,2,\dots, n_v$. $i=1,2,\dots, n_w$ and $n_w$ denotes the total number of the intersections for the $w$-th ray.
\STATE $\bar{\gamma}_{w,i} \gets$ $\|\overline{O_1 E_{w,i}}\|/\|\overline{O_1 V_w}\|$ for $w=1,2,\dots, n_v$ and $i=1,2,\dots, n_w$.
\STATE $G_{w'} \gets$ extreme points of $P_2$, $w'=1,2,\dots, n_g$ and $n_g$ denotes the total number of the extreme points of $P_2$.
\STATE \textbf{if} $n_g=0$ \textbf{do}:
\STATE \hspace{0.5cm} $\gamma^* \gets g(\gamma_0) \cdot \min\{ \bar{\gamma}_{w,i} \}$
\STATE \textbf{else do}:
\STATE \hspace{0.5cm} $F_{w',j} \gets$ intersections between ray $\overrightarrow{O_1 G_{w'}}$ and the sides of $P_1$ for $w'=1,2,\dots, n_g$. $j=1,2,\dots, n_{w'}$ and $n_{w'}$ denotes the total number of intersections for $w'$-th ray.
\STATE \hspace{0.5cm} $\hat{\gamma}_{w',j} \gets \|\overline{O_1 G_{w'}}\|/\|\overline{O_1 F_{w',j}}\|$ for $w'=1,2,\dots, n_g$ and $j=1,2,\dots, n_{w'}$.
\STATE \hspace{0.5cm} 
$\gamma^* \gets g(\gamma_0) \cdot \min\{ \bar{\gamma}_{w,i}, \hat{\gamma}_{w',j} \}$.
\STATE \textbf{end if}
\STATE \textbf{return:} 
\STATE \hspace{0.5cm} $\gamma^*$ as the scaling factor.
\end{algorithmic}
\end{algorithm}

The Interior Point Method (IPM) is widely employed to solve problems such as \eqref{eq:scaling_optimization_linear_form}, but this iterative approach incurs high computational costs—particularly when executed on Graphics Processing Units (GPUs) during parallel DRL training. To address this issue, we first present a theorem stating that when $\gamma^*$ is attained, the corresponding $x^*$ lies on the vertices of ${P_1}^*$ or the extreme points of $P_2$. We then propose a direct ray-casting-based algorithm to solve Problem \eqref{eq:scaling_optimization_linear_form} in 2-D rigid body space, where multiple rays are emitted from the scaling center to the vertices of $P_1$ and extreme points of $P_2$ to obtain the minimum distance ratio.

\begin{theorem}
\label{thm:intersection_are_in_vertices_or_extreme_points}
Let $P_1$ be a polytope, i.e. $\forall x \in P_1, \exists \xi \in \mathbb{R}$ such that $\| x \| \leq \xi$. When the optimal $\gamma^*$ of problem \eqref{eq:scaling_optimization_linear_form} is obtained, the corresponding optimal point $x^*$ can be found at the vertices of ${P_1}^*$ or the extreme points of $P_2$.
\end{theorem}

\begin{proof}
We prove by case analysis based on whether $\gamma^*$ is zero:
\textbf{Case 1}: $\gamma^* = 0$.
Since $P_1$ is a polytope and $O_1 \in \operatorname{int}(P_1)$, $P_1(\gamma)|_{\gamma \to 0^+}$ will converge to $O_1$. Then $x^*= O_1$, which is a vertex of ${P_1}^*$. \\
\textbf{Case 2}: $\gamma^*>0$.
Since $P_1$ and $P_2$ are closed sets, it is reasonable to state that ${P_1}^* \cap P_2 = \partial{{P_1}^*} \cap \partial{P_2}$. Then $x^* \in \partial {P_1}^* \cap \partial P_2$. Each of the two systems of inequalities \eqref{eq:scaling_optimization_linear_form:ego_primitive} and \eqref{eq:scaling_optimization_linear_form:counterpart_primitive} has at least one constraint satisfied as an equality. \\
\textbf{Sub-case 1}: Either \eqref{eq:scaling_optimization_linear_form:ego_primitive} or \eqref{eq:scaling_optimization_linear_form:counterpart_primitive} has at least two constraints satisfied as equalities. 
By definition, $x^*$ is at the vertices of ${P_1}^*$ or the extreme points of $P_2$. \\
\textbf{Sub-case 2}: Each of \eqref{eq:scaling_optimization_linear_form:ego_primitive} and \eqref{eq:scaling_optimization_linear_form:counterpart_primitive} has one constraint satisfied as an equality.
This means that one side for ${P_1}^*$ and one side for $P_2$ overlap with each other. Hence, $x^*$ is on a line segment in $\partial {P_1}^* \cap \partial P_2$ and $x^*$ can be found at the endpoints, which are the vertices of ${P_1}^*$ or the extreme points of $P_2$.

In conclusion, $x^*$ can be found at the vertices of ${P_1}^*$ or the extreme points of $P_2$. The proof is complete.
\end{proof}

The pseudocode of the algorithm is presented in Algorithm \ref{alg:ray_casting_alpha_calc}. Here, the function $g(\cdot)$ is a step function. Now we provide a mathematical analysis of the computational complexity and timing performance. For our formulation with exactly three decision variables and $m_1 + m_2$ constraints, IPM requires $\mathcal{O}(\sqrt{m_1 + m_2}\log \frac{1}{\epsilon_{\text{ipm}}})$ iterations to reach a precision $\epsilon_{\text{ipm}}$. Each iteration involves $\mathcal{O}(m_1 + m_2)$ algebraic operations to compute the corresponding Newton step, resulting in an overall time complexity of $\mathcal{O}((m_1 + m_2)^{\frac{3}{2}} \log \frac{1}{\epsilon_{\text{ipm}}})$ \cite{bazaraa2011linear}. Furthermore, this mathematical complexity represents an expected theoretical cost. In parallel GPUs environments, the execution speed of such iterative algorithms is strictly bottlenecked by the slowest instance within a computation group. Consequently, the overall computation time of a parallel batch is dictated by the worst-case scenario that requires the maximum number of iterations. This synchronization overhead, combined with algorithmic branching, causes severe warp divergence and makes the IPM prohibitively expensive for large-scale DRL training. 

In contrast, our direct ray-casting-based algorithm operates without any iterative loops. According to Algorithm \ref{alg:ray_casting_alpha_calc}, the extraction of $n_g$ extreme points from the half-plane representation of $P_2$ further contributes $\mathcal{O}(m_2 \log m_2)$ to the computational cost. The calculation of $\gamma_0$ requires $\mathcal{O}(m_2)$ operations to evaluate the constraint set. The subsequent ray intersections involve $n_v$ rays from the ego primitive intersecting with $m_2$ sides of the counterpart primitive, and $n_g$ rays from the counterpart primitive intersecting with $m_1$ sides of the ego primitive. Combined with the final minimum value selection among all valid intersections, the overall time complexity is exactly bounded by $\mathcal{O}(m_2 \log m_2 + n_v \cdot m_2 + n_g \cdot m_1)$. This deterministic computational load inherently avoids the iteration and synchronization overheads found in the IPM.

Empirical timing data during the parallel training process further substantiates this efficiency. The equipment used for the training test is the same as that in Section \ref{sec:experiments}. For the computational efficiency benchmark, a batch of $65,536$ primitive evaluations is considered, corresponding to $128$ environments, $128$ rollout steps, $1$ ego vehicle, $2$ obstacles, and $2$ road boundaries. Under the same GPU parallel configuration with two RTX 4090 GPUs, the standard IPM solver requires approximately $5$ s to complete these evaluations, corresponding to an average amortized time of $76.3$ $\mu$s per primitive evaluation. In contrast, the proposed ray-casting algorithm requires approximately $1$ s, corresponding to an average amortized time of $15.3$ $\mu$s per primitive evaluation. These values are calculated by dividing the total execution time by the number of primitive evaluations under the parallel training configuration.

Regarding scalability in dense multi-obstacle scenarios, the computational cost of our algorithm scales linearly as $\mathcal{O}(N)$ with respect to $N$ obstacles. Because the unit execution time is at the microsecond level, evaluating dozens of surrounding elements remains computationally efficient. In practical system deployments, this exact evaluation is mathematically constrained to a localized subset of obstacle primitives $P_2$ satisfying an initial spatial bounding condition. For example, the system evaluates only the obstacles whose Euclidean distance from the ego scaling center is less than or equal to a defined V2X communication range threshold. This ensures that the exact ray-casting algorithm is executed exclusively within the relevant local domain, thereby guaranteeing computational scalability across varying traffic densities.

Based on the calculation of $\gamma^*$, we can efficiently compute the safety factor $H_k$ in \eqref{eq:safe_trajectory_tracking_problem_formulation} as
\begin{equation}
\label{eq:h_definition}
\begin{split}
h_k^i &= 1-{\gamma_k^i}^*, \\
H_k &= \max \{ h_k^1, \ h_k^2, \ \dots, \ h_k^{N_o}  \} 
\end{split}
\end{equation}
where ${\gamma_k^i}^*$ is the optimal scaling factor between the ego vehicle and the $i$-th obstacle at the $k$-th time step.

\begin{remark}
\label{rmk:robustness_in_application}
To enhance robustness in practical implementations, we incorporate modifications into \eqref{eq:h_definition}. $h^i_k$ can be calculated as
\begin{equation}
\label{eq:robust_h_k}
h^i_k = \gamma_{\text{t}} - {\gamma^i_k}^*,
\end{equation}
where $\gamma_{\text{t}}$ is a threshold parameter slightly greater than $1$. This choice of $\gamma_{\text{t}}$ introduces safety margins because even as $h^i_k$ approaches $0$, the scaling factor of the ego vehicle remains above $1$, meaning some distance to collision is preserved. After that, $H_k$ is computed as
\begin{equation}
\label{eq:robust_H_k}
H_k = {H_k}' + \zeta \cdot g({H_k}'),
\end{equation}
where ${H_k}'=\max \{ h_k^1, \ h_k^2, \ \dots, \ h_k^{N_o}  \}$ and $\zeta$ is a positive parameter. Introducing $\zeta$ helps reduce the learning error of the safety critic NN near the boundary between safe and unsafe conditions during training, thereby improving the accuracy of the bisection result.
\end{remark}

\section{Experiments}
\label{sec:experiments}

\begin{figure}[tbp]
\centering
\includegraphics[width=0.95\linewidth]{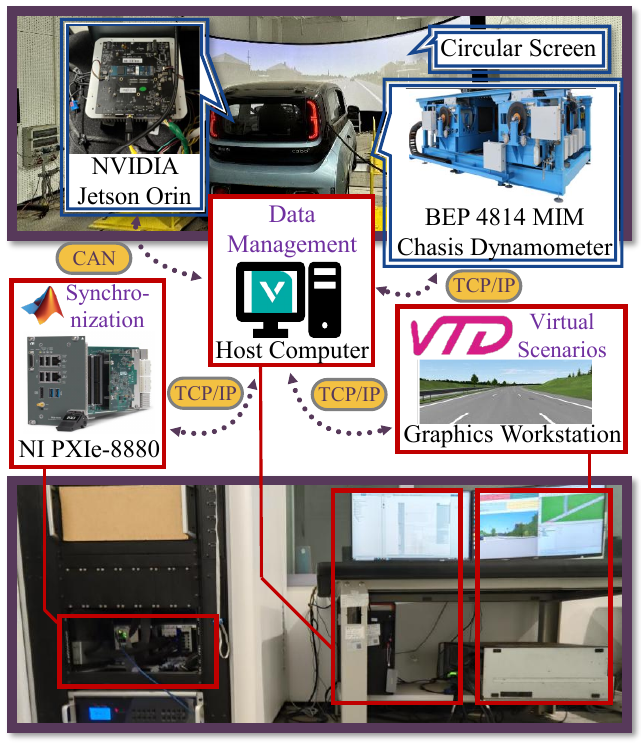}
\caption{The structure of the ViL platform.}
\label{fig:equipment_introduction}
\end{figure}

This section evaluates the proposed STTC-EFRL during training and implementation. To begin with, we benchmark the training efficiency against the state-of-the-art PPO-Lagrangian \cite{gu2023safe} and Deep Deterministic Policy Gradient (DDPG)-Lagrangian methods \cite{jin2025connected} in randomly generated simulation scenarios based on the 3-DOF dynamic model \eqref{eq:dynamic_model_simplified}. Subsequently, we conduct a comparative analysis of trajectory tracking and collision avoidance performance against both the Lagrangian methods and the non-learning-based Uniform Finite Time Safe Path Tracking Control (UFTSTC) approach \cite{han2024uniform} across six fixed scenarios via ViL experiments. The ViL platform is shown in Fig. \ref{fig:equipment_introduction}. In the simulation and ViL experiments, we assume ideal V2X communication, with no communication delays or packet loss. Under this assumption, the experiments evaluate the effectiveness of the proposed STTC-EFRL. For practical CAV deployments, advanced communication protocols \cite{fei2025echelon} and heterogeneous networking architectures \cite{liu2026a} can be considered to help reduce communication delays and packet loss.

\begin{table}[t]
    \begin{center}
    \caption{Random Parameters in Type-1 and Type-2 training scenarios generation.}
    \label{tab:scen_gen_params_straight}
    \begin{tabular}{c l}
    \noalign{\hrule height 1pt}
    \textbf{Name} & \textbf{Range} \\
    \hline
    x coordinate of the start of the lane change segment & $[0,40]$ m \\
    x coordinate of the end of the lane change segment & $[60,100]$ m \\
    x coordinate of the start of the ego vehicle & $[0,30]$ m \\
    x coordinate of the static vehicle & $[50,80]$ m \\
    initial speed of the ego vehicle & $[60,90]$ km/h \\
    speed of the moving vehicle & $[20,120]$ km/h \\
    speed in the reference trajectory & $[60,90]$ km/h \\
    \noalign{\hrule height 1pt}
    \end{tabular}
    \end{center}
\end{table}

\begin{table}[t]
    \begin{center}
    \caption{Random Parameters in Type-3 training scenarios generation.}
    \label{tab:scen_gen_params_intersection}
    \begin{tabular}{c l}
    \noalign{\hrule height 1pt}
    \textbf{Name} & \textbf{Range} \\
    \hline
    x coordinate of the start of the ego vehicle & $[-100,-80]$ m \\
    initial speed of the ego vehicle & $[60,90]$ km/h \\
    speed of the moving vehicle & $[20,100]$ km/h \\
    speed in the reference trajectory at the entrance & $[60,90]$ km/h \\
    speed in the reference trajectory during turning & $[30,40]$ km/h \\
    \noalign{\hrule height 1pt}
    \end{tabular}
    \end{center}
\end{table}

\begin{table}[t]
    \begin{center}
    \caption{Other parameters.}
    \label{tab:other_params}
    \begin{tabular}{c c c c}
    \noalign{\hrule height 1pt}
    \textbf{Name} & \textbf{Value} & \textbf{Name} & \textbf{Value} \\
    \hline
    lane width & $3$ m & turning area & $35$ m $\times$ $35$ m \\
    ego size & $2.63$ m $\times$ $1.65$ m & obstacle size & $4.10$ m $\times$ $2.05$ m \\
    $C_f$ & $95700$ N/rad & $C_r$ & $93020$ N/rad \\
    $l_f$ & $0.8475$ m & $l_r$ & $0.9025$ m \\
    $M$ & $940$ kg & $I_z$ & $752.25$ kg$\cdot\text{m}^2$\\
    $\nu$ & $0.05$ & $\gamma_{\text{t}}$ & $1.05$ \\
    $\zeta$ & $1$ & $\epsilon_p$ & $0.1$ \\
    $\tau$ & $0.05$ s & $T$ & $256$ / $384$ \\
    $\lambda_Q$ & $0.95$ & update epoch & $1$ \\
    \noalign{\hrule height 1pt}
    \end{tabular}
    \end{center}
\end{table}

\subsection{Training Efficiency Tests}
\begin{figure}[htbp]
    \centering
    \subfloat[]{
    \includegraphics[width=0.9\linewidth]{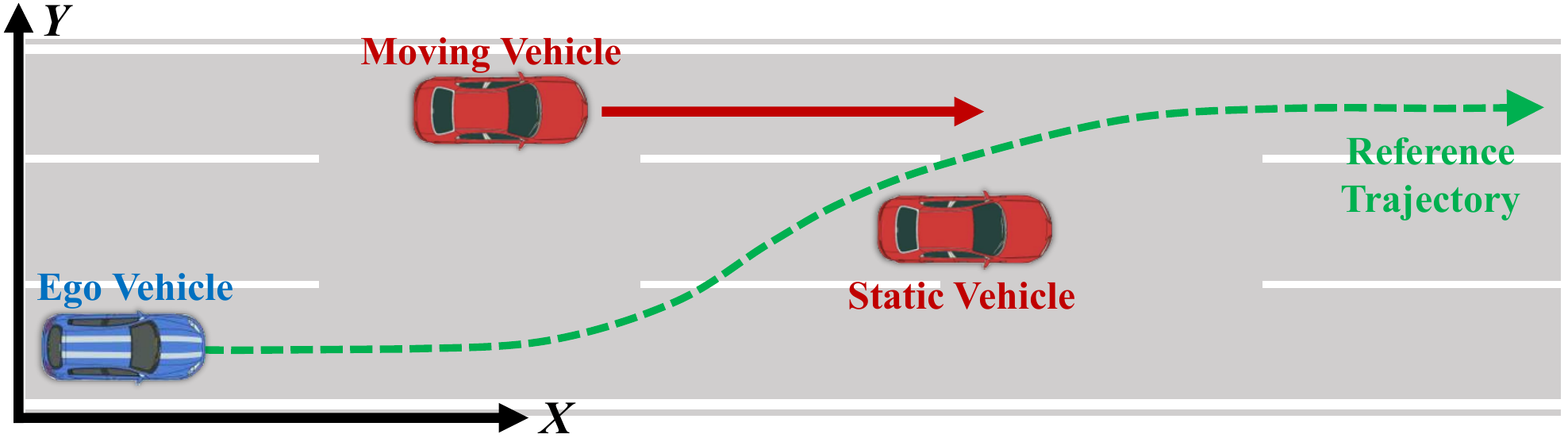}
    \label{fig:lane_change_scenario_2}
    }
    \hfill
    \subfloat[]{
    \includegraphics[width=0.9\linewidth]{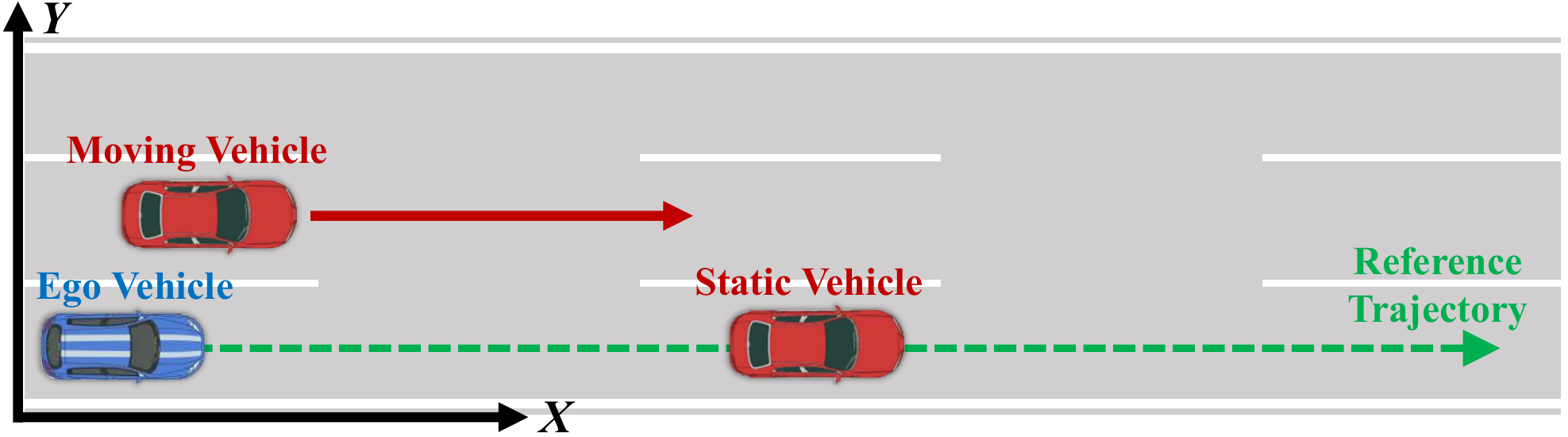}
    \label{fig:overtake_scenario}
    }
    \hfill
    \subfloat[]{
    \includegraphics[width=0.9\linewidth]{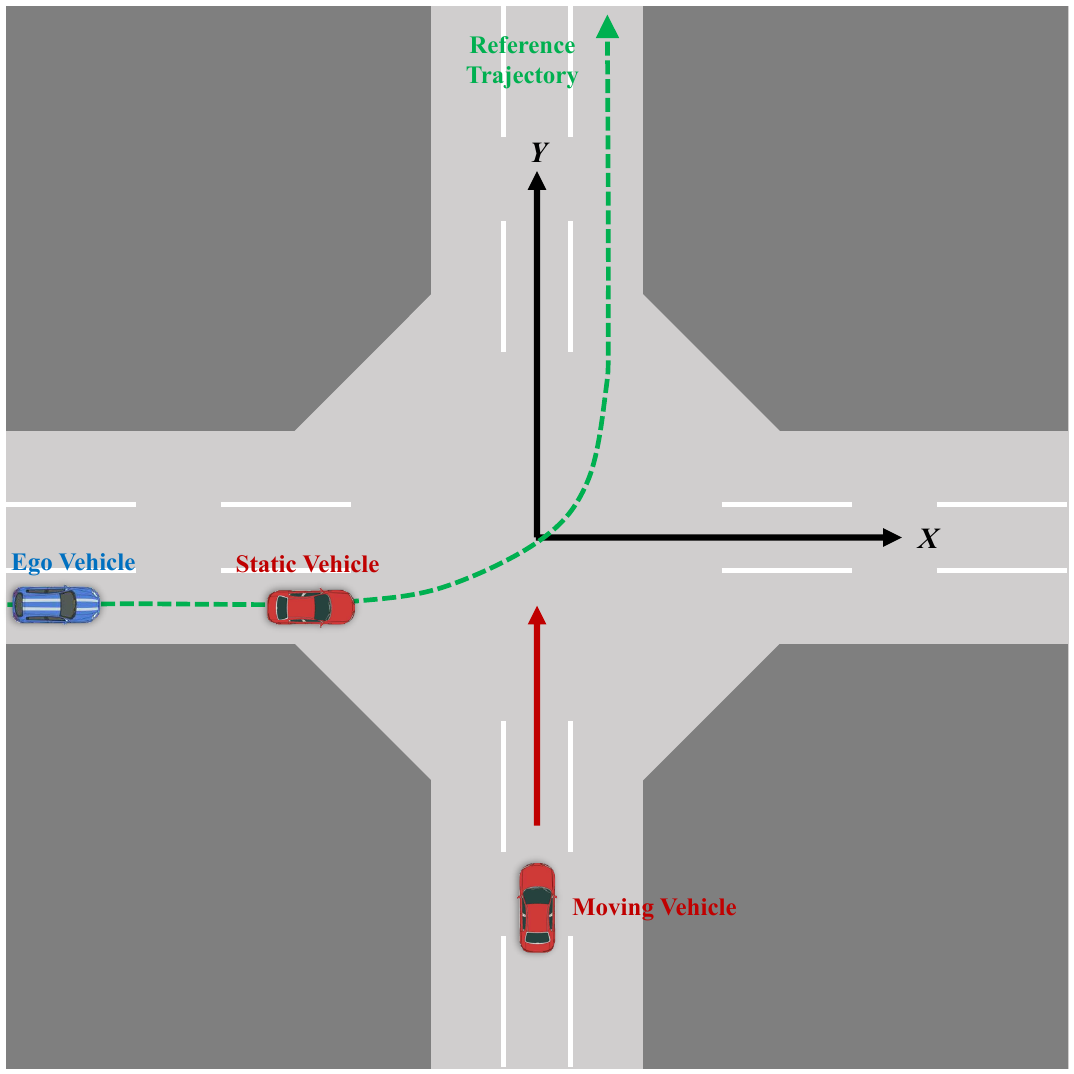}
    \label{fig:intersection_scenario}
    }
    \caption{Schematic of three types of training scenarios. (a) is Type-1: the lane-changing scenario. (b) is Type-2: the overtaking scenario. (c) is Type-3: the intersection scenario.}
    \label{fig:scenarios_for_experiments}
\end{figure}

Three comprehensive types of driving scenarios that align with typical CAV applications are designed and randomly generated for DRL training, as illustrated in Fig. \ref{fig:scenarios_for_experiments}. The road topologies, including lane numbers, lane widths, and lane directions, remain fixed. All scenarios incorporate one static vehicle and one moving vehicle as obstacles, with randomly initialized states. The first type (Type-1) is a lane-changing case, as illustrated in Fig. \ref{fig:lane_change_scenario_2}, where the ego vehicle changes from the first lane to the third lane or vice versa. The second type (Type-2) is an overtaking case, as illustrated in Fig. \ref{fig:overtake_scenario}, where the ego vehicle moves straight forward in the first or third lane. The third type (Type-3) is an unsignalized intersection case, as illustrated in Fig. \ref{fig:intersection_scenario}, where the ego vehicle navigates the intersection by either proceeding straight, turning left, or turning right. In Type-3 scenarios, the static vehicle is randomly deployed along the reference trajectory, and the moving vehicle's initial position is randomly generated among the four entries of the intersection. All scenarios require the ego vehicle to interact with obstacle vehicles within a communication range of $50$ m.

For the reference trajectories in Type-1 and Type-2 scenarios, the lane-changing segment is generated via a quintic polynomial, and the scenario randomization parameters are summarized in Table \ref{tab:scen_gen_params_straight}. For the reference trajectories in the Type-3 scenario, the turning segment is generated via an arc curve, and the scenario randomization parameters are summarized in Table \ref{tab:scen_gen_params_intersection}. When resetting scenarios at each iteration, a dynamic stochastic scenario generation mechanism is applied, where the initial states of the ego and obstacle vehicles are randomized by a pseudo-random number generator (PRNG) bounded by these predefined ranges. During training, the parameters in the 3-DOF dynamic models for the ego vehicle are calibrated using the electric vehicle employed in the ViL experiments. Parameters of the 3-DOF dynamic models and other training environment parameters are listed in Table \ref{tab:other_params}. We first conducted a training efficiency test in Type-1 and Type-2 scenarios, followed by another training efficiency test conducted solely in Type-3 scenarios.

\begin{table}[t]
    \color{black}
    \begin{center}
    \caption{Overall performance of ViL experiments.}
    \label{tab:ViL_results}
    \begin{tabular}{c c|c c c c c}
    \noalign{\hrule height 1pt}
    & & $e^{\text{mean}}_{\text{lateral}}$ & $\sigma_{\text{lateral}}$ & Cost & $H_k^{\max}$ & Safety \\
    \hline
    \multirow{4}{*}{I} & \multicolumn{1}{|c|}{STTC-EFRL} & -1.33 & \textbf{1.39} & \textbf{5.67} & \textbf{-0.61} & $\checkmark$ \\
    & \multicolumn{1}{|c|}{PPO-L} & -1.73 & 1.82 & 5.79 & 0.38 & $\times$ \\
    & \multicolumn{1}{|c|}{DDPG-L} & \textbf{-1.07} & 2.31 & 8.54 & -0.09 & $\checkmark$ \\
    & \multicolumn{1}{|c|}{UFTSTC} & -1.17 & 2.73 & 6.72 & -0.24 & $\checkmark$ \\
    \hline
    \multirow{4}{*}{II} & \multicolumn{1}{|c|}{STTC-EFRL} & \textbf{-0.59} & \textbf{1.02} & \textbf{5.23} & \textbf{-0.23} & $\checkmark$ \\
    & \multicolumn{1}{|c|}{PPO-L} & -1.79 & 1.10 & 6.54 & -0.10 & $\checkmark$ \\
    & \multicolumn{1}{|c|}{DDPG-L} & -1.15 & 2.65 & 8.43 & -0.20 & $\checkmark$ \\
    & \multicolumn{1}{|c|}{UFTSTC} & -1.21 & 2.66 & 6.58 & -0.04 & $\checkmark$ \\
    \hline
    \multirow{4}{*}{III} & \multicolumn{1}{|c|}{STTC-EFRL} & 0.98 & 1.40 & 4.86 & \textbf{-0.75} & $\checkmark$ \\
    & \multicolumn{1}{|c|}{PPO-L} & \textbf{0.37} & \textbf{0.23} & \textbf{0.93} & 1.00 & $\times$ \\
    & \multicolumn{1}{|c|}{DDPG-L} & 2.30 & 1.75 & 11.94 & 1.00 & $\times$ \\
    & \multicolumn{1}{|c|}{UFTSTC} & 3.31 & 2.54 & 8.75 & 0.79 & $\times$ \\
    \hline
    \multirow{4}{*}{IV} & \multicolumn{1}{|c|}{STTC-EFRL} & 1.35 & 1.58 & 4.53 & -0.32 & $\checkmark$ \\
    & \multicolumn{1}{|c|}{PPO-L} & \textbf{0.16} & \textbf{1.03} & \textbf{1.92} & \textbf{-0.38} & $\checkmark$ \\
    & \multicolumn{1}{|c|}{DDPG-L} & 2.76 & 1.88 & 12.56 & -0.03 & $\checkmark$ \\
    & \multicolumn{1}{|c|}{UFTSTC} & 3.45 & 2.27 & 9.40 & -0.35 & $\checkmark$ \\
    \hline
    \multirow{4}{*}{V} & \multicolumn{1}{|c|}{STTC-EFRL} & 0.40 & 0.45 & 2.94 & -0.23 & $\checkmark$ \\
    & \multicolumn{1}{|c|}{PPO-L} & \textbf{0.17} & \textbf{0.41} & \textbf{2.79} & 1.00 & $\times$ \\
    & \multicolumn{1}{|c|}{DDPG-L} & 2.16 & 0.80 & 8.80 & \textbf{-0.25} & $\checkmark$ \\
    & \multicolumn{1}{|c|}{UFTSTC} & 2.04 & 7.64 & 9.37 & 0.24 & $\times$ \\
    \hline
    \multirow{4}{*}{VI} & \multicolumn{1}{|c|}{STTC-EFRL} & 0.39 & \textbf{0.69} & \textbf{3.71} & \textbf{-0.31} & $\checkmark$ \\
    & \multicolumn{1}{|c|}{PPO-L} & -1.26 & 10.08 & 11.64 & 1.00 & $\times$ \\
    & \multicolumn{1}{|c|}{DDPG-L} & \textbf{-0.07} & 3.93 & 9.76 & 1.00 & $\times$ \\
    & \multicolumn{1}{|c|}{UFTSTC} & 1.02 & 2.59 & 6.97 & -0.22 & $\checkmark$ \\
    \noalign{\hrule height 1pt}
    \end{tabular}
    \end{center}
    \small
    Note: The Roman numerals I to VI denote respectively Scenario I to Scenario VI. The bold numbers denote the best performance of this criteria among the compared algorithms in the specific scenario. $e^{\text{mean}}_{\text{lateral}}$ denotes the mean tracking error (m), $\sigma_{\text{lateral}}$ denotes the standard deviation of the tracking error (m), and $H_k^{\max}$ denotes the maximum safety factor in the sampled trajectory of a single scenario using \eqref{eq:h_definition}.
\end{table}

\begin{figure}[htbp]
    \centering
    \subfloat[]{
    \includegraphics[width=0.45\linewidth]{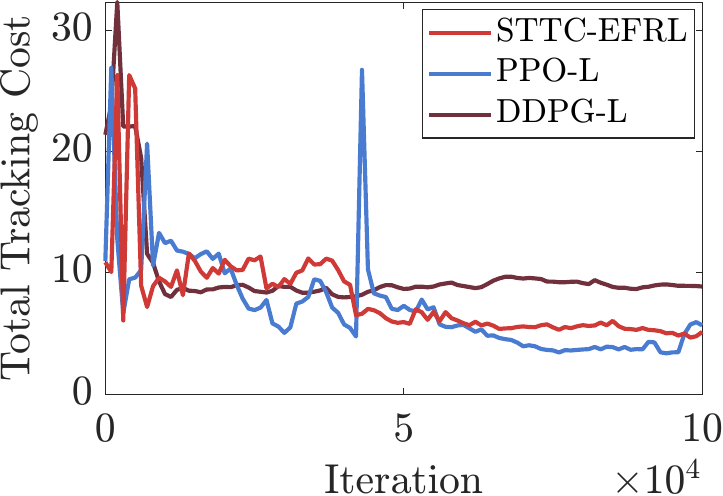}
    \label{fig:training_straight:cost}
    }
    \subfloat[]{
    \includegraphics[width=0.45\linewidth]{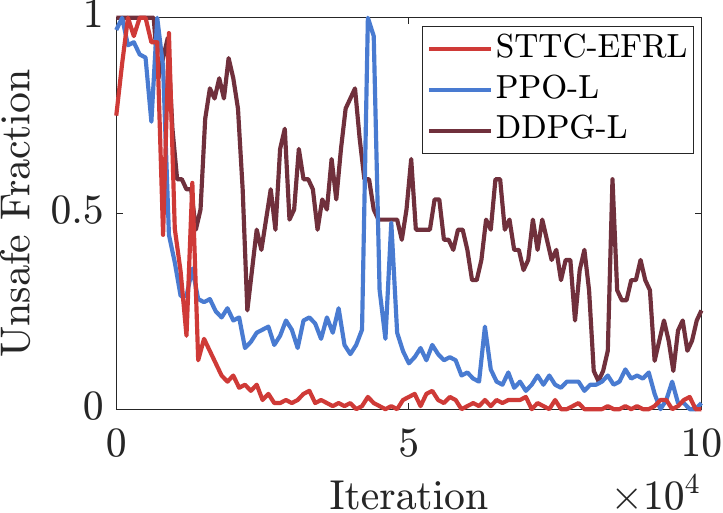}
    \label{fig:training_straight:unsafe_frac}
    }
    \hfill
    \subfloat[]{
    \includegraphics[width=0.45\linewidth]{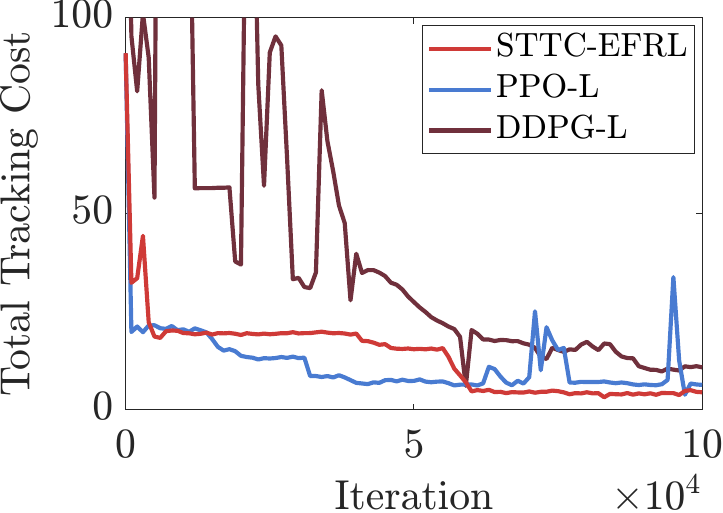}
    \label{fig:training_intersection:cost}
    }
    \subfloat[]{
    \includegraphics[width=0.45\linewidth]{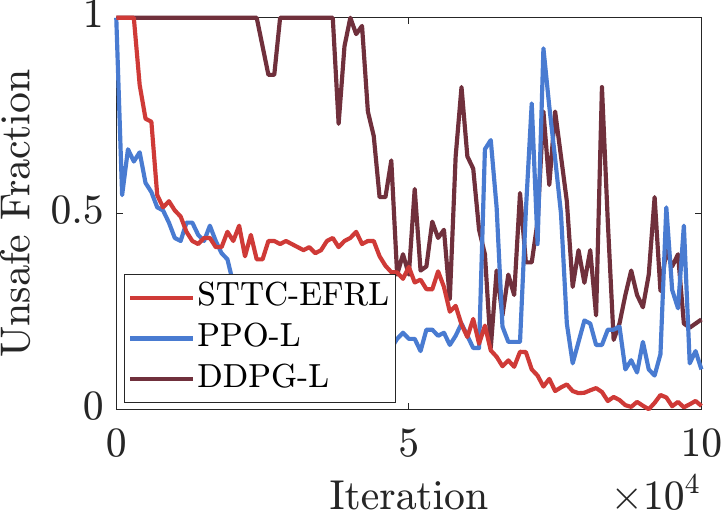}
    \label{fig:training_intersection:unsafe_frac}
    }
    \caption{Total tracking cost in (a) and unsafe fraction in (b) in Type-1 and Type-2 scenarios, and total tracking cost in (c) and unsafe fraction in (d) in Type-3 scenarios, w.r.t. training iterations.}
    \label{fig:training}
\end{figure}

\begin{figure}[htbp]
    \centering
    \subfloat[]{
    \includegraphics[width=0.45\linewidth]{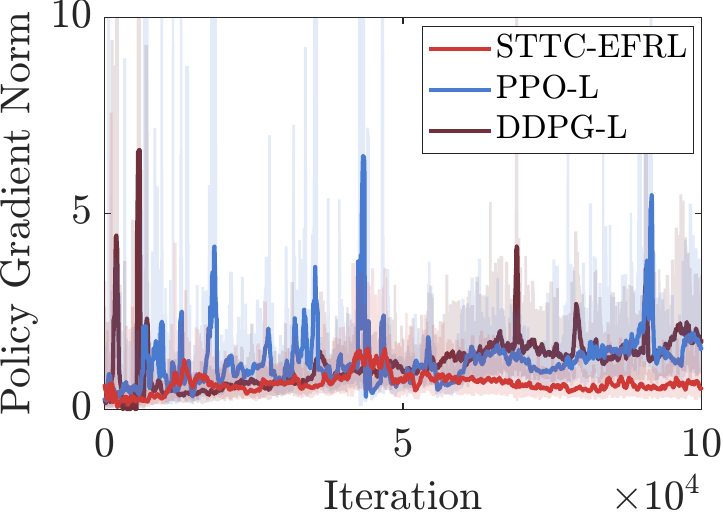}
    \label{fig:training_straight_inter_metric:policy_grad}
    }
    \subfloat[]{%
    \includegraphics[width=0.45\linewidth]{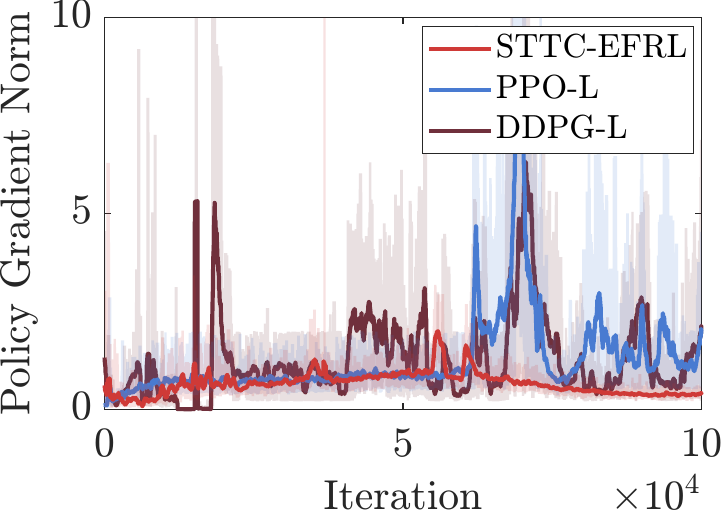}
    \label{fig:training_straight_inter_metric:cost_critic_loss}
    }
    \hfill
    \subfloat[]{%
    \includegraphics[width=0.45\linewidth]{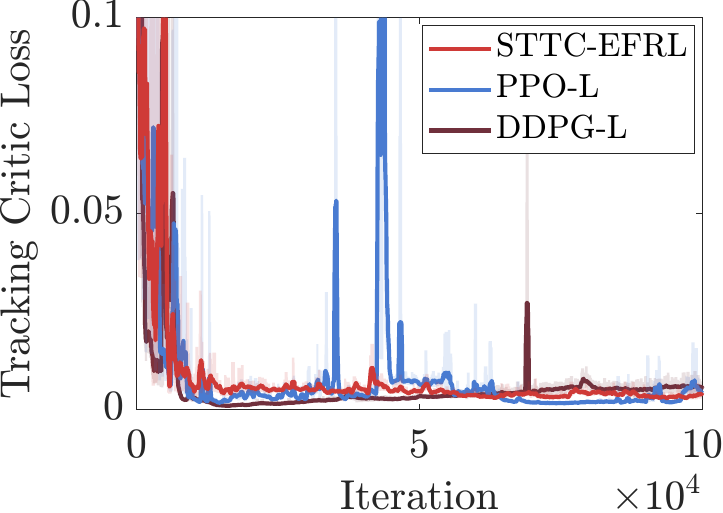}
    \label{fig:training_straight_inter_metric:safety_critic_loss}
    }
    \subfloat[]{%
    \includegraphics[width=0.45\linewidth]{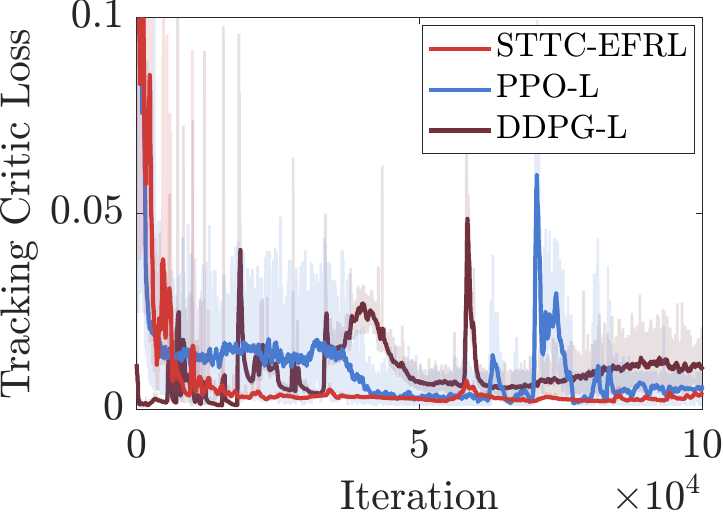}
    \label{fig:training_intersection_inter_metric:policy_grad}
    }
    \hfill
    \subfloat[]{%
    \includegraphics[width=0.45\linewidth]{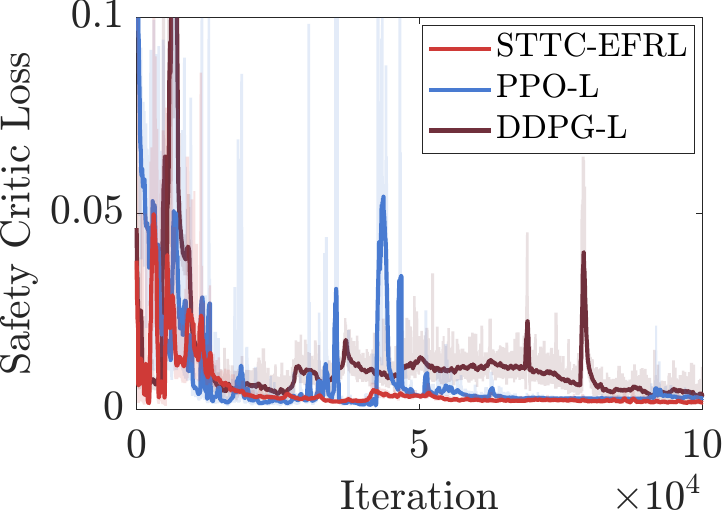}
    \label{fig:training_intersection_inter_metric:cost_critic_loss}
    }
    \subfloat[]{%
    \includegraphics[width=0.45\linewidth]{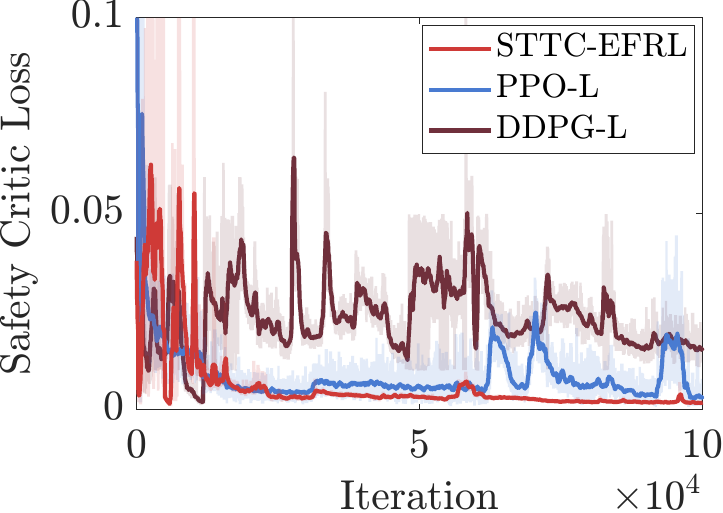}
    \label{fig:training_intersection_inter_metric:safety_critic_loss}
    }
    \caption{Policy gradient norm in (a), tracking critic loss in (c), safety critic loss in (e) in Type-1 and Type-2 scenarios, and policy gradient norm in (b), tracking critic loss in (d), safety critic loss in (f) in Type-3 scenarios, w.r.t. training iterations.}
    \label{fig:training_inter_metric}
\end{figure}

A $4$-layer graph transformer architecture \cite{nayak2023scalable} is adopted to construct both the actor and critic NNs. $W$ in \eqref{eq:safe_trajectory_tracking_problem_formulation} is set to:
\begin{equation}
\label{eq:W_definition}
W = \text{diag} (10^{-4}, \ 10^{-4}, \ 2.5\times10^{-7}, \ 10^{-6}).
\end{equation}

During training, we benchmark our STTC-EFRL ($z_{\max}=10$) against two state-of-the-art Lagrangian methods: PPO-Lagrangian ($\lambda_\text{init}=5$, denoted as PPO-L) and DDPG-Lagrangian ($\lambda_\text{init}=5$, denoted as DDPG-L), where $\lambda_\text{init}$ denotes the initial value of the Lagrangian multiplier. All DRL algorithms are trained on a desktop with an Intel Xeon 8468V CPU and two RTX 4090 GPUs for $100000$ iterations. To accelerate training, $128$ sampled trajectories are rolled out on each GPU simultaneously for each iteration. The trajectories are rolled out for $256$ steps per iteration in the Type-1 and Type-2 training environments, while in the Type-3 scenario, the rollout length is extended to $384$ steps to accommodate the complex intersection dynamics. All learning-based models in this study are trained using a single global random seed, which initializes the aforementioned PRNG to govern the NN initialization, the PPO action sampling at every iteration and the scenario resetting.

The total tracking cost and unsafe fraction curves are shown in Fig. \ref{fig:training}, while the policy gradient norms and critic loss evolution are illustrated in Fig. \ref{fig:training_inter_metric}. As these figures demonstrate, PPO-L and DDPG-L suffer from instability in both policy gradient norms and critic losses. This instability occurs because, upon encountering safety violations, the rapidly growing Lagrangian multiplier directly scales the policy gradient, leading to explosive gradient norms and catastrophic oscillations in the critic loss. This explicitly explains why the Lagrangian policies oscillate in the tracking cost and unsafe fraction curves across all training scenarios, and why they exhibit undesirable, unstable behaviors during ViL experiments in dynamic environments. In contrast, our STTC-EFRL utilizes a monotonically decreasing epigraph scalar, which successfully bounds the policy gradient magnitude. The results show that under this single-seed training setup, STTC-EFRL maintains relatively stable gradient norms and achieves a smoother, steadier convergence trend compared to the Lagrangian baselines by the end of training.

\begin{figure}[bp]
    \centering
    \includegraphics[width=1.0\linewidth]{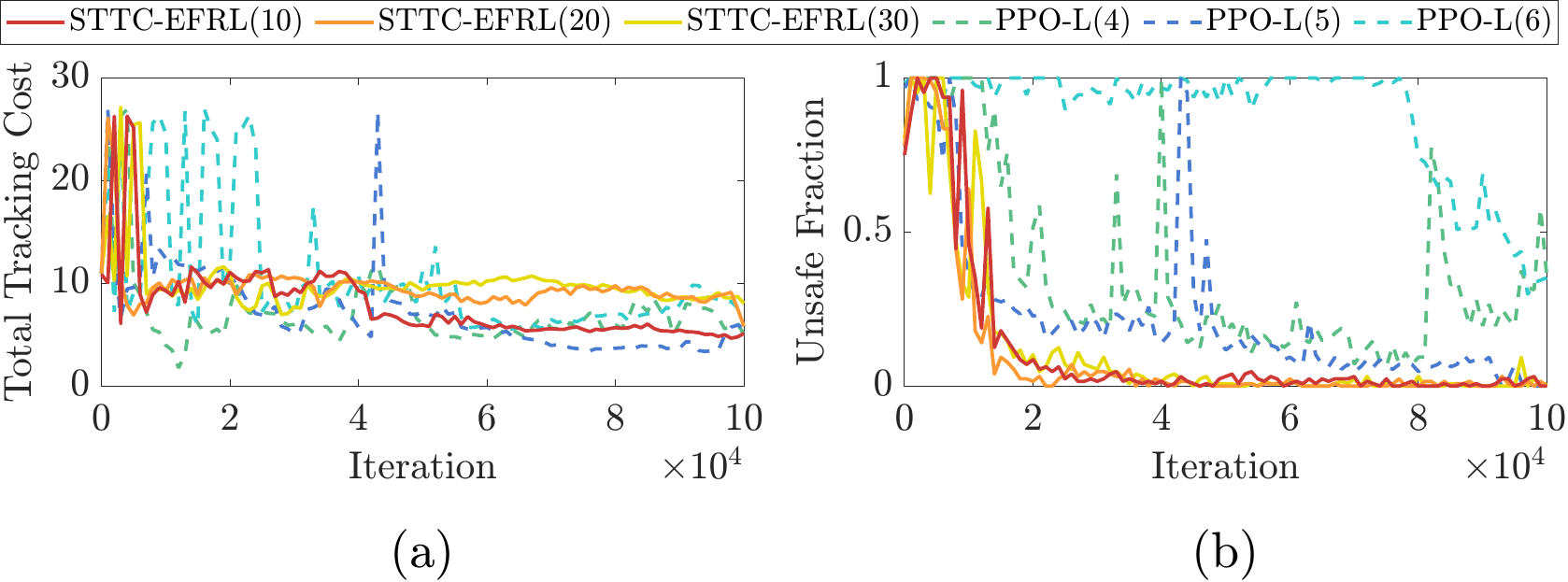}
    \caption{Total tracking cost in (a) and unsafe fraction in (b) w.r.t. training iterations with different hyper parameters. Numbers in brackets beside STTC-EFRL denote $z_{\max}$ and beside PPO-L denote $\lambda_{\text{init}}$.}
    \label{fig:hyper}
\end{figure}

Additionally, we implement another experiment to demonstrate PPO-L's high sensitivity to its hyperparameter $\lambda_{\text{init}}$. As shown in Fig. \ref{fig:hyper}, when $\lambda_{\text{init}}$ ranges from $4$ to $6$ during Type-1 and Type-2 training, the total tracking cost of PPO-L increases without significant improvement in safety performance. Conversely, STTC-EFRL demonstrates better adaptability to its hyperparameter $z_{\max}$, with convergence speed and stability remaining nearly unaffected when $z_{\max}$ varies from $10$ to $30$.

\subsection{Safe Trajectory Tracking Validations}
During ViL experiments, we reconstruct the NN inference process using TensorRT on the NVIDIA Jetson AGX Orin of the ego vehicle. The bisection search takes approximately 2.54 ms per step. The state encoding and actor NN inference take approximately 0.19 ms per step. This computation time is sufficient for the real-time control requirements in our experiment setup.

We then evaluate the STT performance of the proposed STTC-EFRL ($z_{\max}=10$) in comparison with the PPO-L ($\lambda_\text{init}=5$), DDPG-L ($\lambda_\text{init}=5$), and UFTSTC approaches in specific lane-changing, overtaking, and intersection scenarios. The evaluated scenarios are generated based on the three training types. Scenarios I and II correspond to Type-1, while Scenarios III and IV correspond to Type-2, varying only in the speed of the moving vehicle. In Scenarios I to IV, the reference speed is set to $70$ km/h, and the initial speed of the ego vehicle is $80$ km/h. The moving vehicle speed is set to $90$ km/h in fast scenarios and $50$ km/h in slow scenarios. Scenarios V and VI correspond to the new Type-3 scenarios. In Scenario V, the ego vehicle proceeds straight while the moving vehicle is spawned at the south entrance. In Scenario VI, the ego vehicle turns left while the moving vehicle is spawned at the east entrance. For both Scenarios V and VI, the initial speed of the ego vehicle and the speed of the moving vehicle are set to $60$ km/h, with a reference speed of $70$ km/h at the entrance and $30$ km/h during the turn. 

\begin{figure}[tbp]
    \centering
    \subfloat[]{
    \includegraphics[width=0.49\linewidth]{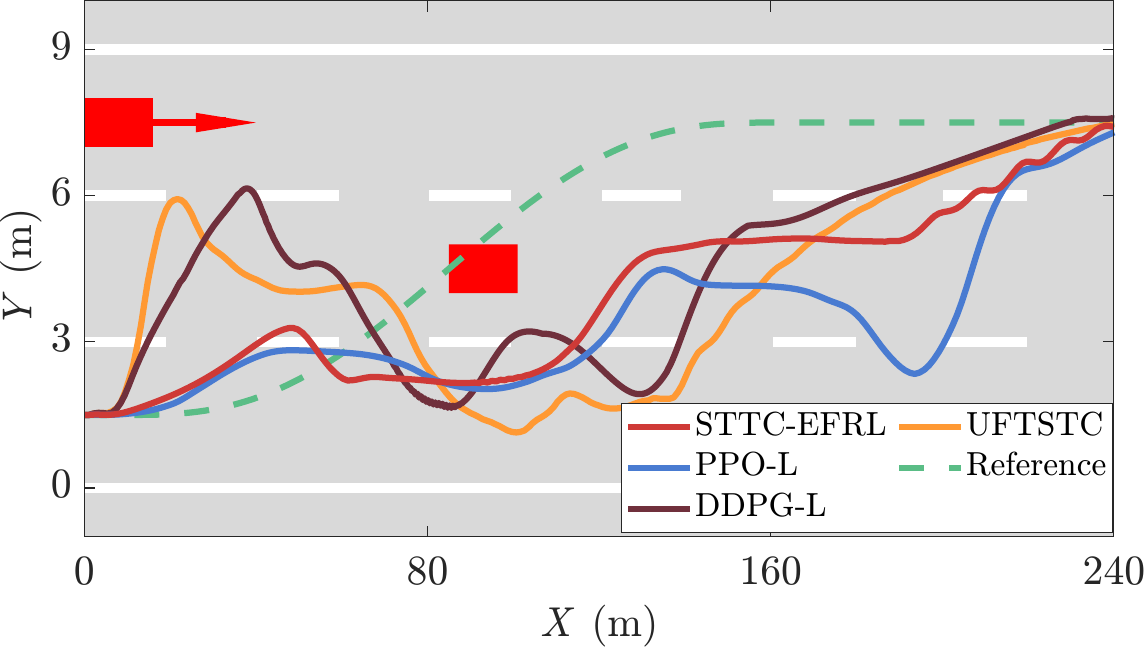}
    \label{fig:x_y_scene_i}
    }
    \subfloat[]{
    \includegraphics[width=0.49\linewidth]{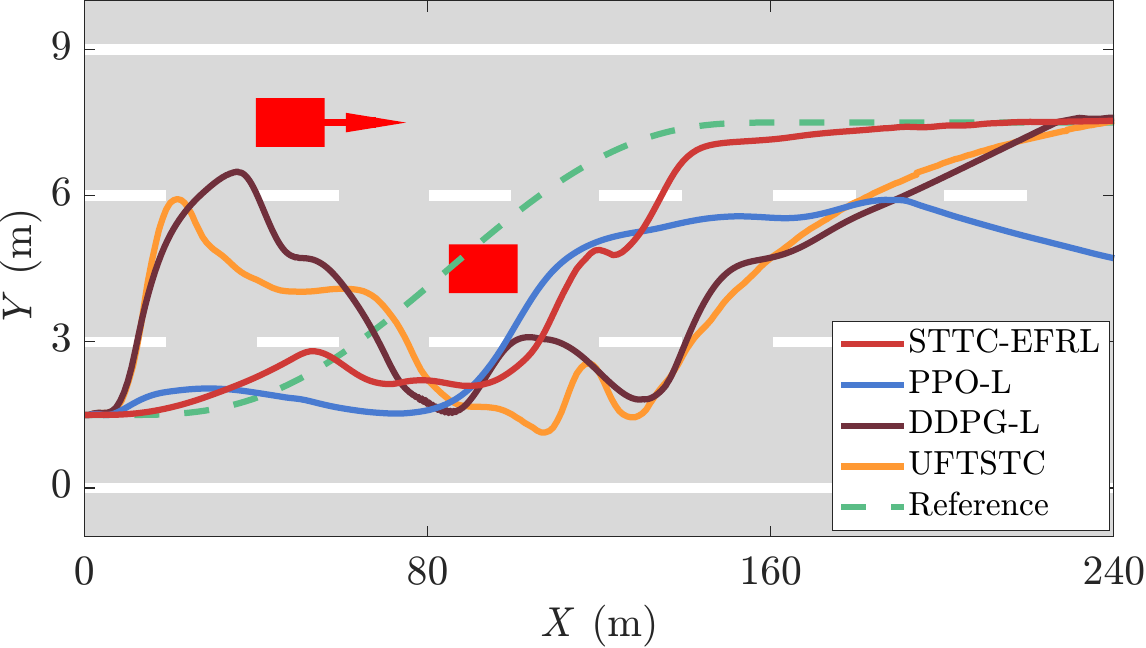}
    \label{fig:x_y_scene_ii}
    }
    \hfill
    \subfloat[]{
    \includegraphics[width=0.49\linewidth]{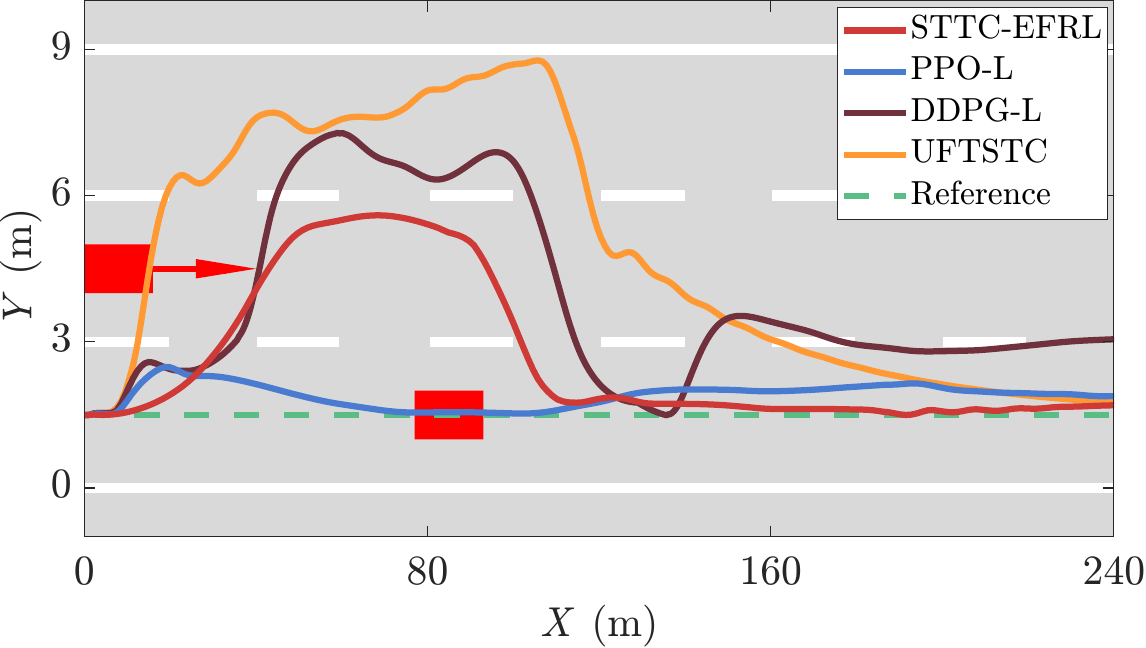}
    \label{fig:x_y_scene_iii}
    }
    \subfloat[]{
    \includegraphics[width=0.49\linewidth]{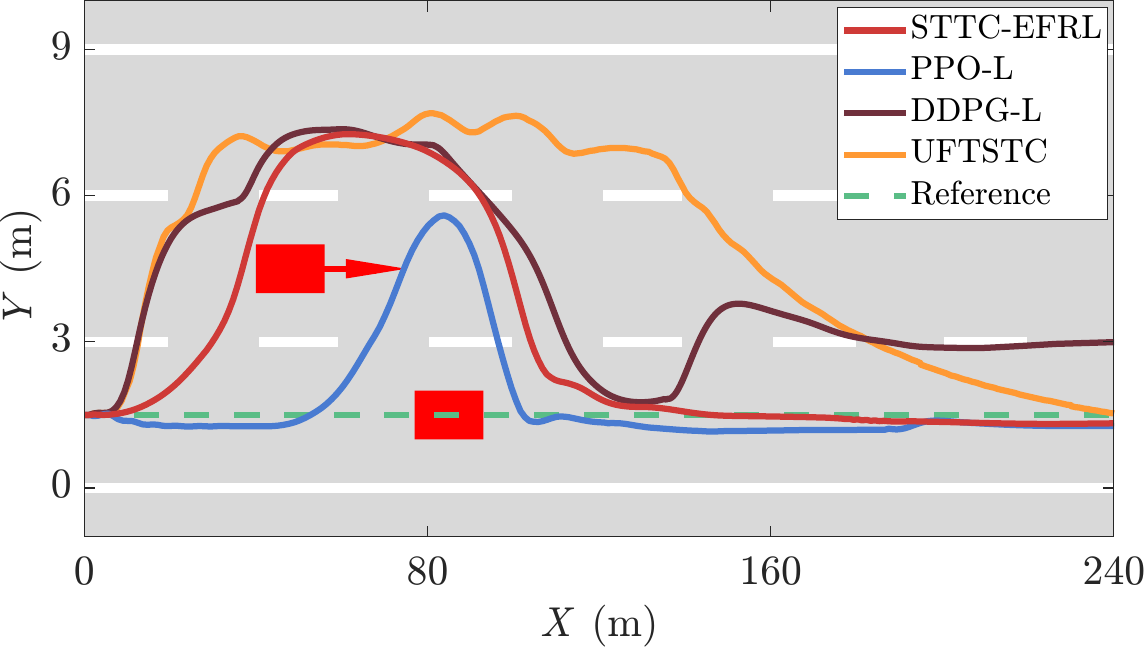}
    \label{fig:x_y_scene_iv}
    }
    \hfill
    \subfloat[]{
    \includegraphics[width=0.49\linewidth]{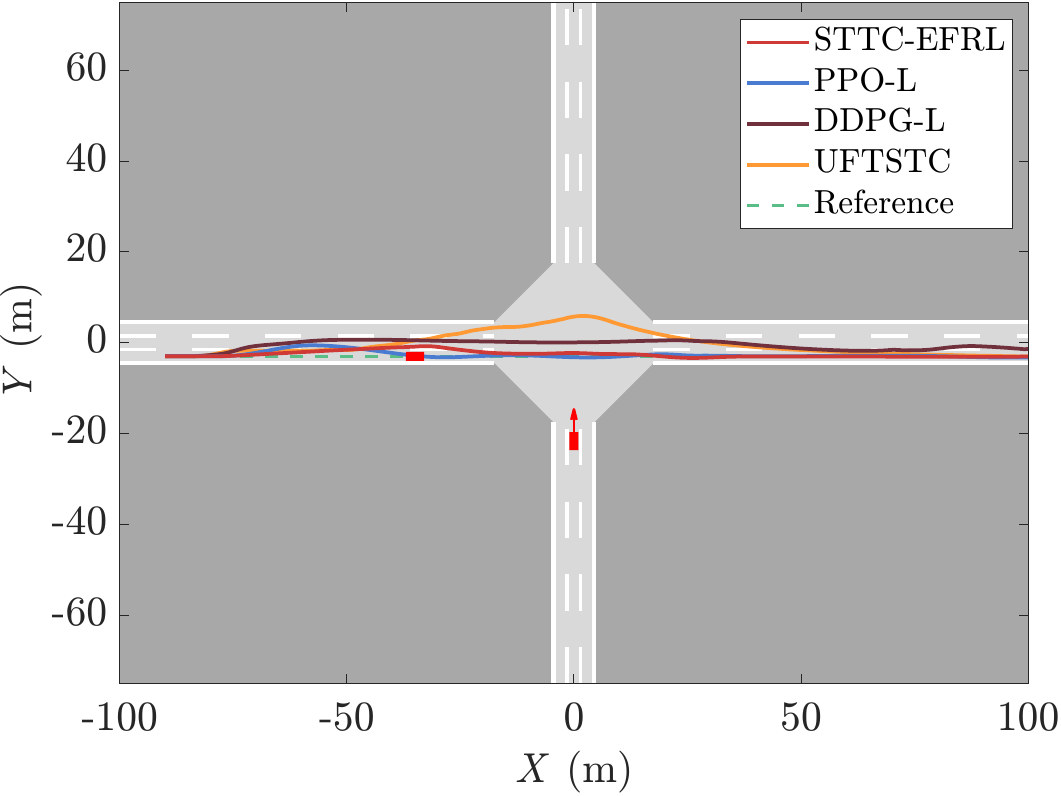}
    \label{fig:x_y_scene_v}
    }
    \subfloat[]{
    \includegraphics[width=0.49\linewidth]{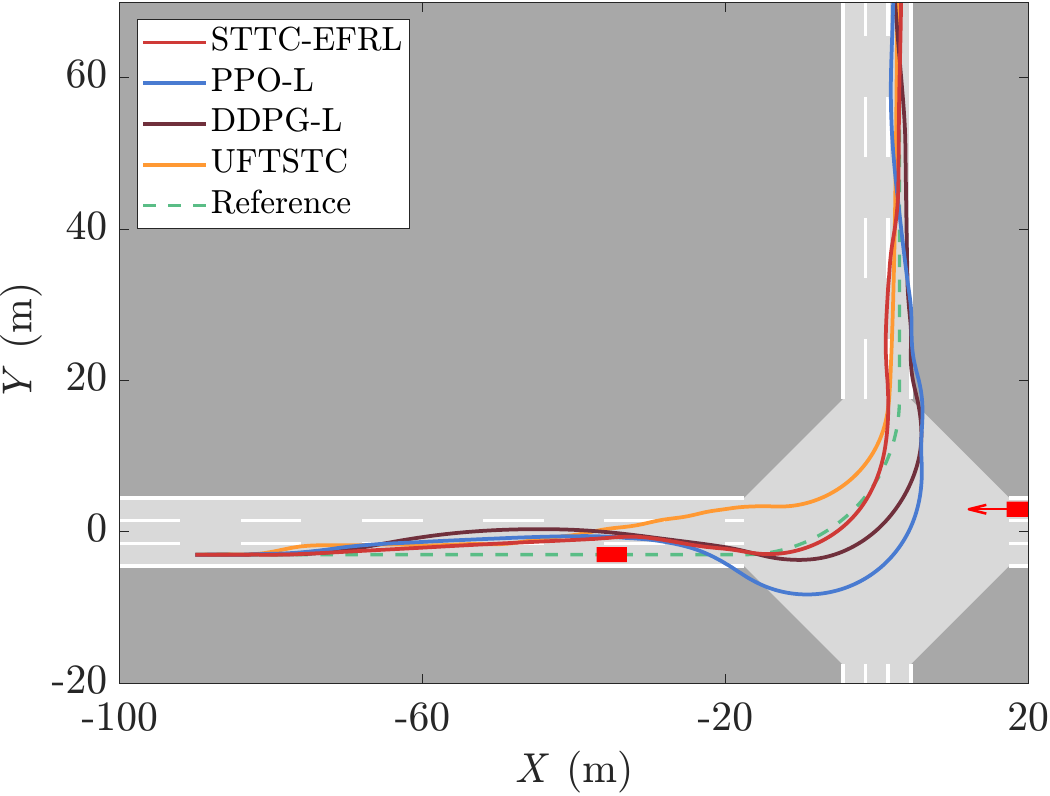}
    \label{fig:x_y_scene_vi}
    }
    \caption{The tracking trajectories in the x–y coordinate system of Scenario I in (a), Scenario II in (b), Scenario III in (c), Scenario IV in (d), Scenario V in (e) and Scenario VI in (f). The red squares denote the static and moving obstacle vehicles.}
    \label{fig:x_y}
\end{figure}

\begin{figure}[tbp]
    \centering
    \subfloat[]{
    \includegraphics[width=0.45\linewidth]{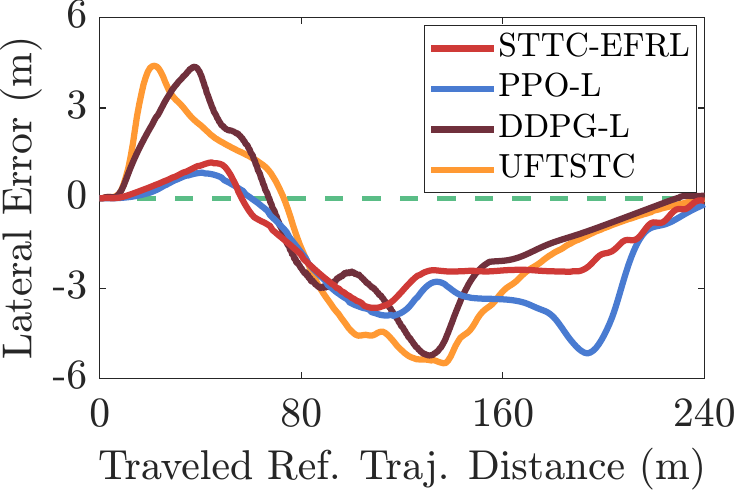}
    \label{fig:y_error_scene_i}
    }
    \subfloat[]{
    \includegraphics[width=0.45\linewidth]{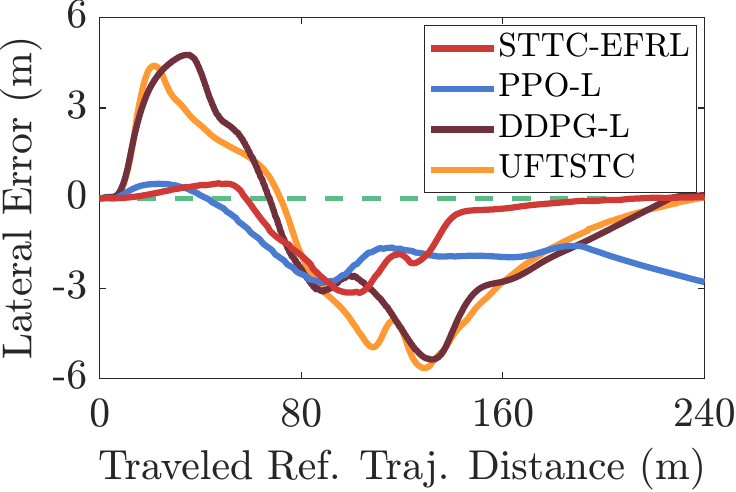}
    \label{fig:y_error_scene_ii}
    }
    \hfill
    \subfloat[]{
    \includegraphics[width=0.45\linewidth]{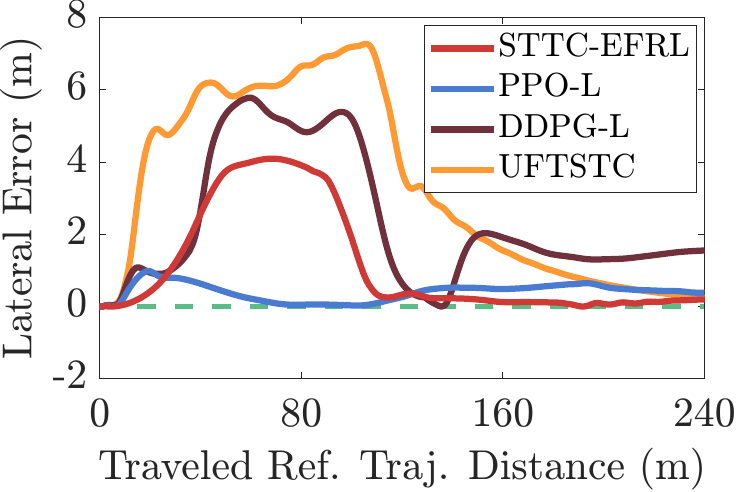}
    \label{fig:y_error_scene_iii}
    }
    \subfloat[]{
    \includegraphics[width=0.45\linewidth]{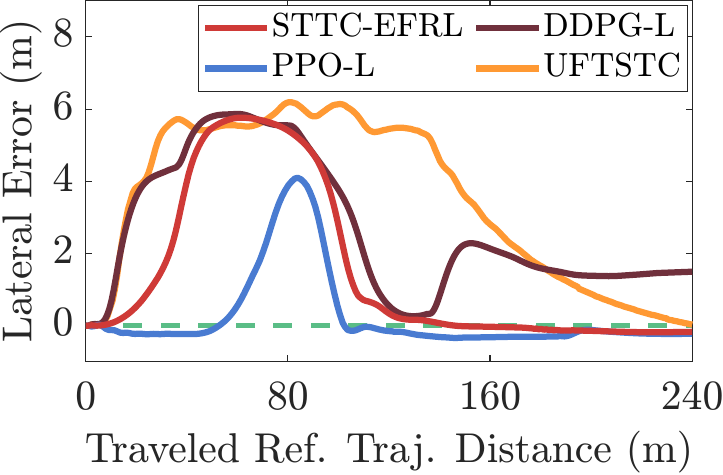}
    \label{fig:y_error_scene_iv}
    }
    \hfill
    \subfloat[]{
    \includegraphics[width=0.45\linewidth]{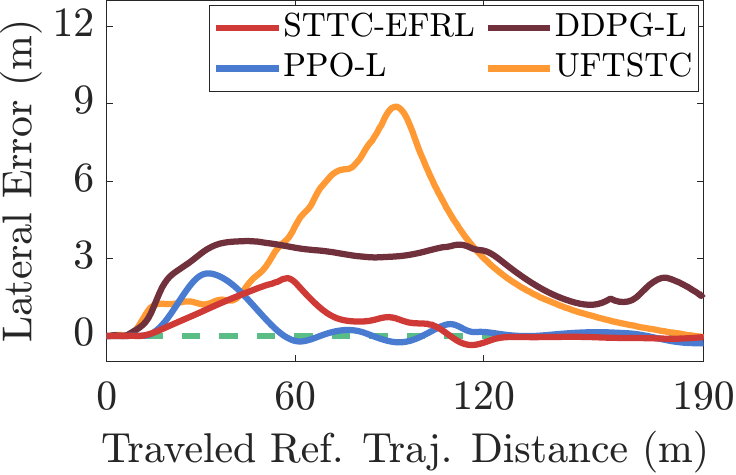}
    \label{fig:y_error_scene_v}
    }
    \subfloat[]{
    \includegraphics[width=0.45\linewidth]{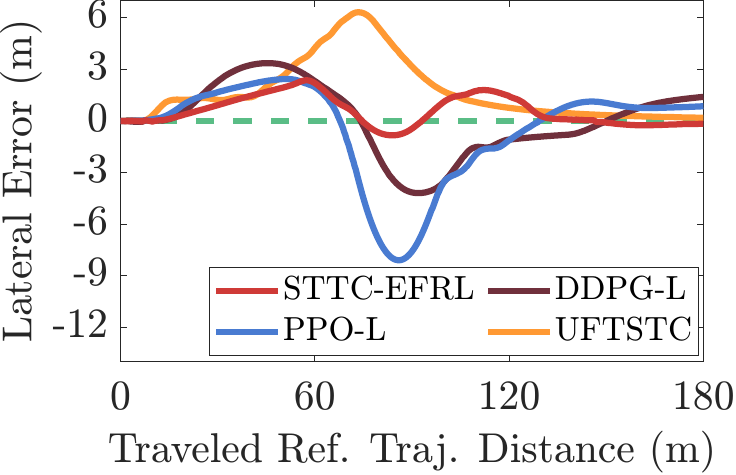}
    \label{fig:y_error_scene_vi}
    }
    \caption{The lateral errors w.r.t. the traveled reference trajectory distance of Scenario I in (a), Scenario II in (b), Scenario III in (c), Scenario IV in (d), Scenario V in (e) and Scenario VI in (f).}
    \label{fig:y_error}
\end{figure}

\begin{figure*}[htbp]
    \centering
    \subfloat[]{
    \includegraphics[width=0.9\linewidth]{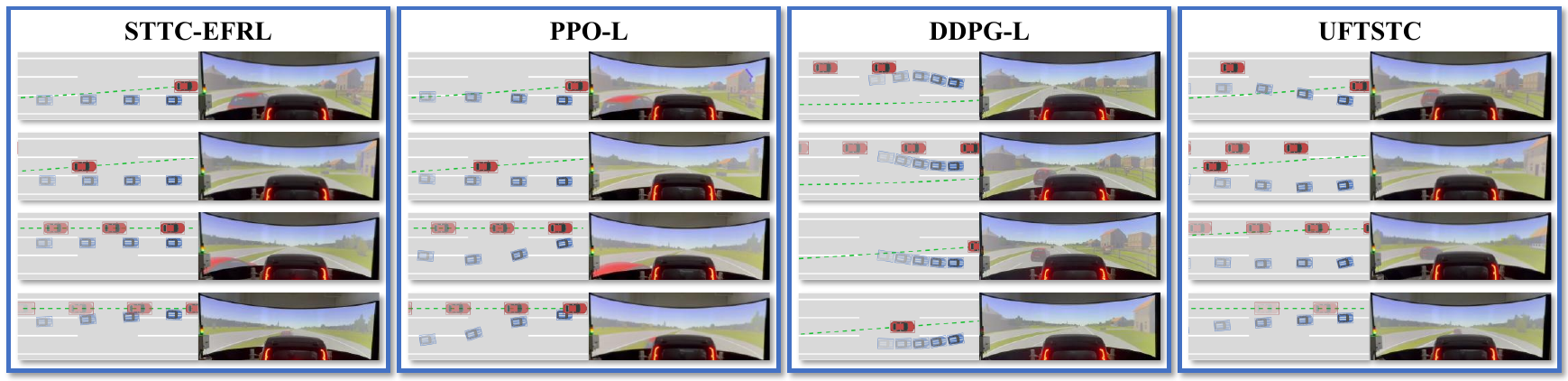}
    \label{fig:video_frame_scene_i}
    }
    \hfill
    \subfloat[]{
    \includegraphics[width=0.9\linewidth]{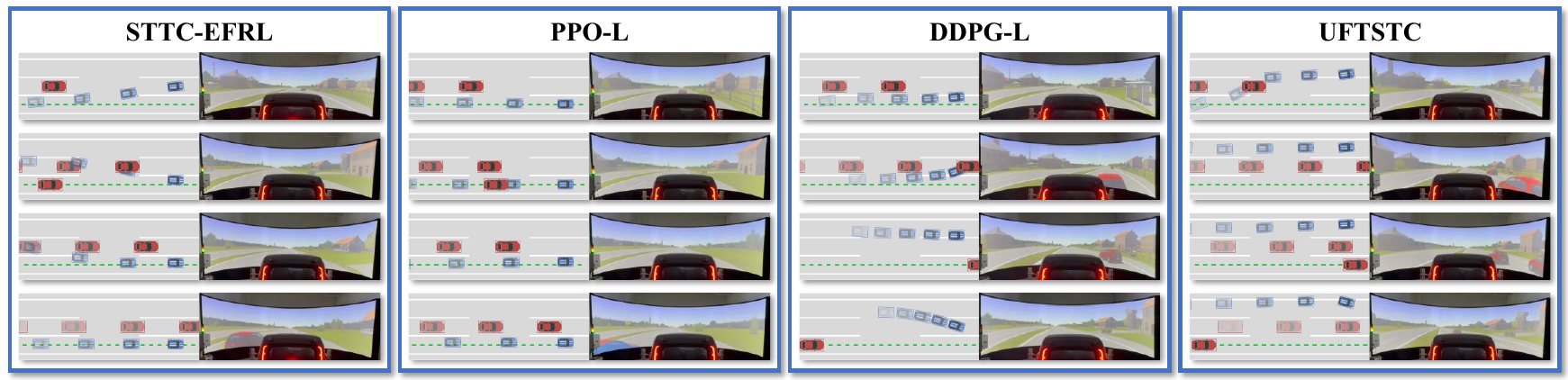}
    \label{fig:video_frame_scene_iii}
    }
    \hfill
    \subfloat[]{
    \includegraphics[width=0.9\linewidth]{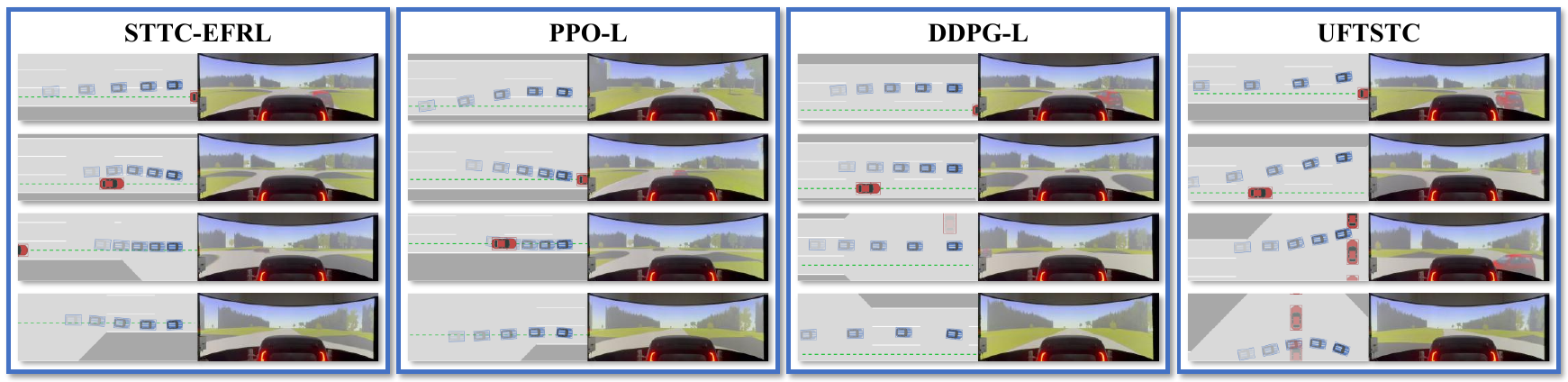}
    \label{fig:video_frame_scene_v}
    }
    \caption{Key intermediate frames of ViL experiments of Scenario I in (a), Scenario III in (b) and Scenario V in (c). Transparent primitives represent vehicles at historical time steps.}
    \label{fig:video_frame}
\end{figure*}

The settings of Scenarios I to VI, including the ego vehicle starting positions, the reference trajectory and the obstacle trajectories, are fixed. During the evaluations, all DRL-based methods execute deterministic policies, and the parameters of the UFTSTC are constant. 

\begin{figure}[H]
    \centering
    \includegraphics[width=0.95\linewidth]{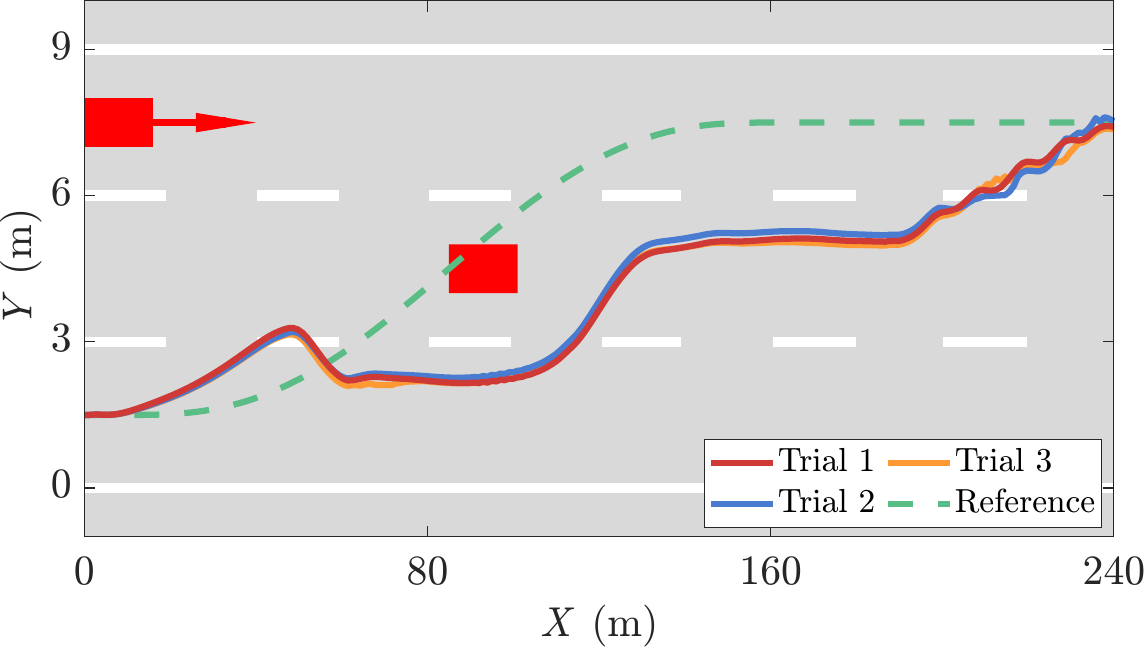}
    \caption{Three repeated ViL trajectories of STTC-EFRL in Scenario I under fixed experimental conditions.}
    \label{fig:trials}
\end{figure}

\begin{table}[H]
    \begin{center}
    \caption{Mean and standard deviation of absolute lateral trajectory differences for STTC-EFRL in Scenario I.}
    \label{tab:trials}
    \begin{tabular}{c c c}
    \noalign{\hrule height 1pt}
    \textbf{Comparison} & \textbf{Mean (m)} & \textbf{Std. (m)} \\
    \hline
    Trial 1 vs. Trial 2 & $0.0606$ & $0.0557$ \\
    Trial 1 vs. Trial 3 & $0.0466$ & $0.0618$ \\
    Trial 2 vs. Trial 3 & $0.0935$ & $0.0831$ \\
    \noalign{\hrule height 1pt}
    \end{tabular}
    \end{center}
\end{table}

To further evaluate the repeatability of STTC-EFRL under fixed experimental conditions, three repeated ViL trials were conducted in Scenario I during the experimental setup and debugging stage. The recorded trajectories are compared in Fig. \ref{fig:trials}, and the corresponding statistical results are summarized in Table \ref{tab:trials}. These results demonstrate that the three trajectories exhibit close agreement, and demonstrate good trajectory repeatability of STTC-EFRL under the evaluated conditions.

For the comparative performance evaluation, one trial is conducted for each algorithm in each scenario. The overall performance is summarized in Table \ref{tab:ViL_results}, with the ego vehicle’s COM position curves plotted in Fig. \ref{fig:x_y} and lateral error curves in Fig. \ref{fig:y_error}. We further present key intermediate frames in Fig. \ref{fig:video_frame}. The ViL results show that STTC-EFRL achieves satisfactory tracking accuracy while satisfying the evaluated safety constraints in all six predefined scenarios. Compared with the considered baseline controllers, STTC-EFRL exhibits favorable overall performance in tracking accuracy and safety compliance across the evaluated scenarios. Conversely, both PPO-L and DDPG-L controllers exhibit unstable tracking and unreliable obstacle avoidance. In Scenario II, the PPO-L controlled ego vehicle fails to follow the reference trajectory by the end of the test. Furthermore, it struggles with highly dynamic, complex obstacles as seen in Scenarios I, III, V, and VI, and even fails to avoid static vehicles, merely following a straight reference trajectory in Scenarios III and V. In Scenario VI, PPO-L also violates the road boundary constraints. We attribute these critical failures to the severe training instability previously analyzed. 

The DDPG-L controller behaves similarly to PPO-L due to their shared Lagrangian framework, but DDPG-L exhibits a slower training speed as shown in Fig. \ref{fig:training_straight:unsafe_frac}, \ref{fig:training_intersection:cost}, and \ref{fig:training_intersection:unsafe_frac}, resulting in even poorer performance. Specifically, across Scenarios III and IV, the DDPG-L controller not only diverges from the spatial reference trajectory towards the end of the simulation but also fails to maintain the desired velocity profile, leading to substantially higher control costs. We attribute this slow training speed to its deterministic policy derivation, which relies on manually designed noise for exploration and is thus less suited for the complex STT task compared to the stochastic, rapid exploration of PPO. 

Finally, the non-learning-based UFTSTC controller struggles to simultaneously steer the highly dynamic ego vehicle, avoid obstacles, and track the reference trajectory across all scenarios, leading to significant tracking errors. It violates road boundary constraints in Scenario III and, in Scenario V, incorrectly attempts to avoid the moving vehicle by steering in the same direction as the obstacle's trajectory. We attribute these undesirable behaviors to the UFTSTC's inability to gracefully handle the inherent high dynamics and control sensitivity of high-speed vehicle motion.

\begin{remark}
\label{rmk:model_mismatch_in_deployment}
As a high-frequency closed-loop DRL controller, STTC-EFRL continuously updates the control input based on the observed vehicle state, enabling feedback correction of tracking errors during execution. The effectiveness of this closed-loop mechanism is supported by the ViL experimental results.The ViL hardware setup involves lower-level execution dynamics that are not fully captured by the idealized 3-DOF model, yet the controller maintains satisfactory tracking and safety performance under the evaluated experimental conditions.For full-scale physical deployments with more severe model mismatches, the framework can be comprehensively adapted via robust training and data-driven calibration. On the one hand, domain randomization can be introduced during the DRL training phase by randomizing key parameters of the 3-DOF model, such as vehicle mass and cornering stiffness. The agent can then learn a generalized policy robust to a wide range of real-world dynamics. On the other hand, real-world data can be collected to perform system parameters correction or train a residual dynamics model. The sim-to-real gap can then be progressively minimized without altering the core epigraph-form algorithmic structure.
\end{remark}

\section{Conclusion}
\label{sec:conclusion}
This paper proposes a novel integrated approach for STT of CAVs to optimize tracking accuracy under safety constraints. We first transform the constrained OCP of STT into a min-max optimization formulation to introduce the epigraph. Subsequently, the STTC-EFRL algorithm is developed to solve this reformulated problem. By introducing a monotonically decreasing epigraph variable, the algorithm guides the policy gradient descent without magnitude scaling during training, thereby enhancing training stability. Theoretical analysis establishes stability guarantees for the ego vehicle under the learned policy in the unconstrained condition. The ViL experiments demonstrate that STTC-EFRL achieves favorable overall tracking and safety performance compared with the considered Lagrangian-based DRL and non-learning-based baseline methods under the six evaluated scenarios.

Despite these advancements, key limitations remain regarding real-world physical deployment. First, the current framework assumes ideal continuous V2X communication. In severe real-world environments, significant packet loss or communication delays can occur. The missing obstacle state information compromises the safety and tracking accuracy of the STTC-EFRL controller. To address this issue, our future work will design a communication-robust DRL method utilizing techniques like augmenting the state space with observation delays or implementing multiple tracking critic NNs under unknown channel qualities. Second, the current primitive evaluations rely on a nominal dynamic model. The framework lacks formal mathematical robustness when the system experiences unpredictable real-world noises, specifically unmodeled matched disturbances on the steering angle. To address this limitation, our future work will integrate CBFs into the STTC-EFRL architecture. By leveraging the theoretical framework of Input-to-State Safety (ISSf), we will establish rigorous mathematical bounds. These bounds will guarantee safety constraint satisfaction even in the presence of bounded matched disturbances on the control input. Third, regarding statistical evaluation, the reliance on a single training seed limits the assessment of sensitivity to initialization, and conducting only one comparative ViL trial per scenario per algorithm restricts comprehensive statistical validation. Future studies will incorporate multiple training seeds and repeated experimental trials to provide a more rigorous statistical evaluation. Such evaluations would provide a more comprehensive assessment of the consistency and robustness of the proposed method under diverse experimental conditions.

\bibliographystyle{IEEEtran}
\bibliography{ref}

@article{zhang2021secure,
  title={Secure adaptive trajectory tracking control for nonlinear robot systems under multiple dynamic obstacles: Safety barrier certificates},
  author={Zhang, Lili and Yang, Guang-Hong},
  journal={IEEE Transactions on Industrial Electronics},
  volume={69},
  number={11},
  pages={11549--11559},
  year={2021},
  publisher={IEEE}
}

@inproceedings{hsu2021safety,
  title={Safety and Liveness Guarantees through Reach-Avoid Reinforcement Learning},
  author={Hsu, Kai Chieh and Rubies-Royo, Vicen{\c{c}} and Tomlin, Claire J and Fisac, Jaime F},
  booktitle={17th Robotics: Science and Systems, RSS 2021},
  year={2021},
  organization={MIT Press Journals}
}

@article{chen2025range,
  title={Range-only distributed safety-critical formation control based on contracting bearing estimators and control barrier functions},
  author={Chen, Jin and Li, Ming and Marcantoni, Matteo and Jayawardhana, Bayu and Wang, Yafei},
  journal={IEEE Internet of Things Journal},
  year={2025},
  publisher={IEEE}
}

@article{li2024quadrotor,
  title={Quadrotor stabilization with safety guarantees: A universal formula approach},
  author={Li, Ming and Sun, Zhiyong and Weiland, Siep},
  journal={arXiv preprint arXiv:2401.03500},
  year={2024}
}

@article{chen2017obstacle,
  title={Obstacle avoidance for low-speed autonomous vehicles with barrier function},
  author={Chen, Yuxiao and Peng, Huei and Grizzle, Jessy},
  journal={IEEE Transactions on Control Systems Technology},
  volume={26},
  number={1},
  pages={194--206},
  year={2017},
  publisher={IEEE}
}

@article{gu2024review,
  title={A review of safe reinforcement learning: Methods, theories and applications},
  author={Gu, Shangding and Yang, Long and Du, Yali and Chen, Guang and Walter, Florian and Wang, Jun and Knoll, Alois},
  journal={IEEE Transactions on Pattern Analysis and Machine Intelligence},
  year={2024},
  publisher={IEEE}
}

@article{gu2023safe,
  title={Safe multi-agent reinforcement learning for multi-robot control},
  author={Gu, Shangding and Kuba, Jakub Grudzien and Chen, Yuanpei and Du, Yali and Yang, Long and Knoll, Alois and Yang, Yaodong},
  journal={Artificial Intelligence},
  volume={319},
  pages={103905},
  year={2023},
  publisher={Elsevier}
}

@article{zhang2021safe,
  title={Safe reinforcement learning with stability guarantee for motion planning of autonomous vehicles},
  author={Zhang, Lixian and Zhang, Ruixian and Wu, Tong and Weng, Rui and Han, Minghao and Zhao, Ye},
  journal={IEEE transactions on neural networks and learning systems},
  volume={32},
  number={12},
  pages={5435--5444},
  year={2021},
  publisher={IEEE}
}

@article{gu2023safevehicle,
  title={Safe-state enhancement method for autonomous driving via direct hierarchical reinforcement learning},
  author={Gu, Ziqing and Gao, Lingping and Ma, Haitong and Li, Shengbo Eben and Zheng, Sifa and Jing, Wei and Chen, Junbo},
  journal={IEEE Transactions on Intelligent Transportation Systems},
  volume={24},
  number={9},
  pages={9966--9983},
  year={2023},
  publisher={IEEE}
}

@article{hou2024safe,
  title={Safe reinforcement learning for autonomous driving by using disturbance-observer-based control barrier functions},
  author={Hou, Zhengyu and Liu, Wenjun and Knoll, Alois},
  journal={IEEE Transactions on Intelligent Vehicles},
  year={2024},
  publisher={IEEE}
}

@article{altarovici2013general,
  title={A general Hamilton-Jacobi framework for non-linear state-constrained control problems},
  author={Altarovici, Albert and Bokanowski, Olivier and Zidani, Hasnaa},
  journal={ESAIM: Control, Optimisation and Calculus of Variations},
  volume={19},
  number={2},
  pages={337--357},
  year={2013},
  publisher={EDP Sciences}
}

@article{lin2025a,
  author={Lin, Zhihao and Tian, Zhen and Lan, Jianglin and Zhang, Qi and Ye, Ziyang and Zhuang, Hanyang and Zhao, Xianxian},
  journal={IEEE Transactions on Intelligent Transportation Systems}, 
  title={A Conflicts-Free, Speed-Lossless KAN-Based Reinforcement Learning Decision System for Interactive Driving in Roundabouts}, 
  year={2025},
  volume={26},
  number={10},
  pages={16377-16390},
  doi={10.1109/TITS.2025.3578279}}

@article{li2023safe,
  author   = {Li, Daofei and Liu, Ao and Pan, Hao and Chen, Wentao},
  title    = {Safe, Efficient and Socially-Compatible Decision of Automated Vehicles: A Case Study of Unsignalized Intersection Driving},
  journal  = {Automotive Innovation},
  year     = {2023},
  volume   = {6},
  number   = {2},
  pages    = {281--296},
  issn     = {2522-8765},
  doi      = {10.1007/s42154-023-00219-2},
  url      = {https://doi.org/10.1007/s42154-023-00219-2},
}

@article{liu2025lane,
  title={Lane-change trajectory planning method for connected autonomous vehicles in freeway ramp merging areas},
  author={Liu, Yongtao and Zhu, Yichen and Kang, Haoyu and Liu, Chuanpan and Hou, Kaiwen and Song, Dongdong and Cheng, Qingwei},
  journal={Computer-Aided Civil and Infrastructure Engineering},
  year={2025},
  publisher={Wiley Online Library}
}

@article{reza2021integrated,
  author={Hajiloo, Reza and Abroshan, Mehdi and Khajepour, Amir and Kasaiezadeh, Alireza and Chen, Shih-Ken},
  journal={IEEE Transactions on Intelligent Transportation Systems}, 
  title={Integrated Steering and Differential Braking for Emergency Collision Avoidance in Autonomous Vehicles}, 
  year={2021},
  volume={22},
  number={5},
  pages={3167-3178},
  doi={10.1109/TITS.2020.2984210}}

@book{bazaraa2011linear,
  title={Linear programming and network flows},
  author={Bazaraa, Mokhtar S and Jarvis, John J and Sherali, Hanif D},
  pages={75--81},
  year={2011},
  publisher={John Wiley \& Sons}
}

@incollection{grune2016nonlinear,
  title={Discrete Time and Sampled Data Systems},
  author={Gr{\"u}ne, Lars and Pannek, J{\"u}rgen},
  booktitle={Nonlinear model predictive control: Theory and algorithms},
  pages={29--36},
  year={2016},
  publisher={Springer}
}

@article{ames2016control,
  title={Control barrier function based quadratic programs for safety critical systems},
  author={Ames, Aaron D and Xu, Xiangru and Grizzle, Jessy W and Tabuada, Paulo},
  journal={IEEE Transactions on Automatic Control},
  volume={62},
  number={8},
  pages={3861--3876},
  year={2016},
  publisher={IEEE}
}

@article{tracy2022differentiable,
  title={Differentiable collision detection for a set of convex primitives},
  author={Tracy, Kevin and Howell, Taylor A and Manchester, Zachary},
  journal={arXiv preprint arXiv:2207.00669},
  year={2022}
}

@article{schulman2017proximal,
  title={Proximal policy optimization algorithms},
  author={Schulman, John and Wolski, Filip and Dhariwal, Prafulla and Radford, Alec and Klimov, Oleg},
  journal={arXiv preprint arXiv:1707.06347},
  year={2017}
}

@inproceedings{so23solving,
  author={Oswin So and Chuchu Fan},
  title={Solving Stabilize-Avoid via Epigraph Form Optimal Control using Deep Reinforcement Learning},
  year={2023},
  cdate={1672531200000},
  url={https://doi.org/10.15607/RSS.2023.XIX.085},
  booktitle={Robotics: Science and Systems},
}

@inproceedings{kong2015kinematic,
  title={Kinematic and dynamic vehicle models for autonomous driving control design},
  author={Kong, Jason and Pfeiffer, Mark and Schildbach, Georg and Borrelli, Francesco},
  booktitle={2015 IEEE intelligent vehicles symposium (IV)},
  pages={1094--1099},
  year={2015},
  organization={IEEE}
}

@article{schulman2015high,
  title={High-dimensional continuous control using generalized advantage estimation},
  author={Schulman, John and Moritz, Philipp and Levine, Sergey and Jordan, Michael and Abbeel, Pieter},
  journal={arXiv preprint arXiv:1506.02438},
  year={2015}
}

@article{han2024uniform,
  title={Uniform finite time safe path tracking control for obstacle avoidance of autonomous vehicle via barrier function approach},
  author={Han, Jinheng and Zhang, Junzhi and He, Chengkun and Lv, Chen and Wei, Henglai and Zhang, Junfeng and Zhao, Shiyue},
  journal={IEEE Transactions on Intelligent Transportation Systems},
  volume={25},
  number={8},
  pages={9512--9523},
  year={2024},
  publisher={IEEE}
}

@inproceedings{nayak2023scalable,
  title={Scalable multi-agent reinforcement learning through intelligent information aggregation},
  author={Nayak, Siddharth and Choi, Kenneth and Ding, Wenqi and Dolan, Sydney and Gopalakrishnan, Karthik and Balakrishnan, Hamsa},
  booktitle={International conference on machine learning},
  pages={25817--25833},
  year={2023},
  organization={PMLR}
}

@article{erlien2016shared,
  author={Erlien, Stephen M. and Fujita, Susumu and Gerdes, Joseph Christian},
  journal={IEEE Transactions on Intelligent Transportation Systems}, 
  title={Shared Steering Control Using Safe Envelopes for Obstacle Avoidance and Vehicle Stability}, 
  year={2016},
  volume={17},
  number={2},
  pages={441-451},
  doi={10.1109/TITS.2015.2453404}}

@article{zhang2021human,
  author={Zhang, Lin and Chen, Hong and Huang, Yanjun and Wang, Ping and Guo, Konghui},
  journal={IEEE Transactions on Vehicular Technology}, 
  title={Human-Centered Torque Vectoring Control for Distributed Drive Electric Vehicle Considering Driving Characteristics}, 
  year={2021},
  volume={70},
  number={8},
  pages={7386-7399},
  doi={10.1109/TVT.2021.3087476}}

@article{paden2016survey,
  title={A survey of motion planning and control techniques for self-driving urban vehicles},
  author={Paden, Brian and {\v{C}}{\'a}p, Michal and Yong, Sze Zheng and Yershov, Dmitry and Frazzoli, Emilio},
  journal={IEEE Transactions on intelligent vehicles},
  volume={1},
  number={1},
  pages={33--55},
  year={2016},
  publisher={IEEE}
}

@ARTICLE{chen2026,
  author={Chen, Jin and Li, Lidong and Wang, Yafei and Zhang, Qi and Chen, Hailong and Zhou, Jiaming and Wu, Mingyu},
  journal={IEEE Transactions on Industrial Informatics}, 
  title={Direct Data-Driven Trajectory Tracking Control for Autonomous Vehicles via Algebraic Regulator Under Limited Data}, 
  year={2026},
  volume={},
  number={},
  pages={1-12},
  doi={10.1109/TII.2026.3658768}}

@article{chen2022reinforcement,
  title={Reinforcement learning for selective key applications in power systems: Recent advances and future challenges},
  author={Chen, Xin and Qu, Guannan and Tang, Yujie and Low, Steven and Li, Na},
  journal={IEEE Transactions on Smart Grid},
  volume={13},
  number={4},
  pages={2935--2958},
  year={2022},
  publisher={IEEE}
}

@article{funada2025collision,
  author={Funada, Riku and Nishimoto, Koju and Ibuki, Tatsuya and Sampei, Mitsuji},
  journal={IEEE Transactions on Control Systems Technology}, 
  title={Collision Avoidance for Ellipsoidal Rigid Bodies With Control Barrier Functions Designed From Rotating Supporting Hyperplanes}, 
  year={2025},
  volume={33},
  number={1},
  pages={148-164},
  doi={10.1109/TCST.2024.3467809}}

@article{singletary2022safety,
  author={Singletary, Andrew and Guffey, William and Molnar, Tamas G. and Sinnet, Ryan and Ames, Aaron D.},
  journal={IEEE Robotics and Automation Letters}, 
  title={Safety-Critical Manipulation for Collision-Free Food Preparation}, 
  year={2022},
  volume={7},
  number={4},
  pages={10954-10961},
  doi={10.1109/LRA.2022.3192634}}

@article{zha2025heterogeneous,
  title={Heterogeneous multiscale cooperative perception for connected autonomous vehicles via V2X interaction},
  author={Zha, Yuanyuan and Shangguan, Wei and Chen, Junjie and Chai, Linguo and Qiu, Weizhi and L{\'o}pez, Antonio M},
  journal={IEEE Internet of Things Journal},
  year={2025},
  publisher={IEEE}
}

@article{huang2022efficient,
  title={Efficient deep reinforcement learning with imitative expert priors for autonomous driving},
  author={Huang, Zhiyu and Wu, Jingda and Lv, Chen},
  journal={IEEE transactions on neural networks and learning systems},
  volume={34},
  number={10},
  pages={7391--7403},
  year={2022},
  publisher={IEEE}
}

@article{qi2022learning,
  title={Learning-based path planning and predictive control for autonomous vehicles with low-cost positioning},
  author={Qi, Ziheng and Wang, Tong and Chen, Jian and Narang, Deepak and Wang, Yuexuan and Yang, Huayong},
  journal={IEEE Transactions on Intelligent Vehicles},
  volume={8},
  number={2},
  pages={1093--1104},
  year={2022},
  publisher={IEEE}
}

@article{zhang2025control,
  title={Control barrier function-guided deep reinforcement learning for decision-making of autonomous vehicle at on-ramp merging},
  author={Zhang, Changzhu and Dai, Lifei and Zhang, Hao and Wang, Zhuping},
  journal={IEEE Transactions on Intelligent Transportation Systems},
  year={2025},
  publisher={IEEE}
}

@inproceedings{ma2021model,
  title={Model-based constrained reinforcement learning using generalized control barrier function},
  author={Ma, Haitong and Chen, Jianyu and Eben, Shengbo and Lin, Ziyu and Guan, Yang and Ren, Yangang and Zheng, Sifa},
  booktitle={2021 IEEE/RSJ International Conference on Intelligent Robots and Systems (IROS)},
  pages={4552--4559},
  year={2021},
  organization={IEEE}
}

@article{jin2025connected,
  title={A connected-automated vehicles-based dynamic speed limit control strategy for improving safety and efficiency of freeway tunnels: an augmented lagrange safe reinforcement learning framework},
  author={Jin, Jieling and Li, Ye and Huang, Helai and Dai, Jianjun},
  journal={IEEE Internet of Things Journal},
  year={2025},
  publisher={IEEE}
}

@article{mhaskar2006stabilization,
  title={Stabilization of nonlinear systems with state and control constraints using Lyapunov-based predictive control},
  author={Mhaskar, Prashant and El-Farra, Nael H and Christofides, Panagiotis D},
  journal={Systems \& Control Letters},
  volume={55},
  number={8},
  pages={650--659},
  year={2006},
  publisher={Elsevier}
}

@ARTICLE{fei2025echelon,
  author={Hui, Fei and Zhou, Xingkai and Liu, Jiajia and Wang, Gang and Zhang, Junfei},
  journal={IEEE Internet of Things Journal}, 
  title={Echelon-Based Collaborative Resource Allocation for Platoon Communication in C-V2X Networks}, 
  year={2025},
  volume={12},
  number={20},
  pages={43002-43017},
  doi={10.1109/JIOT.2025.3594904}}

@ARTICLE{liu2026a,
  author={Liu, Ruifeng and Wang, Wenwei and Cao, Wanke and Deng, Bo and Wang, Bingbing},
  journal={IEEE Internet of Things Journal}, 
  title={A Heterogeneous Satellite-Vehicular Passive Optical Network with Fault-Tolerant Predictive Control for Large-Scale CACC Platoons}, 
  year={2026},
  volume={},
  number={},
  pages={1-1},
  doi={10.1109/JIOT.2026.3715023}}

@article{yang2026optimal,
  title={Optimal Trajectory Tracking Control of Quadrotor Autonomous Aerial Vehicle: A Predefined-time Reinforcement Learning Approach},
  author={Yang, Feisheng and Gong, Zhenyu and Ge, Shuzhi Sam},
  journal={IEEE Transactions on Automatic Control},
  year={2026},
  publisher={IEEE}
}

@article{yang2025secure,
  title={Secure containment control for multi-UAV systems by fixed-time convergent reinforcement learning},
  author={Yang, Feisheng and Gong, Zhenyu and Wei, Qinglai and Lei, Yifei},
  journal={IEEE Transactions on Cybernetics},
  volume={55},
  number={4},
  pages={1981--1994},
  year={2025},
  publisher={IEEE}
}

@article{yang2025aav,
  title={AAV weapon target assignment via state split flow deep Q-learning network with threat degree index},
  author={Yang, Feisheng and Gong, Zhenyu and Zhao, Ziming and Fang, Chengliang},
  journal={IEEE Internet of Things Journal},
  year={2025},
  publisher={IEEE}
}

@article{gong2025secure,
  title={Secure formation control of multiagent system against FDI attack using fixed-time convergent reinforcement learning},
  author={Gong, Zhenyu and Yang, Feisheng and Yuan, Yuan and Ma, Qian and Zheng, Wei Xing},
  journal={IEEE Transactions on Control of Network Systems},
  volume={12},
  number={2},
  pages={1203--1214},
  year={2025},
  publisher={IEEE}
}

\vspace{50pt}
\begin{IEEEbiography}[{\includegraphics[width=1in,height=1.25in,clip,keepaspectratio]{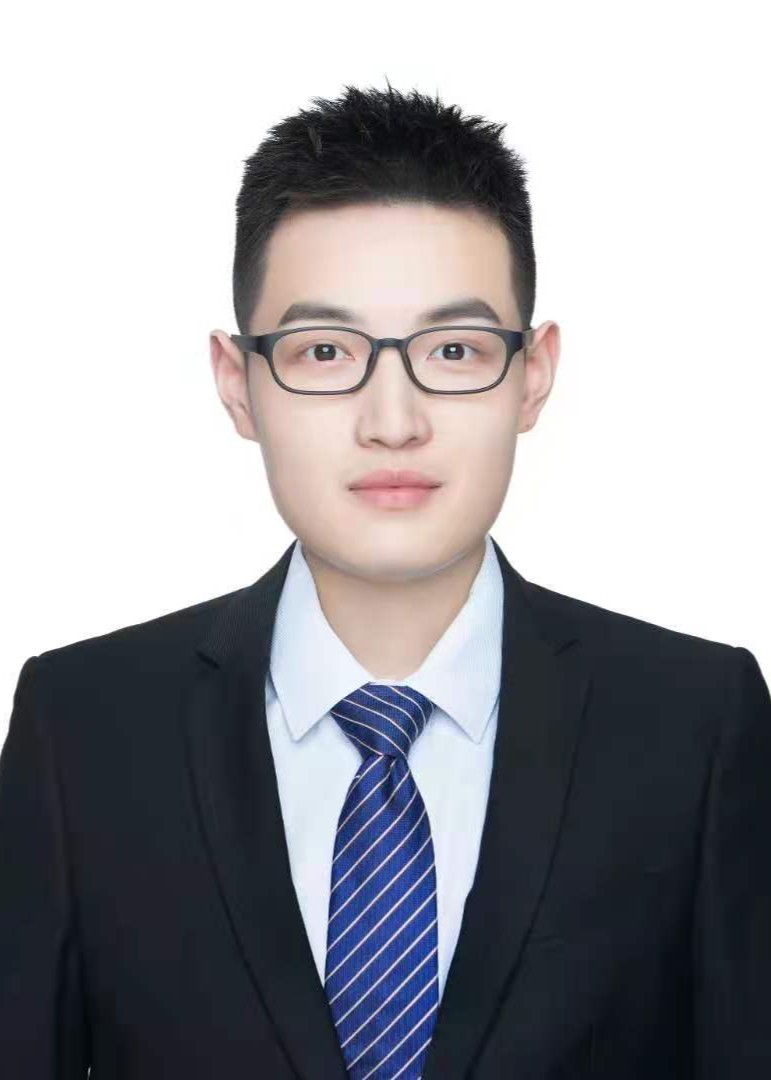}}]{Xiaoke Yang}
received the B.S. degree in mechanical engineering from Shanghai Jiao Tong University, China, in 2020. He is currently working toward the Ph.D. degree in mechanical engineering at Shanghai Jiao Tong University, Shanghai, China. His research interests focus on safety-critical control of automated vehicles and reinforcement learning.
\end{IEEEbiography}
\vspace{30pt}
\begin{IEEEbiography}[{\includegraphics[width=1in,height=1.25in,clip,keepaspectratio]{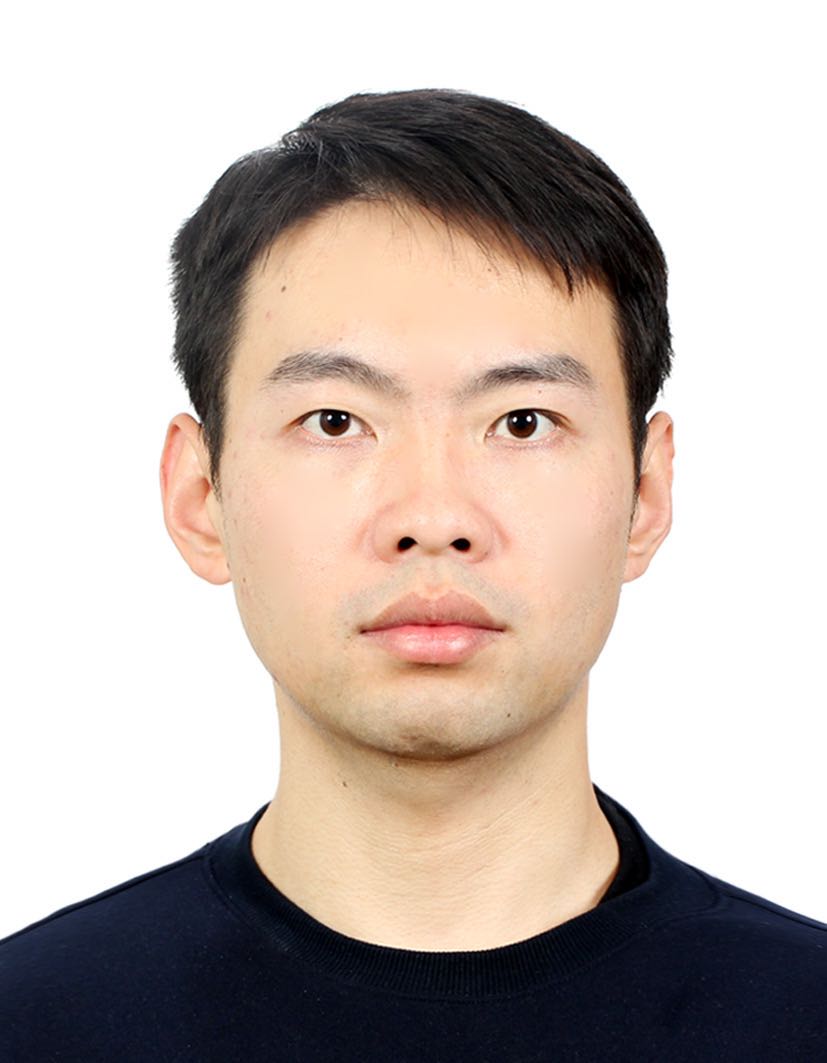}}]{Jin Chen}
received the B.S. degree in Mechanical Engineering and Automation from Jilin University, Changchun, China, in 2015, the M.S. degree in vehicle engineering from Shanghai Jiao Tong University, Shanghai, China, in 2020, and the Ph.D. degree in Automation at the University of Groningen, the Netherlands, in 2025. He is currently a Postdoctoral Researcher at the School of Mechanical Engineering, Shanghai Jiao Tong University, China. His research interests include nonlinear control, formation control, and cooperative planning and control for autonomous vehicles.
\end{IEEEbiography}
\vspace{30pt}
\begin{IEEEbiography}[{\includegraphics[width=1in,height=1.25in,clip,keepaspectratio]{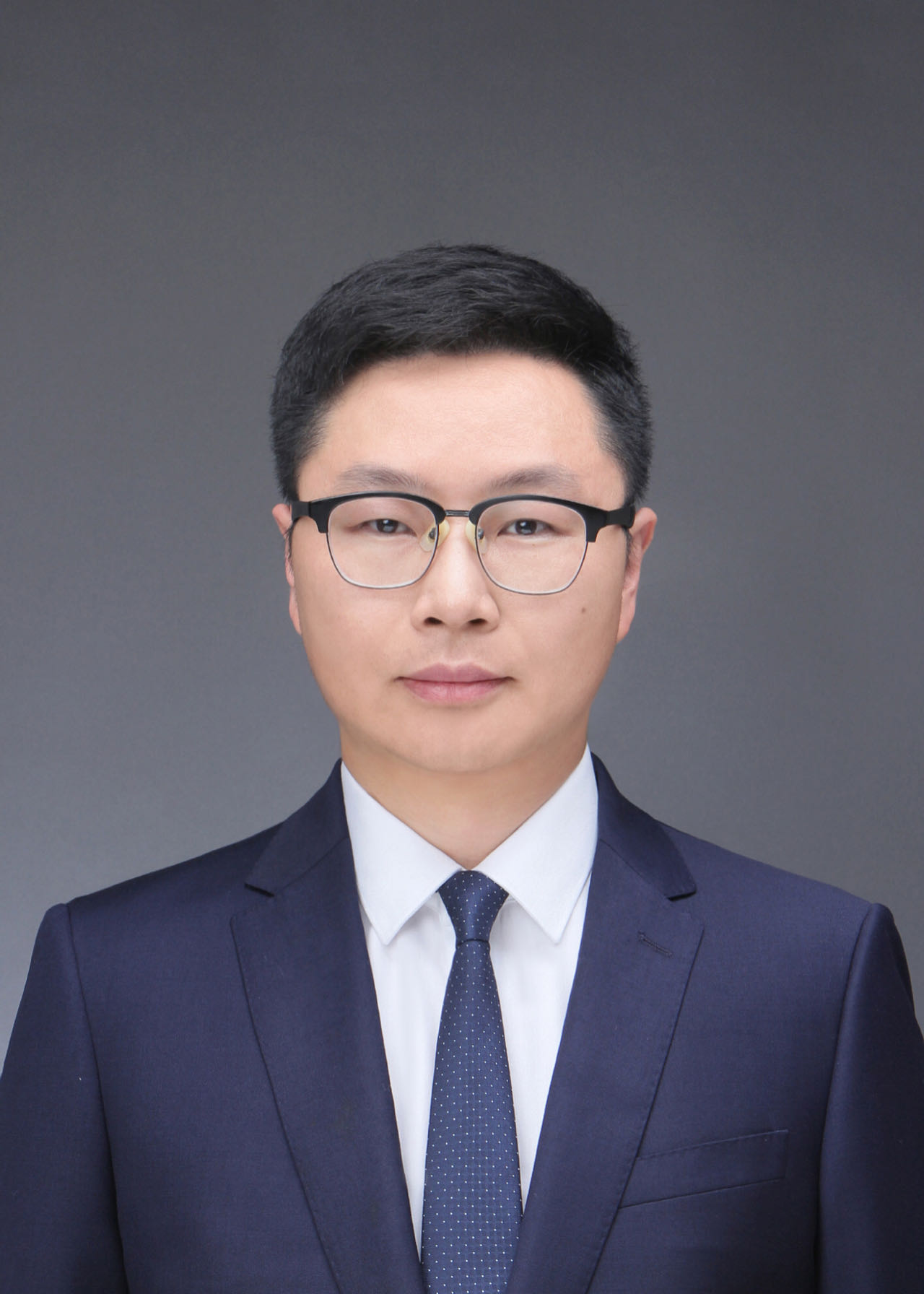}}]{Yafei Wang}
received the B.S. degree in internal combustion engine from Jilin University, Changchun, China, in 2005, the M.S. degree in vehicle engineering from Shanghai Jiao Tong University, Shanghai, China, in 2008, and the Ph.D. degree in electrical engineering from The University of Tokyo, Tokyo, Japan, in 2013. From 2013 to 2016, he was a Postdoctoral Researcher at the University of Tokyo. He is currently a Professor with the School of Mechanical Engineering, Shanghai Jiao Tong University, Shanghai, China. His research interests include state estimation and control for connected and automated vehicles.
\end{IEEEbiography}
\vspace{30pt}
\begin{IEEEbiography}[{\includegraphics[width=1in,height=1.25in,clip,keepaspectratio]{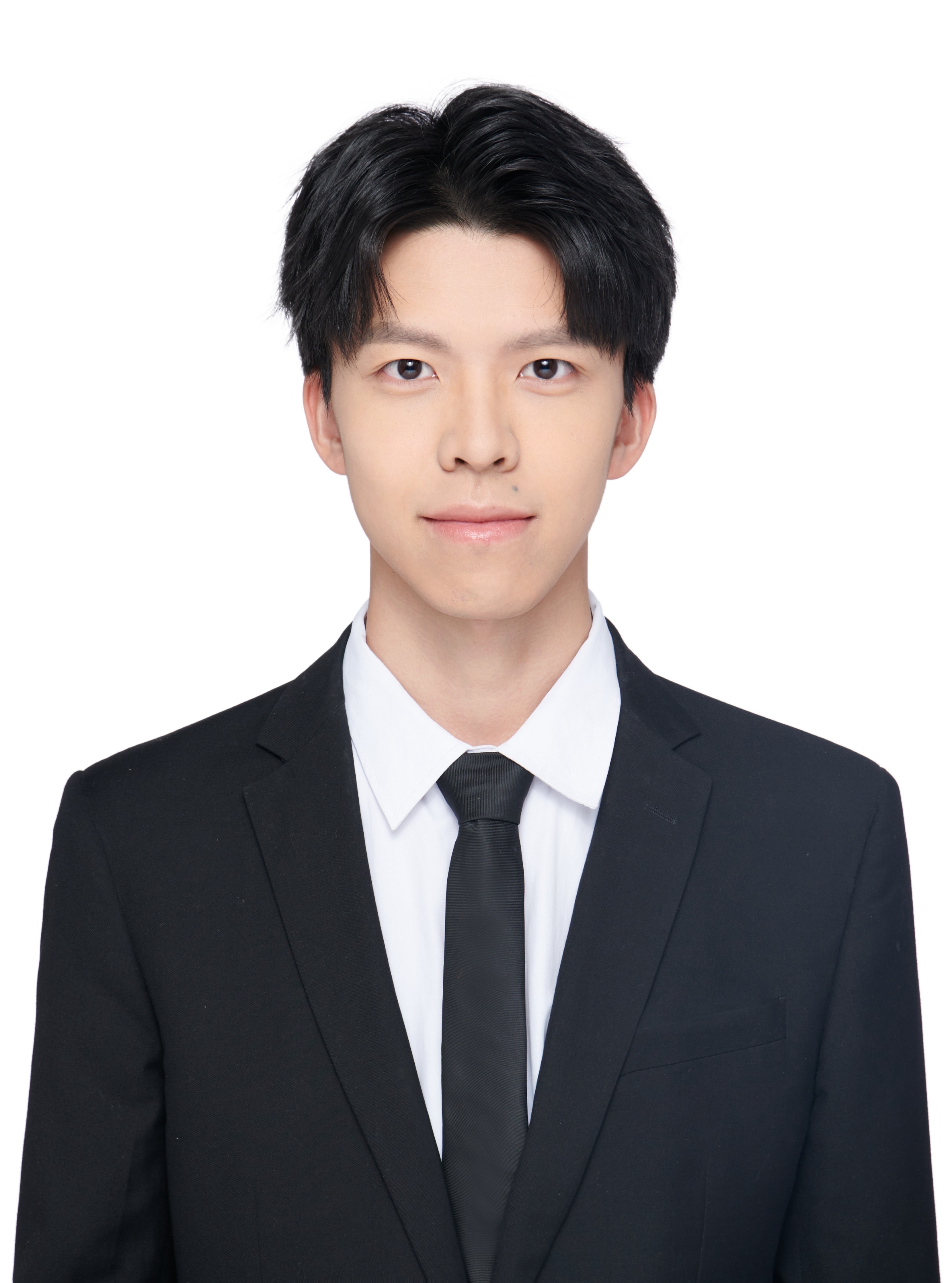}}]{Qi Zhang}
received the B.S. degree in robotics engineering from the University of Shanghai for Science and Technology, Shanghai, China, in 2024. He is currently working toward the M.Sc. degree in mechanical engineering at Shanghai Jiao Tong University, Shanghai, China. His research interests focus on planning and control for autonomous vehicles.
\end{IEEEbiography}
\vspace{30pt}
\begin{IEEEbiography}[{\includegraphics[width=1in,height=1.25in,clip,keepaspectratio]{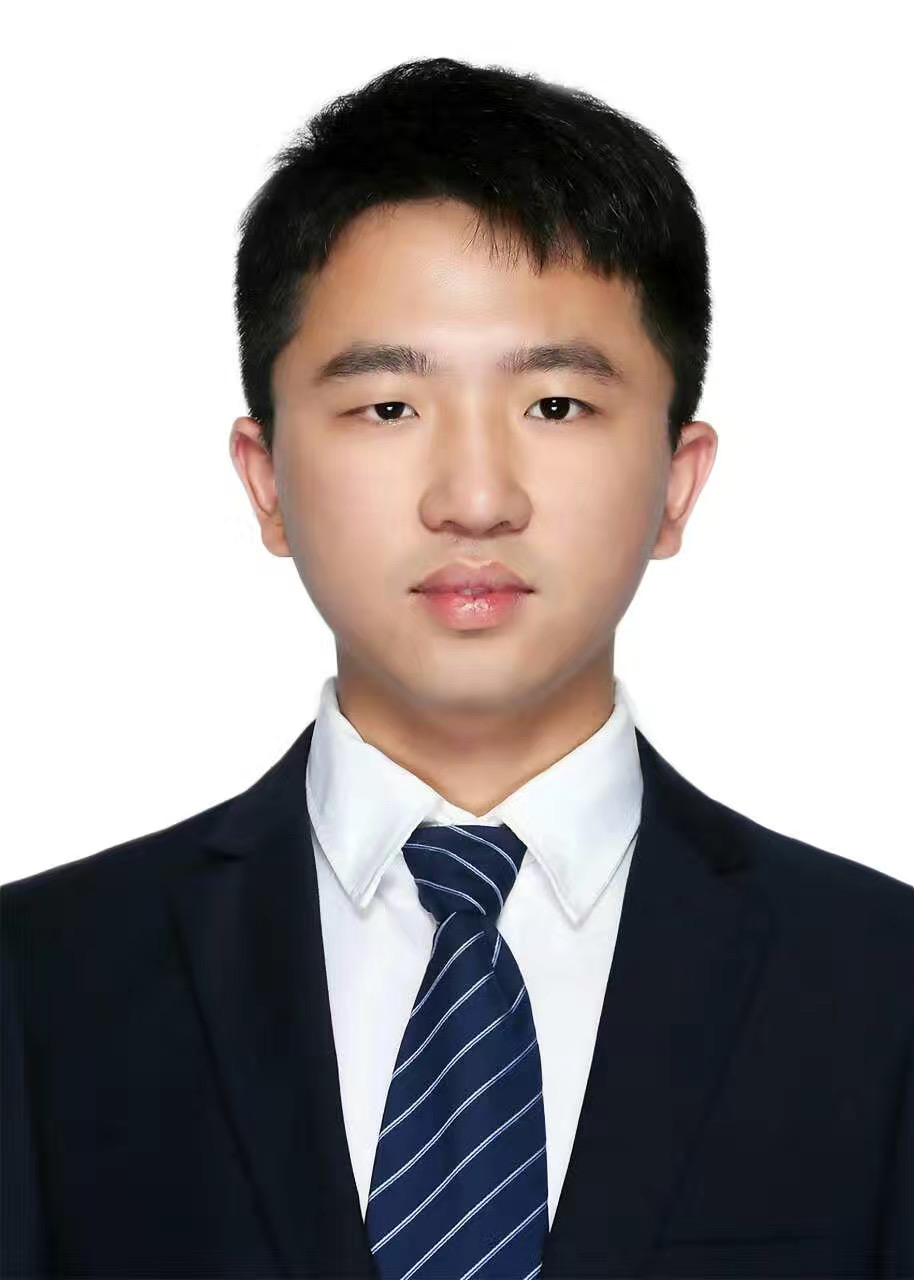}}]{Ming Li}
obtained his Ph.D. degree from the Electrical Engineering Department at Eindhoven University of Technology (TU/e) in December 2024. He is currently a Postdoctoral Researcher at KTH Royal Institute of Technology. His research interests include safety-critical control, multi-agent coordination, localization, and multi-sensor fusion.
\end{IEEEbiography}
\vspace{30pt}
\begin{IEEEbiography}[{\includegraphics[width=1in,height=1.25in,clip,keepaspectratio]{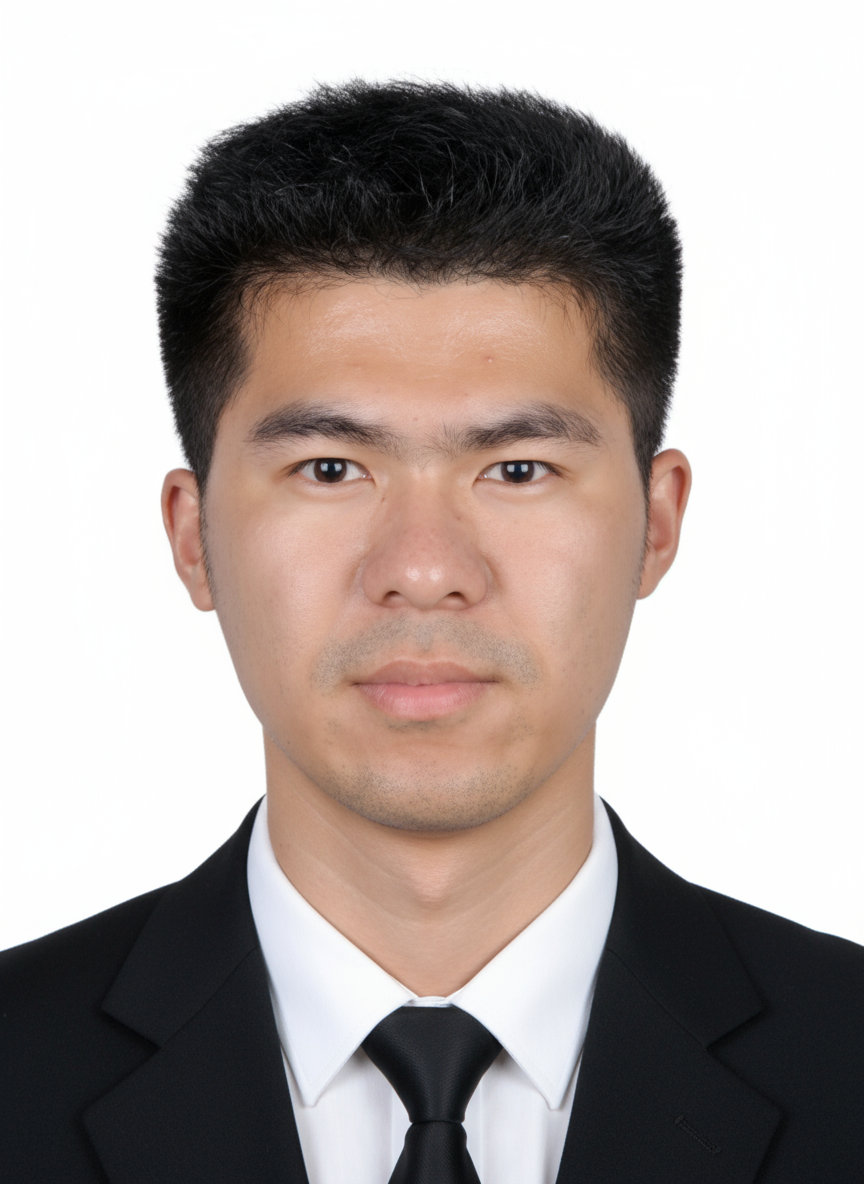}}]{Haiwen Wu} received the B.Sc. degree in automatic control from the Nanjing University of Science and Technology, Nanjing, China, in 2014, and the Ph.D. degree in electrical and electronics engineering with the Faculty of Science and Engineering, University of Groningen, Groningen, The Netherlands, in 2022. He is currently a Postdoctoral Researcher with the Hong Kong Center for Logistics Robotics and the Chinese University of Hong Kong, Hong Kong. His research interest includes the analysis and control of nonlinear systems, multiagent systems, and robotic systems.
\end{IEEEbiography}
\vspace{30pt}
\begin{IEEEbiography}[{\includegraphics[width=1in,height=1.25in,clip,keepaspectratio]{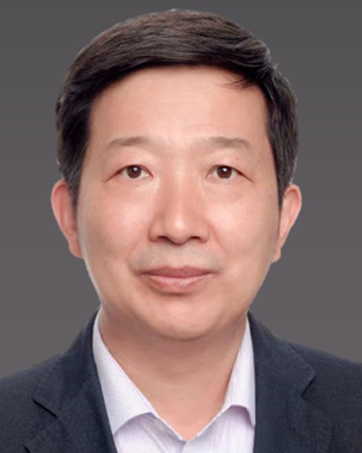}}]{Chengliang Yin} received the master’s and Ph.D. degrees in vehicle engineering from Jilin Industrial
University, Changchun, China, in 1996 and 2000, respectively. He is currently a Full Professor with Shanghai Jiao Tong University, Shanghai, China, where he is also the Vice Dean of the Institute of Automotive Engineering and the Vice Director of the National Engineering Laboratory for Automotive Electronic Control Technology. He is also an Advanced Technical Adviser with Shanghai Automobile and Dongfeng Automotive Group. His research interests include the control of automotive electronics and electric vehicles, particularly research and development of hybrid electric vehicles.
\end{IEEEbiography}
\vspace{30pt}
\begin{IEEEbiography}[{\includegraphics[width=1in,height=1.25in,clip,keepaspectratio]{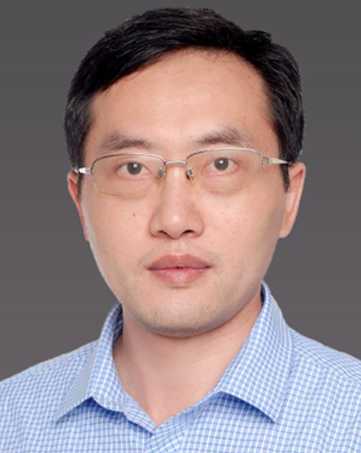}}]{Yong Zhang} received the B.S. and M.S. degrees in vehicle engineering from Chang’an University, Xi’an, China, in 1999 and 2002, respectively, and the Ph.D. degree in vehicle engineering from Shanghai Jiao Tong University, Shanghai, China, in 2005. He is currently an Assistant Professor with the School of Mechanical Engineering and the National Engineering Laboratory for Automotive Electronic Control Technology, Shanghai Jiao Tong University. His research interests include testing platform, vehicle electronics, and the control system of autonomous vehicles and electric vehicles
\end{IEEEbiography}

\end{document}